\documentclass{article}

\usepackage[main, final]{neurips_2025}
\usepackage{microtype}
\usepackage{graphicx}
\usepackage{placeins}
\usepackage{booktabs} %
\usepackage{multirow}
\usepackage{algorithm}
\usepackage{algorithmic}
\usepackage{subcaption}
\usepackage{hyperref}
\usepackage{wrapfig}
\usepackage[table]{xcolor}

\usepackage{array}
\usepackage{amsthm}
\usepackage{amssymb}
\usepackage{amsmath}
\usepackage{bbold}

\theoremstyle{plain}
\newtheorem{prop}{Proposition}

\theoremstyle{definition}

\theoremstyle{remark}

\newcommand{\Esp}{\mathbb{E}}

\newcommand{\Lrond}{\mathcal L}

\newcommand{\Xrond}{\mathcal X}
\newcommand{\Yrond}{\mathcal Y}

\newcommand{\xbold}{\mathbf{x}}

\newcommand{\zbold}{\mathbf{z}}

\newcommand{\vertiii}[1]{{\left\vert\kern-0.25ex\left\vert\kern-0.25ex\left\vert #1 
    \right\vert\kern-0.25ex\right\vert\kern-0.25ex\right\vert}}

\newcommand{\norm}[1]{\left|\left|#1\right|\right|}

\newcommand{\set}[1]{\left\{ #1\right\}}

\renewcommand{\leq}{\leqslant}

\usepackage{amsmath,amsfonts,bm}

\def\eqref#1{equation~\ref{#1}}

\def\1{\bm{1}}

\def\rmX{{\mathbf{X}}}
\def\rmY{{\mathbf{Y}}}

\DeclareMathAlphabet{\mathsfit}{\encodingdefault}{\sfdefault}{m}{sl}
\SetMathAlphabet{\mathsfit}{bold}{\encodingdefault}{\sfdefault}{bx}{n}

\newcommand{\R}{\mathbb{R}}

\DeclareMathOperator*{\argmax}{arg\,max}

\newcommand{\teach}{\mathsf{T}}
\newcommand{\teachk}{{\mathsf{T}_k}}
\newcommand{\stud}{{\mathsf{S}}}

\usepackage{amsmath}
\usepackage{amssymb}
\usepackage{mathtools}
\usepackage{amsthm}

\usepackage[capitalize,noabbrev]{cleveref}
\usepackage{minitoc}

\theoremstyle{plain}
\newtheorem{theorem}{Theorem}[section]
\newtheorem{proposition}[theorem]{Proposition}

\newtheorem{corollary}[theorem]{Corollary}
\theoremstyle{definition}

\theoremstyle{remark}
\newtheorem{remark}[theorem]{Remark}

\usepackage[textsize=tiny]{todonotes}

\begin{document}

\doparttoc %
\faketableofcontents %
\title{Learning Task-Agnostic Representations through Multi-Teacher Distillation}

\newcommand{\ups}{Universite Paris-Saclay}
\newcommand{\mila}{Mila - Quebec AI Institute}
\newcommand{\ets}{ETS Montreal}
\newcommand{\ills}{ILLS }
\newcommand{\livia}{LIVIA}
\newcommand{\mcgill}{McGill University}
\newcommand{\cnrs}{CNRS }

\author{
Philippe Formont\thanks{Equal Contribution}\\
\ups - \ets\\\mila\\\livia - \ills
\And
Maxime Darrin\footnotemark[1]\\
\mcgill - \ups\\\mila\\\ills
\And
Banafsheh Karimian\footnotemark[1]\\
\ets\\\ills - \livia
\And
Jackie CK Cheung\\
\mcgill \\ \mila
\And
Eric Granger\\
\ets \\ \ills - \livia\\
\And
Ismail Ben Ayed\\
\ets \\ \ills - \livia
\And
Mohammadhadi Shateri\\
\ets \\ \livia
\And
Pablo Piantanida\\
 \cnrs  - CentraleSupelec - \ups \\ \ills - \mila 
}

\maketitle

\begin{abstract}
Casting complex inputs into tractable representations is a critical step across various fields. Diverse embedding models emerge from differences in architectures, loss functions, input modalities and datasets, each capturing unique aspects of the input. Multi-teacher distillation leverages this diversity to enrich representations but often remains tailored to specific tasks. In this paper, we introduce a task-agnostic framework based on a ``majority vote" objective function. 
We demonstrate that this function is bounded by the mutual information between student and teachers' embeddings, leading to a task-agnostic distillation loss that eliminates dependence on task-specific labels or prior knowledge. 
Our evaluations across text, vision models, and molecular modeling show that our method effectively leverages teacher diversity, resulting in representations enabling better performance for a wide range of downstream tasks such as classification, clustering, or regression. Additionally, we train and release state-of-the-art embedding models, enhancing downstream performance in various modalities.
\end{abstract}

\section{Introduction}

Transforming complex inputs into tractable representations is crucial for numerous applications across different domains, from natural language processing~\citep{li2023angleoptimized, pimentel2023usefulness}, computer vision~\citep{kubota2024impressionclip, bhalla2024interpreting} to bioinformatics~\citep{morganfp, wang2022chemicalreactionaware}. This is done using embedders, often large pretrained models~\citep{touvron2023llama, jiang2023mistral}, that project  objects (image, text, molecules, \ldots) into numerical representations, enabling various downstream tasks~\citep{murphy2013machine, DBLP:journals/corr/VilnisM14}. 

Variations in model architecture, training paradigms (e.g., unsupervised vs. supervised), and objective functions (e.g., masked language modeling and contrastive learning) result in embedders that capture different aspects of the same input. 
To leverage this diversity, a common practice is to combine them into a single model through multi-teacher Knowledge Distillation (KD)~\citep{zhang2023adaptive}.

Not only are these methods cost-effective at inference time~\citep{hinton2015distilling, frosst2017distilling}, they are also extremely useful to compress knowledge from larger models into smaller ones for resource-constrained environments~\citep{pan2022extreme, wang2023learnable, zhang2023adaptive}, or mend the weights of models whose architectures have been altered~\citep{muralidharan2024compact}.
Most existing approaches, however, focus on single-task distillation. In this setting, the student model either learns to mimic teacher representations for a specific task~\citep{dvornik2019diversity}, or the distillation process is explicitly paired with task-specific information. While effective, such methods cannot be used for or generalized to unseen tasks, requiring a new distillation process to be performed for every new task. \textbf{Our goal is to learn a highly informative representation that retains maximal utility across a wide range of downstream tasks. } In other words, we aim to maximize information density within a single representation, enabling general-purpose adaptability without sacrificing performance.

Task-agnostic multi-teacher distillation aims to compress teacher representations into a single student embedder, such that the student representation captures as much information as all the teachers combined.
To our knowledge, few works address task-agnostic distillation from multiple teachers. Existing approaches often rely on mean squared error (MSE) loss and cross-encoder heads~\citep{navaneet2022simregregressionsimpleeffective}, which can be unstable in high-dimensional spaces~\citep{farebrother2024stopregressingtrainingvalue}.

To overcome these limitations, we introduce a novel task-enabling setting to task-agnostic multi-teacher distillation.
Our goal is to develop representations that capture the maximum amount of information about the data distribution, ensuring their applicability to a wide range of tasks, even in the absence of prior knowledge about those tasks.
We train the student model to learn representations that, when applied to downstream tasks, generate predictions consistent with the majority of predictions from the teachers' representations.
This approach allows our method to leverage the collective knowledge of the teachers' ensemble. To achieve this, we introduce an ensembling loss that measures the agreement between the Bayesian predictor based on the student's embeddings and the Bayesian predictors based on the teachers' embeddings. We show that this loss can be bounded independently of the task, using the conditional differential entropy of the teachers' embeddings given the student's output, thus providing a task-agnostic student-teacher reconstruction loss.

\textbf{Contributions.}
In this study, we investigate the following research question: 
\textbf{How can the knowledge from multiple large embedding models be effectively distilled and integrated into a smaller one to produce a more general-purpose representation?} Our main contributions are threefold:
\begin{enumerate}
    \item \textbf{A task-enabling setting.}  
    We frame the multi-teacher distillation problem in a task-enabling setting, in which we study the relationship between the Bayes classifiers obtained from the students and the teachers' embeddings.
    We prove a simple, yet powerful result: the conditional entropy of the teachers given the student's output controls the probability of the student's Bayesian predictor disagreeing with the teachers' for any task.
     
    \item \textbf{A tractable implementation.} We leverage a recent differentiable high-dimensional Gaussian-Mixture based estimator of the differential conditional entropy to formulate an information-theoretic loss. This loss maximizes the mutual information between the student and all teachers, resulting in a principled, task-agnostic distillation objective.

    \item \textbf{High-quality generalized embedders.}
    Our method enhances distillation capabilities across three application domains: molecular modeling,  natural language processing and computer vision. We release trained students achieving competitive performance on a wide range of downstream tasks, e.g., classification, regression, clustering, and sentence similarity.
\end{enumerate}

\section{Related Work}

\paragraph{Task-oriented distillation.} KD is widely used for transferring knowledge from one or a set of teachers to a student model \citep{Gou_2021} to improve the performance of the student on a given task \citep{zhang2019your, 8100237}.
This is typically done by transferring logits \citep{sun2024logit}; \textit{i.e.} the models' output, features \citep{wang2023learnable, sarkar2024xkd}, relational information \citep{dong2024cluster, dong-etal-2021-hrkd}, or a mixture of them \citep{liu2021exploring}. Similarly, \citep{qiu2024transferring} uses a regularization term to distill the task-relevant information from the large teacher to the small student. We depart from these methods by focusing on distilling task-agnostic representations.

\paragraph{Task-oriented multi-teacher distillation.} A common method for multi-teacher KD is averaging the teachers' logits and transferring the result to the student \citep{dvornik2019diversity, hinton2015distilling}. However, this approach is not ideal when the performance of the teachers is uncertain. Alternative methods include using gate networks \citep{zhu2020ensembled}, reinforcement learning agents \citep{yuan2020reinforced}, and other methods \citep{MA2024123892, att2022, zhang2023adaptive} to perform teacher selection or evaluation. Due to challenges in distilling knowledge among diverse architectures, multi-teacher KD research mainly focuses on logit distillation. Other techniques were also explored, such as multi-teacher feature ensemble \citep{ye2024knowledge}, contrasting feature distillation \citep{Li2024}, and cosine similarity-based methods for various tasks \citep{Ma2024, aslam2024multiteacherprivilegedknowledge, aslam2023privileged}. Ensemble-based methods have also been proposed to mitigate over-smoothing and leverage teacher diversity, such as by aggregating structured predictions before distillation \citep{shayegh2024ensemble}. Although successful, most multi-teacher feature distillation methods remain oriented to only one or a few tasks.

\paragraph{Task-agnostic and self-supervised features distillation.} To the best of our knowledge, few works address task-agnostic representation distillation. Several approaches assume strong limitations, such as requiring the student to have the same architecture as the teachers~\citep{lianghomodistil, xu-etal-2022-importance}, or requiring fine-tuning the teachers to then distill their representations~\citep{liu2023ernie}. Other methods induce requirements on the students, limiting their extension to a general multi-teacher setting.  Notably~\citep{gao2022discoremedyselfsupervisedlearning} relies on vision-specific data augmentation, RoB~\citep{duval2023simplerecipecompetitivelowcompute} focuses on the distillation of joint-embedding approaches, AttnDist~\citep{wang2022attention} is only applicable to single teacher, ~\citep{song2023multi} need the teacher and student to have the same architecture, and SEED~\citep{fang2021seedselfsuperviseddistillationvisual} requires the student and the teacher to have the same embedding dimension.
Finally, CompRess~\citep{NEURIPS2020_975a1c8b} introduced a distillation method ensuring that the embeddings of the student and the teacher encode a similar nearest-neighbor graph, which would be unstable in a multi-teacher setting. Other approaches such as contrastive learning~\citep{feng2024unicornunifiedcontrastivelearning, liu2022pretraining, xu2022baginstancesaggregationboosts} focus on distilling relational relationships between the samples, such as nearest neighbors preservation~\citep{noroozi2018boosting} or angle preserving distillation\citep{park2019relational}. SimReg~\citep{navaneet2022simregregressionsimpleeffective}, however, trains the student jointly with cross-encoding heads to directly reconstruct the teacher's features using an MSE loss.

\begin{figure*}
\centering
\includegraphics[width=0.95\textwidth]{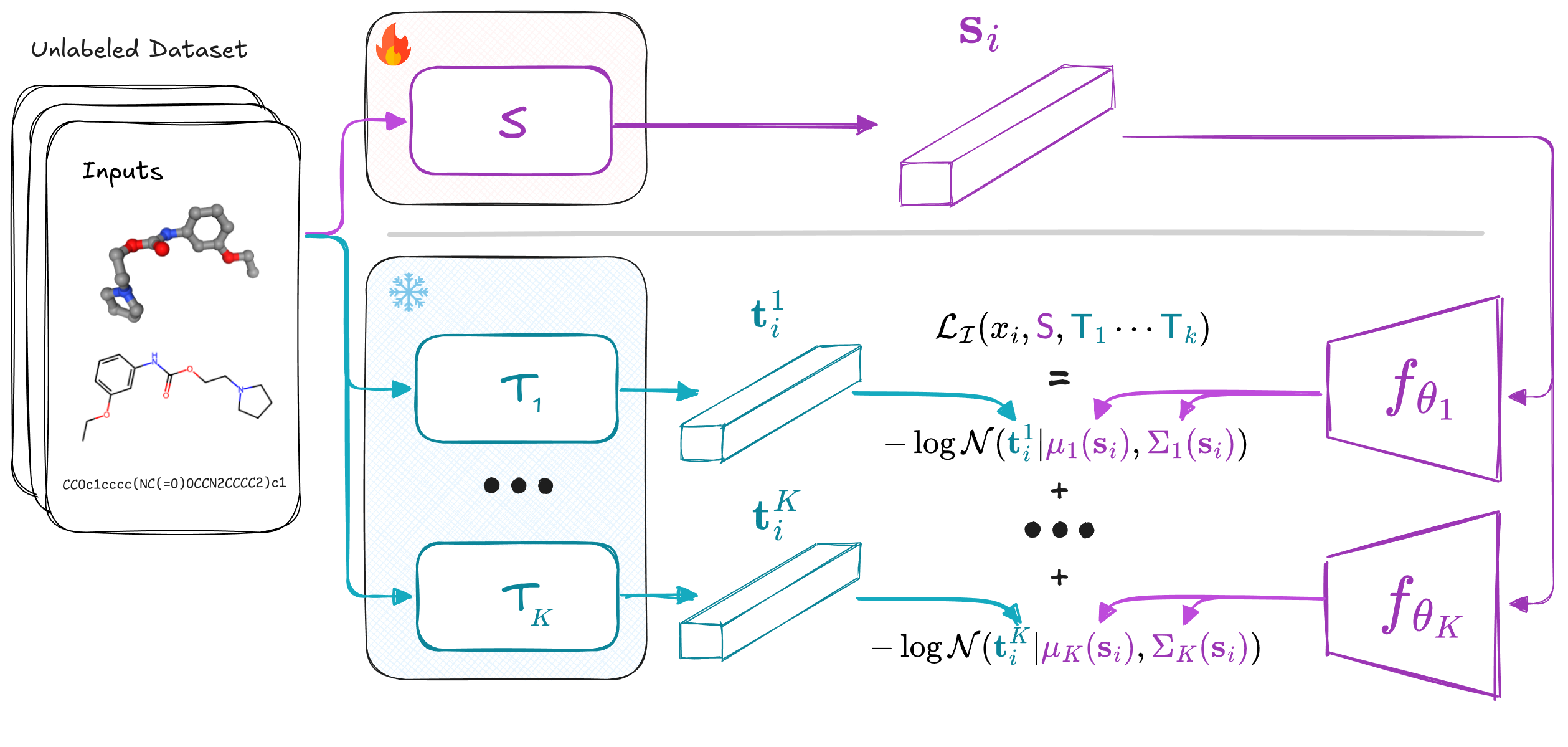}
\caption{
    \textbf{Unsupervised training of our student through task-agnostic distillation.}
    The student embedder $S$ is trained to minimize the negative log-likelihood of multiple teachers' outputs conditioned on the student’s predictions. 
    During this multi-teacher distillation procedure, both the student's weights and those of the teacher-specific Gaussian kernels $\{f_{\theta_k}\}_{k\leq K}$ are updated in an end-to-end fashion.
    Post-training, we discard the Gaussian kernels and evaluate the student embedders by freezing their weights and training a feed-forward network on their embeddings for an unseen dataset. 
    }
\label{fig:method}
\vspace{-0.4cm}
\end{figure*}

\paragraph{Interval estimation.} 
While SimReg performs its distillation through pointwise estimation with MSE, it is well known in the reinforcement learning literature that these standard regression methods are difficult to train~\citep{farebrother2024stopregressingtrainingvalue}.
On the other hand, replacing traditional regression scheme by maximum-likelihood training of Gaussian kernels appears to be more stable~\citep{stewart2023regressionclassificationinfluencetask} and effective in Value learning~\citep{bellemare2017distributionalperspectivereinforcementlearning}. 
We extend this idea in the context of embedder distillation by using Gaussian kernels to estimate the conditional distribution of the teachers' embeddings given the student embedding and show that it is directly connected to maximizing the mutual information between the student and the teacher.

\section{Distilling Representation Through Gaussian Kernels}

We denote the input space by $\Xrond$ and the corresponding input distribution by $P_\rmX$. We assume we have access to a dataset $\mathcal{D}=\set{\xbold_i}_{i=1}^n$, where samples are drawn i.i.d. according to $P_\rmX$. We consider a set of $K$ different teacher embedders, $\teachk : \Xrond \rightarrow \R^{d_k}$, for $k \in \set{1, \ldots, K}$, each mapping  inputs to potentially different embedding spaces of dimension $d_k$.

\subsection{From a task-oriented setting to a task-agnostic loss}
\label{sec:from_setting_to_loss}

Our goal is to train a representation model capable of effectively handling any downstream task, by leveraging diverse representations from diverse pretrained teachers (\autoref{fig:method}). To do so, we first measure the agreement between the student's Bayes classifier and the teachers' for any given task. First, we demonstrate that it can be bounded by the conditional entropy of the teacher's embedding given the student's, which does not depend on the considered task.

Let us consider a task characterized by a target set $\Yrond$ of discrete concepts and the feature space $\mathcal{X}$ with joint probability measure $P_{\rmY\rmX} \in \mathcal{P}(\mathcal{Y}\times \mathcal{X})$. 
For every projection of the features through the different teachers, the Bayes decision rule is given by 
$\displaystyle c_{\teachk}^* \triangleq  \argmax_{c : \R^{d_k} \rightarrow \Yrond} \Esp_{\rmX, \rmY}\left[\mathbb{1}\left[c(\teachk(\rmX)) = \rmY\right]\right]$ and for the student: 
$\displaystyle c_{\stud}^* \triangleq \argmax_{c : \R^d \rightarrow \Yrond} \Esp_{\rmX, \rmY}\left[\mathbb{1}\left[c(\stud(\rmX)) = \rmY\right]\right]$.

Our goal is to minimize the probability that the student's Bayesian classifier deviates from the predictions of the teachers'. This approach has been shown to enhance performance in most cases by reducing both bias and variance, while improving robustness and generalizability \citep{dietterich-iwmcs-00, scimeca2023shortcut, allen2020towards, theisen2024ensembles}. In other words, we aim to minimize the probability that the student's decision differs from that of each teacher:
\begin{equation}
    \label{eq:ideal_loss}
    \mathcal{L}^*(\rmX, \rmY,\stud, \teach_1, \ldots, \teach_K) = \frac1K\sum_{k=1}^K 
    \underbrace{\Pr \left( c_{\stud}^*(\stud(\rmX)) \neq c_{\teachk}^*(\teachk(\rmX)) \right)}_{ \text{\shortstack{Probability that the student Bayesian classifier's \\ output is different from the $k^{\text{th}}$ teacher's}}}.
\end{equation}
where the loss depends on the joint distribution $(\rmX,\rmY)$, through the definition of the Bayesian classifiers.

We leverage recent results on the performance of the Bayes classifiers to bound the probability of getting two different outcomes using the Bayes classifiers operating on two different
projections of the input space.
\begin{proposition}[\cite{darrin2024textttcosmicmutualinformationtaskagnostic}]\label{prop:cosmic}
    Let $C_\teachk = c^*_\teachk(\teachk(X))$ and $C_\stud = c^*_\stud(\stud(t))$ denote the out-
come of the Bayes classifier observing the output of the teacher $\teachk$ and the student $\stud$ on a given task $Y$, respectively.
\[
    \Pr \left( C_\stud \neq C_\teachk \right) \leq 1- \exp{\left(-h\left(\teachk(X)|\stud(X)\right)\right)}
\]

\end{proposition}

\begin{corollary}[Training objective]
\label{prop:loss_upper_bound}
    By applying \hyperref[prop:cosmic]{Prop. 3.1} to \autoref{eq:ideal_loss}
    for any given joint distribution $P_{\rmX\rmY}$, we have 
    \begin{equation}
        \label{eq:ideal_loss_bound}
        \mathcal{L}^*(\rmX,\rmY,\stud, \teach_1, \ldots, \teach_K) \leq 
        1 - \exp\Big( - \underbrace{\frac{1}{K} \sum_{k=1}^K 
        h(\teachk(\rmX) | \stud(\rmX))}_{\text{Negative log likelihood}} \Big).
    \end{equation}
      This corollary directly follows from the concavity of $t\rightarrow 1 - \exp(-t) $ (see \autoref{sec:appendix_theoretical_result}).
\end{corollary}

\begin{remark}
This bound over our ideal loss $\mathcal{L}^*$ is independent of the specific task and depends solely on the conditional entropy of the teacher embeddings given the student embeddings. Therefore, optimizing the student to minimize this loss provides a task-agnostic approach to aligning its Bayesian classifier predictions with the ensemble of teachers' predictions, regardless of the downstream task.
\end{remark}

\subsection{Student training}

\paragraph{Estimation of the conditional entropy.} To evaluate the conditional entropy of the teachers' embeddings given the student's embedding, we need a kernel to learn their conditional distribution $\hat{p}(\teachk(\rmX) | \stud(\mathbf{X}))$ as presented in \autoref{fig:method}.
To this end, we use a parametric Gaussian model whose parameters $\mu_k(\stud(\mathbf{X}))$ and $\Sigma_k(\stud(\mathbf{X}))$ are learned during the student's training~\citep{pichler2022differential}.

\paragraph{Loss function.} Following the above reasoning, we propose to train the student embedder $\stud$ by minimizing the negative log-likelihood  of the teachers' embeddings given the student's embedding, where the likelihood is estimated using Gaussian Kernels as follows:\begin{multline}
{
    \hat\Lrond(\rmX,\stud, \teach_1, \ldots, \teach_K) = \frac1K\sum_{k=1}^K h(\teachk(\rmX) | \stud(\rmX)) } \\ { \leq 
     \frac1K\sum_{k=1}^K 
    \Esp_{\rmX} \Big[ - \log \mathcal{N}\big(\teachk(\rmX) \,\big|\, \mu_k(\stud(\rmX)), \Sigma_k(\stud(\rmX)) \big) \Big],}
    \label{eq:actual_loss}
\end{multline}
where $\mathcal{N}(\cdot | \mu, \Sigma)$ is the Gaussian distribution with mean $\mu$ and covariance $\Sigma$. In our setting, minimizing the conditional entropy $h(\teachk(\rmX) | \stud(\rmX))$, exactly corresponds to maximizing the mutual information $I(\teachk(\rmX) ; \stud(\rmX)) = h(\teachk(\rmX)) - h(\teachk(\rmX) | \stud(\rmX))$ since for each teacher $h(\teachk(\rmX))$ is constant w.r.t of the student. This also applies to the bound in~\autoref{eq:ideal_loss_bound}.

\paragraph{Training procedure.} 
We train both the student and the different kernels in an end-to-end fashion by minimizing the loss function $  \hat{\mathcal{L}}$. It boils down to minimizing the negative log-likelihood of the teachers' embeddings given the student's embedding. We use the Adam optimizer to minimize the loss function. See~\autoref{sec:appendices_detailed_method} for the detailed training algorithm.
To reduce the computational cost, we first embedded the entirety of the training set using the teachers and store them. We can then build training batches by sampling from the pre-computed embeddings. 

\paragraph{Baselines and Evaluation.} We consider two widely used multi-teacher feature distillation methods, MSE, used in SimReg~\citep{navaneet2022simregregressionsimpleeffective} and Cosine similarity (see~\autoref{sec:appendix_baselines} for more information).
To evaluate the representations learned by the student, for each modality, we run different benchmarks evaluating its performance on a wide variety of downstream tasks. For classification and regression tasks, we train a small feedforward network on top of the embeddings (the backbones are considered frozen) on different tasks and evaluate its performance.

\section{Text Embedders}
\label{sec:nlp}

\subsection{Experimental setting}
We focus on distilling high-performing and large models into significantly smaller ones. Indeed, modern models in NLP are extremely large and costly to train\footnote{\url{https://github.com/ills-montreal/nlp-distill}}.
Thus, we aim to produce the best possible models for a given weight category, pushing the size/performance of the Pareto frontier (\autoref{fig:pareto_frontier_nlp}), and not necessarily competing with the largest models. We distill from four teachers ranging from $433M$ parameters to $7B$ into students ranging from $20$M to $335$M parameters based on the nowflakes~\citep{merrick2024arcticembedscalableefficientaccurate} embedders.

\textbf{Teachers and student.} We select four freely available embedding models from the Huggingface hub~\citep{wolf2020huggingfacestransformersstateoftheartnatural} (See \autoref{sec:appendix_nlp_teachers_list} for a detailed list of the teachers) whose evaluations are available in the MTEB benchmark~\citep{muennighoff2023mteb}. 
To ensure having a point of comparison, we select teachers of different sizes and performances. 
Notably, SFR-Embeddings-R\_2 is more than ten points stronger than the other three (smaller) teachers. As students we use snowflakes~\citep{merrick2024embeddingclusteringdataimprove, merrick2024arcticembedscalableefficientaccurate} models xs ($22$M), s ($33$M), m ($109$M) and l ($335$M) and we further train them using our distillation method (See \autoref{sec:appendix_nlp_hyperparameters}). 

\textbf{Embedder evaluation.} Evaluating NLP models is notably challenging, and the common practice of evaluating a model using multi-task benchmarks may not be indicative of model capabilities~\citep{ecbd-bench}. 
For lack of better options and because it is currently the most widely accepted benchmark, we rely on the evaluation provided by the MTEB benchmark~\citep{muennighoff2023mteb} on $33$ tasks encompassing clustering ($11$ datasets), sentence similarity ($10$ datasets) and classification tasks ($12$ datasets). We compare our models with distilled and non-distilled ones from the MTEB leaderboard.

\textbf{Training set.} We gathered different common datasets used for training embedders and collected $6$ million entries from the Huggingface Hub, including Specter~\citep{specter2020cohan}, T5~\citep{ni2021sentencet5}, Amazaon QA~\citep{10.1145/2507157.2507163}, IMDB~\citep{maas-EtAl:2011:ACL-HLT2011}, SNLI~\citep{bowman-etal-2015-large}, QQP triplets from Quora, AG News~\citep{Zhang2015CharacterlevelCN}, MEDI dataset~\citep{su2023embeddertaskinstructionfinetunedtext} and the DAIL Emotion dataset~\citep{saravia-etal-2018-carer}. We provide the dataset statistics in \autoref{sec:appendix_nlp_model_dataset_statistics}. 
The datasets are all flattened, such that if the original had two columns (e.g., sentence $1$ and $2$ in the SNLI dataset), we end up with twice the number of entries, one for each sentence, and we deduplicated the dataset. Models are trained for two epochs with batch size $16$ on NVIDIA V100.

\subsection{Distillation performance}
\label{sec:nlp_perfs}

\begin{table}
\caption{Performance of our distilled models compared to the stronguest models of similar sizes from the MTEB Benchmark on classification tasks. Our $109$M parameters model outperform significantly models $3$ times bigger exhibiting exceptional information density.}
\label{tab:mteb_classification_per_size_merged_downsampled}
\resizebox{\textwidth}{!}{
    \begin{tabular}{lllc|cccccccccccc|c}
    \toprule
     &  & Task & Size & \rotatebox{90}{\shortstack{Amazon \\ Counterfactual}} & \rotatebox{90}{\shortstack{Amazon \\ Polarity}} & \rotatebox{90}{\shortstack{Amazon \\ Reviews}} & \rotatebox{90}{\shortstack{Banking77}} & \rotatebox{90}{\shortstack{Emotion}} & \rotatebox{90}{\shortstack{Imdb}} & \rotatebox{90}{\shortstack{MTOPDomain}} & \rotatebox{90}{\shortstack{MTOPIntent}} & \rotatebox{90}{\shortstack{Massive \\ Intent}} & \rotatebox{90}{\shortstack{Massive \\ Scenario}} & \rotatebox{90}{\shortstack{Toxic \\ Conversations}} & \rotatebox{90}{\shortstack{Tweet \\ Sentiment \\ Extraction}} & Avg. \\
         &   & Model &  &  &  &  &  &  &  &  &  &  &  &  &  &  \\
    \midrule
    \multirow[c]{5}{*}{xs} & \multirow[c]{3}{*}{Bas.} & GIST & 23M & \underline{72.9} & \bfseries \underline{87.2} & 42.6 & \underline{84.2} & 52.1 & 78.5 & \underline{94.8} & \underline{77.7} & 73.2 & 76.7 & \bfseries \underline{72.9} & \underline{59.9} & 72.7 \\
     &  & Ivysaur & 23M & 72.1 & \underline{86.7} & \bfseries \underline{42.7} & 81.9 & 45.4 & 80.8 & 92.1 & 71.9 & 70.3 & 74.9 & 65.5 & 58.7 & 70.2 \\
     &  & gte-tiny & 23M & 71.8 & 86.6 & \underline{42.6} & 81.7 & 44.7 & 80.5 & 91.8 & 69.9 & 70.1 & 74.9 & \underline{71.0} & 58.6 & 70.3 \\
    \cmidrule(lr){2-17}
     & MSE & Student-xs & 23M & 71.6 & 86.2 & 42.3 & 83.6 & \underline{57.5} & \bfseries \underline{83.5} & 94.5 & 75.4 & \underline{74.3} & \bfseries \underline{80.4} & 66.3 & 59.3 & \underline{72.9} \\
    \cmidrule(lr){2-17}
     & NLL & Student-xs & 23M & \bfseries \underline{76.5} & 84.9 & 42.4 & \bfseries \underline{85.8} & \bfseries \underline{58.0} & \underline{81.1} & \bfseries \underline{95.2} & \bfseries \underline{79.9} & \bfseries \underline{75.8} & \underline{80.4} & 68.1 & \bfseries \underline{60.1} & \bfseries \underline{74.0} \\
    \cmidrule(lr){1-17} \cmidrule(lr){1-17} \cmidrule(lr){2-17}
    \multirow[c]{5}{*}{s} & \multirow[c]{3}{*}{Bas.} & bge-small-en-v1.5 & 33M & 73.8 & 92.8 & 47.0 & 85.7 & 47.8 & \bfseries \underline{90.6} & 93.4 & 74.8 & 74.8 & 78.7 & \underline{69.9} & 60.5 & 74.1 \\
     &  & GIST & 33M & 75.3 & \underline{93.2} & \underline{49.7} & \underline{86.7} & 55.9 & 89.5 & \underline{95.5} & 79.1 & 75.5 & 79.2 & \bfseries \underline{72.8} & \underline{61.0} & \bfseries \underline{76.1} \\
     &  & NoInstruct & 33M & \underline{75.8} & \bfseries \underline{93.3} & \bfseries \underline{50.0} & 86.4 & 55.1 & \underline{90.2} & 95.3 & \underline{79.6} & \underline{76.0} & 79.3 & 69.4 & \bfseries \underline{61.3} & \underline{76.0} \\
    \cmidrule(lr){2-17}
     & MSE & Student-s & 33M & 72.6 & 90.3 & 44.3 & 84.2 & \underline{56.5} & 88.8 & 94.9 & 77.2 & 75.4 & \bfseries \underline{81.2} & 64.9 & 60.4 & 74.2 \\
    \cmidrule(lr){2-17}
     & NLL & Student-s & 33M & \bfseries \underline{77.3} & 89.2 & 43.8 & \bfseries \underline{86.7} & \bfseries \underline{58.0} & 88.3 & \bfseries \underline{95.5} & \bfseries \underline{81.9} & \bfseries \underline{76.7} & \underline{80.7} & 66.1 & 60.6 & 75.4 \\
    \cmidrule(lr){1-17} \cmidrule(lr){1-17} \cmidrule(lr){2-17}
    \multirow[c]{6}{*}{m} & \multirow[c]{4}{*}{Bas.} & bge-base-en-v1.5 & 109M & 76.2 & \underline{93.4} & \underline{48.9} & 87.0 & 51.9 & \bfseries \underline{90.8} & 94.2 & 76.9 & 76.2 & 80.2 & 71.6 & 59.4 & 75.5 \\
     &  & GIST & 109M & 76.0 & \bfseries \underline{93.5} & \bfseries \underline{50.5} & \underline{87.3} & 54.7 & \underline{89.7} & 95.3 & 78.1 & 76.0 & 79.6 & \bfseries \underline{72.4} & 59.3 & \underline{76.0} \\
     &  & e5-base-4k & 112M & \underline{77.8} & 92.8 & 46.7 & 83.5 & 47.0 & 86.2 & 93.7 & 75.3 & 73.0 & 77.7 & \underline{72.1} & 60.4 & 73.8 \\
     &  & e5-base-v2 & 110M & \underline{77.8} & 92.8 & 46.7 & 83.5 & 47.0 & 86.2 & 93.7 & 75.3 & 73.0 & 77.7 & \underline{72.1} & 60.4 & 73.8 \\
    \cmidrule(lr){2-17}
     & MSE & Student-m & 109M & 76.6 & 89.1 & 44.7 & 87.2 & \bfseries \underline{60.8} & 88.0 & \underline{95.7} & \underline{81.6} & \underline{77.7} & \underline{82.2} & 67.3 & \underline{60.5} & 76.0 \\
    \cmidrule(lr){2-17}
     & NLL & Student-m & 109M & \bfseries \underline{79.6} & 89.5 & 45.8 & \bfseries \underline{88.0} & \underline{59.7} & 88.3 & \bfseries \underline{96.2} & \bfseries \underline{83.9} & \bfseries \underline{78.6} & \bfseries \underline{82.7} & 67.1 & \bfseries \underline{61.3} & \bfseries \underline{76.7} \\
    \cmidrule(lr){1-17} \cmidrule(lr){1-17} \cmidrule(lr){2-17}
    \multirow[c]{7}{*}{l} & \multirow[c]{5}{*}{Bas.} & bge-large-en-v1.5 & 335M & 75.8 & 92.4 & 48.2 & 87.8 & 51.5 & \bfseries \underline{92.8} & 94.6 & 79.5 & \bfseries \underline{77.6} & 80.5 & 70.9 & 59.9 & 76.0 \\
     &  & GIST & 335M & 75.6 & \underline{93.4} & \underline{49.1} & \bfseries \underline{88.1} & 54.7 & 91.2 & \underline{95.2} & 78.2 & 76.2 & 79.3 & \bfseries \underline{71.9} & 59.2 & \underline{76.0} \\
     &  & UAE-Large-V1 & 335M & 75.5 & 92.8 & 48.3 & 87.7 & 51.8 & 92.8 & 94.0 & 76.9 & 76.5 & 79.8 & 71.1 & 59.8 & 75.6 \\
     &  & ember-v1 & 335M & 76.1 & 92.0 & 47.9 & \underline{87.9} & 52.0 & 92.8 & 94.6 & 79.3 & 77.4 & 80.5 & 71.4 & 60.0 & 76.0 \\
     &  & mxbai-embed-large-v1 & 335M & 75.0 & \bfseries \underline{93.8} & \bfseries \underline{49.2} & 87.8 & 50.9 & \underline{92.8} & 94.0 & 76.8 & 76.2 & 80.0 & \underline{71.5} & 59.7 & 75.6 \\
    \cmidrule(lr){2-17}
     & MSE & Student-l & 335M & \underline{77.3} & 84.5 & 43.4 & 86.0 & \underline{60.0} & 82.7 & 95.1 & \underline{79.8} & 76.3 & \underline{81.3} & 65.8 & \underline{60.2} & 74.4 \\
    \cmidrule(lr){2-17}
     & NLL & Student-l & 335M & \bfseries \underline{81.5} & 88.1 & 45.9 & 86.9 & \bfseries \underline{60.4} & 88.2 & \bfseries \underline{95.6} & \bfseries \underline{83.2} & \underline{77.5} & \bfseries \underline{81.4} & 67.7 & \bfseries \underline{62.2} & \bfseries \underline{76.5} \\
    \cmidrule(lr){1-17} \cmidrule(lr){1-17} \cmidrule(lr){2-17}
    \bottomrule
    \end{tabular}
}
\end{table}

\begin{figure}[htbp]
    \centering
    \begin{subfigure}[b]{0.48\textwidth}
        \centering
        \includegraphics[width=\linewidth]{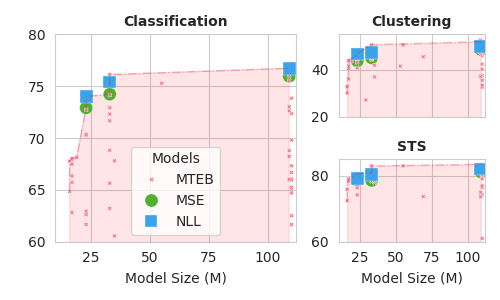}
        \caption{}
        \label{fig:pareto_frontier_nlp}
    \end{subfigure}
    \hfill
    \begin{subfigure}[b]{0.48\textwidth}
        \centering
        \includegraphics[width=\linewidth]{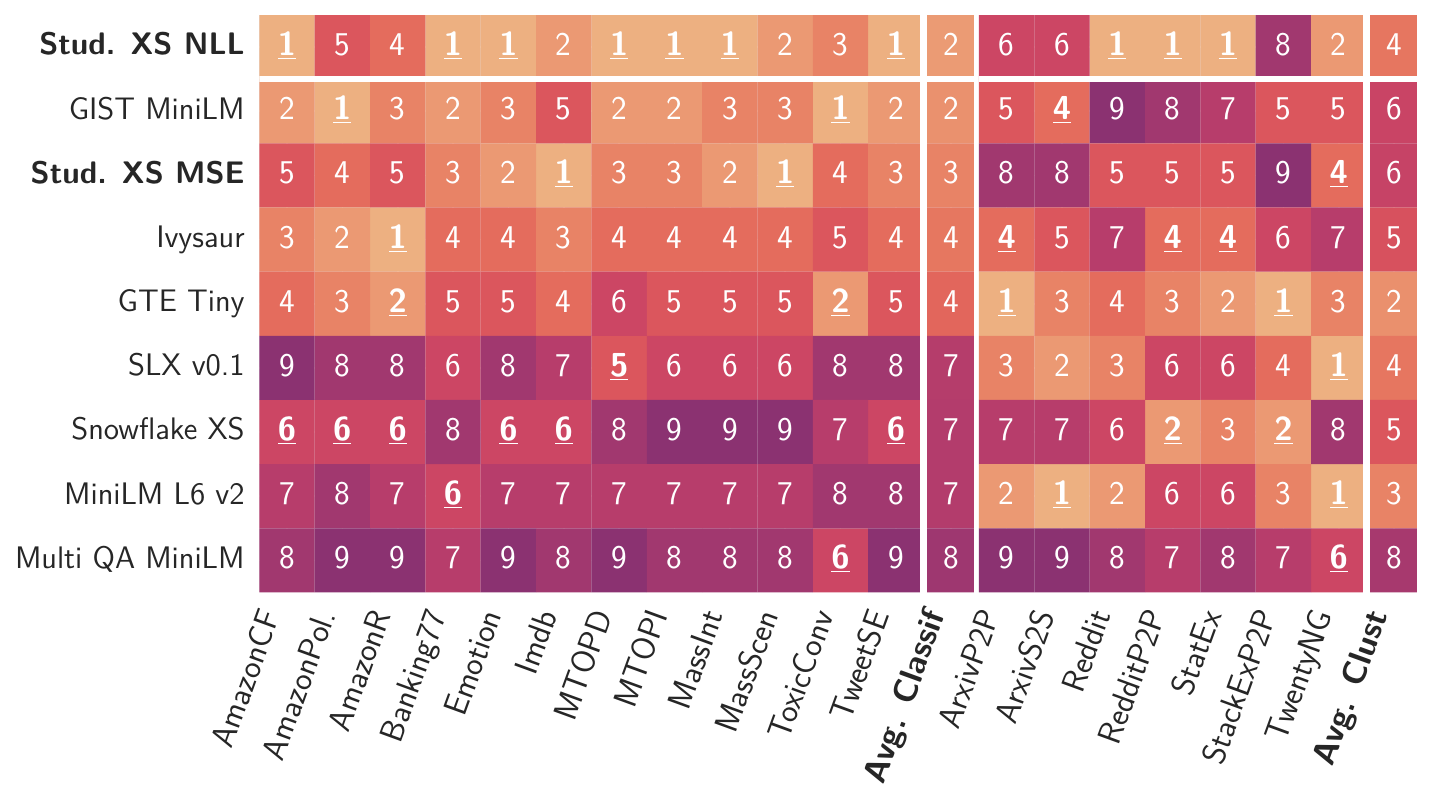}
        \caption{}
        \label{fig:global_ranking_xs}
    \end{subfigure}
    \caption{(a) \textbf{Pareto frontier size/performance in NLP.} Our method (in blue) yields Pareto optimal model. (b) \textbf{Global ranking of embedders} on clustering and classification tasks for our xs model ($23$M). The NLL-distilled model rank $1$ in most tasks and in average, outperforming all other baselines of its weight category and closing the gap with models 10 times bigger.}
    \label{fig:combined_nlp}
\end{figure}

\textbf{Task performance.} Our method produces models that exhibit strong performance on a large variety of tasks, ranking first amongst all models of similar size in the MTEB benchmark on most of the tasks (\autoref{fig:global_ranking_xs}). Notably, we observe that our method produces models that are competitive for almost all the tasks, whereas other models appear more specialized. 
We provide the actual accuracy of our models on classification tasks in \autoref{tab:mteb_classification_per_size_merged_downsampled}. 
We provide the full results for all model sizes in \autoref{sec:appendix_nlp_classification}. 

\textbf{Pareto frontier.} Our goal with distillation is to increase information density of models to reduce computational costs and memory footprint, we show in \autoref{fig:pareto_frontier_nlp} that our method can pack more information into fixed-size models. Interestingly, our medium-sized model ($109$M parameters) outperforms all the models three times its size and even our $335$M model under the same training setting. In addition, our small models outperform all previous model of their weight category, notably yielding a $2$-point gain on average classification accuracy on the MTEB over the previous \textit{state-of-the-art} efficient GIST-based embedders~\citep{solatorio2024gistembed}. 

\textbf{Comparison with standard MSE distillation.} Consistent with results from reinforcement learning and interval estimation\citep{stewart2023regressionclassificationinfluencetask}, training the student to match the teachers' embeddings using MSE loss results in consistently worse models.

\textbf{Limitations of the embedding space structure.} 
Our metric, which optimizes mutual information between the student and teachers, does not impose structure on the embedding space. Given that information remains invariant under invertible transformations, let \( f_1 \) and \( f_2 \) be differentiable and invertible mapping functions (diffeomorphisms); thus, \( I(X;Y) = I(f_1(X); f_2(Y)) \). Consequently, our objective does not ensure the preservation of structural properties, such as pairwise cosine similarity, in the teachers' embedding space. Nonetheless our method maintains competitive performance in both clustering and Semantic Textual Similarity (STS) (see Appendix \ref{sec:appendix_nlp_detailed_results}).

\section{Molecular Embedders}
\label{sec:mol_model}

We further our method in molecular modeling, enabling the distillation of a student with models leveraging different modalities to represent a molecule: text, graph, and 3D point clouds. 

\begin{figure}[h]
    \centering
    \includegraphics[width=1\linewidth,trim={0 0.3cm 0 0.3cm},clip]{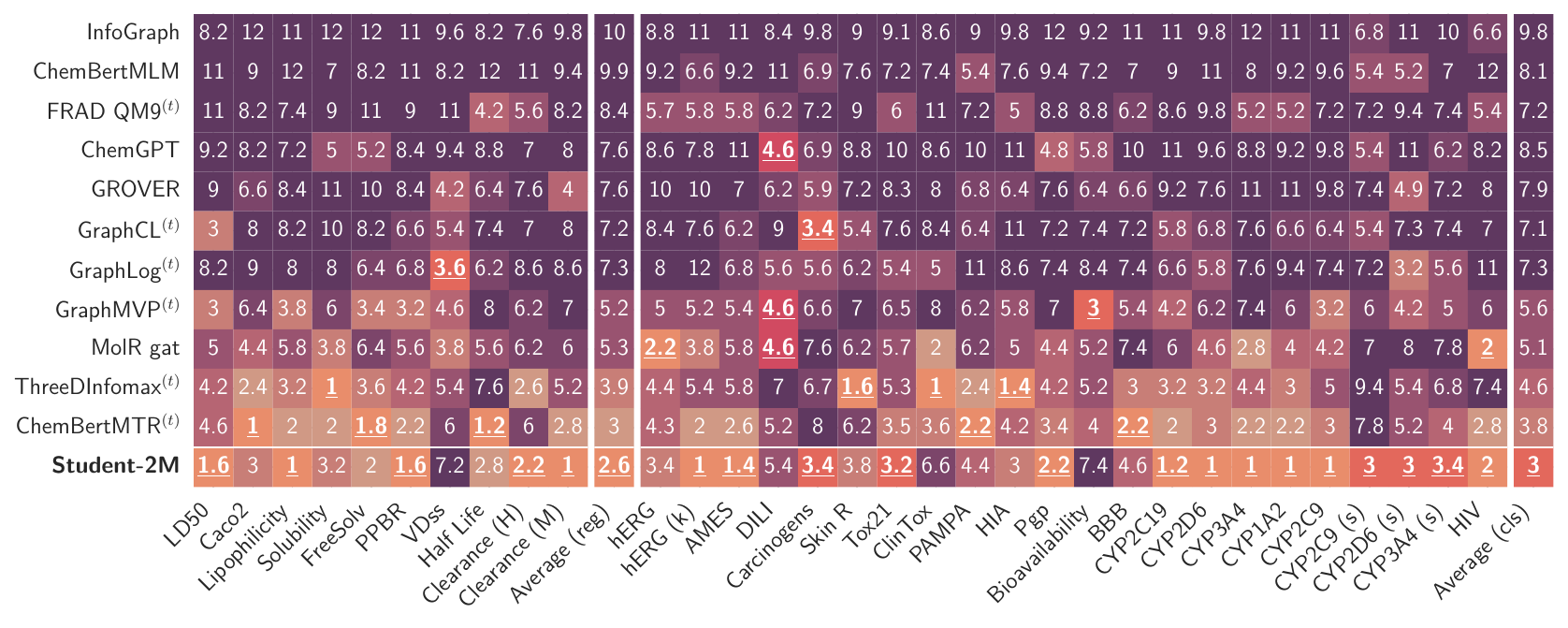}
    \caption{
    \textbf{Ranking on the TDC ADMET tasks.}
    Our student consistently achieves competitive performances across the evaluated tasks compared to its teachers (denoted by ${}^{(t)}$) and the other baselines, achieving the best average rank for both regression and classification tasks.
    }
    \label{fig:hmp_rankings}
\end{figure}

\begin{figure*}
    \centering
    \begin{subfigure}{0.56\linewidth}
        \includegraphics[width=\linewidth]{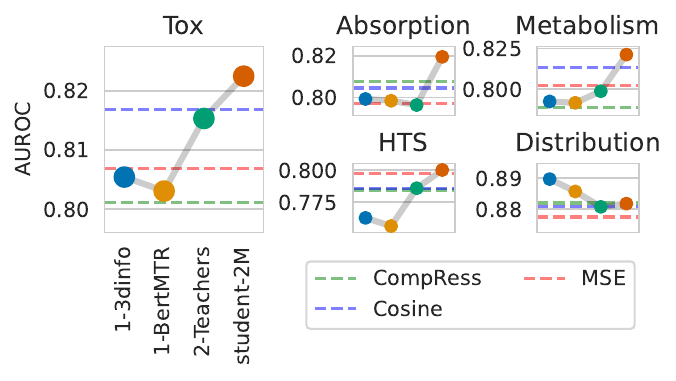}
        \caption{
            \textbf{Number of teachers and classification performances.}
            Performance of students trained with 1 teacher ``1-.", 2 teachers, and 8 teachers (student-2M).
            We compare our NLL distillation method to MSE, Cosine, and CompRess for eight teachers.
        }
        \label{fig:multi_vs_single_reg}
    \end{subfigure}
    ~
    \begin{subfigure}{.4\linewidth}
        \includegraphics[width=\linewidth]{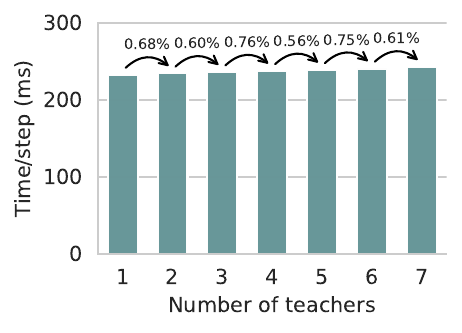}
        \caption{
        \textbf{Computational overhead.}
        Evolution of runtime for a training step as a function of the number of teachers.
        The computational overhead induced by an additional teacher represents less than 1\% of the total runtime on a batch.
        }
        \label{fig:cost-n-teacher}
    \end{subfigure}
    \vspace{-0.4cm}
\end{figure*}

\subsection{Experimental setting}
\label{sec:mol_model_experimental_setting}

\textbf{Teachers and architecture.} 
We use eight teachers trained on different modalities: SMILES (textual representation of the molecular graph)~\citep{ahmad2022chemberta2}, 2D molecular graphs~\citep{You2020GraphCL,xu2021selfsupervised,liu2022pretraining,stark2021_3dinfomax}, and 3D structures~\citep{pmlr-v202-feng23c}. We identify the teachers with ${}^{(t)}$ such as $\text{ChemBERTaMTR}^{(t)} $, and use a 2D-GNN (Graph Isomorphism Network: GIN~\citep{hu2020strategiespretraininggraphneural}) for our student (for more details see~\autoref{sec:appendix_mol_model_model_architecture})\footnote{\url{https://github.com/ills-montreal/mol-distill}}.

\textbf{Evaluation setting.}
We evaluated all models on the ADMET (Absorption, Distribution, Metabolism, Excretion, Toxicity) tasks of the Therapeutic Data Commons platform (TDC)~\citep{Huang2021tdc} and on a high-throughput screening task (HTS), (HIV~\citep{wu_moleculenet_2018}).
We record the test performance over five runs (details on the evaluation procedure in~\autoref{sec:appendix_mol_model_eval_details}).
We trained our models on MOSES, a processed version of the ZINC Clean Leads dataset~\citep{moses}, containing 2 million samples, and on ZINC-250k~\citep{irwin_zinc_2005}, consisting of 250,000 samples.
The performances of the model trained on 250k samples can be found in~\autoref{sec:appendix_mol_model_model_architecture}.
Both are public datasets of commercially available compounds designed to be used in various therapeutic projects. 
\subsection{Results}
\label{sec:mol_model_results}
\textbf{Overall performance.}
We compare the performance of the student model with the teachers and other baseline embedders on the different tasks.
The results (average rank) for each task are presented in~\autoref{fig:hmp_rankings}. 
Our student model achieves the best performance on both the regression and classification tasks, delivering the most accurate predictions across a majority of tasks.
This suggests that our method generates informative representations, providing high-quality molecular descriptors.

\textbf{Single teacher vs. multi-teachers.}
To assess the impact of training a student with multiple teachers, we trained students to distill the knowledge of a single teacher and two teachers, and compared the results to those of our student trained with eight teachers.
We selected two of the best-performing baselines as teachers: ChemBERTaMTR-77M~\citep{ahmad2022chemberta2} and 3D-infomax~\citep{stark2021_3dinfomax}. We then trained student models on the 2M-molecules dataset.
\autoref{fig:multi_vs_single_reg} displays the performances of each of these student models on the regression tasks.
Training with multiple teachers consistently outperforms training with a single teacher, except on the Blood-Brain Barrier (BBB) task (the only Distribution classification task), which is also one of the tasks our model struggles the most with.
For the BBB benchmark, we noticed it is one of the datasets where all results are among the most tightly packed (variations within 1.45 times the average standard deviation of the results), and whose data distribution differs the most from the training set, which could explain the slightly lower average performance of the 8-teacher student compared to the 1 or 2-teacher students.
Overall, using multiple teachers significantly improves performance, with the best performance achieved when training with all eight teachers (additional results are available in~\autoref{sec:appendix_mol_single_teach}).

\textbf{Comparison to baselines.}
\autoref{fig:multi_vs_single_reg} also compares the performance of our NLL distillation method to MSE, cosine, and CompRess distillation for eight teachers.
Overall, in the evaluation of classification tasks, our NLL distillation method outperformed the Cosine and MSE distillation methods.
This observation goes beyond the results of classification tasks, as we also observed that the NLL distillation method consistently outperforms the other two methods on all evaluated task categories (see~\autoref{sec:appendix_mol_model_results_TDC} for more details).

\textbf{Computational complexity.} \label{sec:mol:time_complexity}
 Training our molecular embedders on the largest dataset (2 M molecules) takes approximately 50 hours on 6 A6000 GPUs. We evaluated the computational overhead induced by the multi-teacher setting in~\autoref{fig:cost-n-teacher}. The runtime of a training step increases linearly with the number of teachers: $+1.57 ms$ per teacher, representing less than $ 1\%$ of the total runtime.

\section{Image Embedders}
For our final modality, vision, we aim to assess whether our method can deliver competitive performance compared to other baseline models (teachers, and MSE, Cosine, and CompRess student), especially on fine-grained vision classification tasks. In the following subsections, we outline the experimental setup used to investigate these questions and present the results. Additional details, including hyperparameter tuning and the augmentations applied, can be found in ~\autoref{sec:appendix_vision_det}.

\subsection{Experimental setting}
\textbf{Teachers and evaluations.} 
Given the increasing use of Vision Transformers, we used large transformer models (Swin \citep{liu2021swin}, DINOv2 \citep{oquab2023dinov2}, ViT \citep{dosovitskiy2021imageworth16x16words}, and BEiT \citep{bao2022beitbertpretrainingimage}, with around 87 million parameters)  as teachers, and selected a smaller Vision Transformer, PVTv2 \citep{wang2022pvt}, with 3.7 million parameters, as the student. We also use some CNN based modes with different sizes as baselines to have a more comprehensive comparison of our student's representation abilities (refer to ~\autoref{sec:appendix_vision_models} for more details).

\textbf{Training set.} We include fine-grained datasets such as DTD \citep{cimpoi14describing}, FGVCAircraft \citep{maji13fine-grained}, and CUB \citep{WelinderEtal2010}, alongside CIFAR10 \citep{krizhevsky2009learning}, SVHN \citep{netzer2011reading}, STL10 \citep{coates2011analysis} for the vision experiment. These allows us to assess the performance of our approach on a variety of challenging and detailed classification tasks. Refer to ~\autoref{sec:appendix_vision_training_set} for details of the datasets meta-data \footnote{\url{https://github.com/ills-montreal/vision-distill/}}.

\subsection{Results on Vision Transformer}

\begin{figure}
\begin{center}
\centerline{\includegraphics[
width=\columnwidth,
trim = 0.2cm 0 0.2cm 0, clip
]{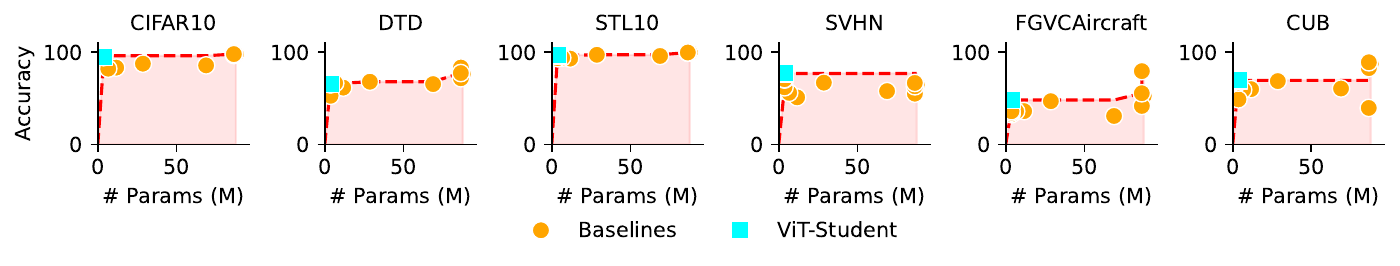}}
\caption{\textbf{Pareto frontier of vision models.} The figure compares the performance of student model distilled using our method (named ViT-Student shown with color blue) with baselines (shown in yellow) across various datasets. The distilled student consistently lies on the Pareto frontier. }%
\vspace{-0.8cm}
\label{fig:vision_pareto}
\end{center}
\end{figure}
To further evaluate our method, we conducted experiments using Vision Transformer (ViT) teachers. As shown in ~\autoref{fig:vision_pareto}, the distilled student model trained with our approach consistently lies on the Pareto frontier, for each task, showing a superior trade-off between accuracy and model size. Notably, our distilled student achieves the best performance among other distillation methods and other baseline models within its respective size categories, with results comparable to large ViT teachers (20$\times$ more parameters). This demonstrates our method's ability to effectively transfer knowledge from large, complex teacher models to smaller, more efficient student models, while maintaining comparable performance. Additional results in ~\autoref{sec:appendix_vision_comp_results} show that our method generalizes well to unseen vision datasets, improving other distillation baselines, and effectively integrates diverse task-specific teachers without performance conflicts, confirming its robustness across domains.

\section{Limitations}

Our method focuses on training student embedding models for diverse, unknown tasks; for single, pre-defined tasks, task-specific distillation may be more effective. As with any distillation approach—especially multi-teacher distillation—there is an overhead, either computational (if teacher embeddings are generated on-the-fly) or memory-intensive (if precomputed). We mitigate this by precomputing and storing embeddings, requiring approximately 100GB of disk space for our largest text-based teacher.
The quality of our student embeddings depends on the relevance of the teachers to the downstream tasks. While task-specific teachers provide limited benefits outside their domain, they do not degrade performance when combined with task-relevant teachers (\autoref{sec:appendix_vision_comp_results}).
Our optimization metric maximizes mutual information between student and teachers but does not explicitly structure the embedding space, potentially limiting performance in tasks like clustering. For textual embeddings, we observe significant gains in classification (where embeddings train a small classifier) but more modest improvements in clustering and STS tasks, which rely on embedding dot products for similarity assessment (\autoref{sec:appendix_nlp_similarity_clustering}).

\section{Conclusions and Future Work}
\label{sec:conclusion}

We proposed a theoretically grounded task-agnostic distillation mechanism that leverages interval estimation through Gaussian kernels in high dimensions to distill a more informative representation from multiple teachers to a single student. We demonstrated that our objective serves as a proxy for maximizing the mutual information and reconstructive capacity of the student model in relation to the teachers. We experimentally validated that our method is more efficient than point estimation-based multi-teacher feature distillation methods such as MSE or cosine-based distillation mechanisms. We demonstrated the superior performance of our method compared to others across three different modalities and numerous downstream tasks. In future work, we aim to extend this distillation approach to cross-modal distillation, enhancing the model's capabilities by leveraging task-agnostic cross-modal information.

\clearpage

\clearpage
\bibliography{biblio}

\begin{thebibliography}{122}
\providecommand{\natexlab}[1]{#1}
\providecommand{\url}[1]{\texttt{#1}}
\expandafter\ifx\csname urlstyle\endcsname\relax
  \providecommand{\doi}[1]{doi: #1}\else
  \providecommand{\doi}{doi: \begingroup \urlstyle{rm}\Url}\fi

\bibitem[Abbasi~Koohpayegani et~al.(2020)Abbasi~Koohpayegani, Tejankar, and Pirsiavash]{NEURIPS2020_975a1c8b}
Abbasi~Koohpayegani, S., Tejankar, A., and Pirsiavash, H.
\newblock Compress: Self-supervised learning by compressing representations.
\newblock In Larochelle, H., Ranzato, M., Hadsell, R., Balcan, M., and Lin, H. (eds.), \emph{Advances in Neural Information Processing Systems}, volume~33, pp.\  12980--12992. Curran Associates, Inc., 2020.
\newblock URL \url{https://proceedings.neurips.cc/paper_files/paper/2020/file/975a1c8b9aee1c48d32e13ec30be7905-Paper.pdf}.

\bibitem[Ahmad et~al.(2022)Ahmad, Simon, Chithrananda, Grand, and Ramsundar]{ahmad2022chemberta2}
Ahmad, W., Simon, E., Chithrananda, S., Grand, G., and Ramsundar, B.
\newblock Chemberta-2: Towards chemical foundation models, 2022.

\bibitem[Allen-Zhu \& Li(2020)Allen-Zhu and Li]{allen2020towards}
Allen-Zhu, Z. and Li, Y.
\newblock Towards understanding ensemble, knowledge distillation and self-distillation in deep learning.
\newblock \emph{arXiv preprint arXiv:2012.09816}, 2020.

\bibitem[Aslam et~al.(2023)Aslam, Zeeshan, Pedersoli, Koerich, Bacon, and Granger]{aslam2023privileged}
Aslam, M.~H., Zeeshan, M.~O., Pedersoli, M., Koerich, A.~L., Bacon, S., and Granger, E.
\newblock Privileged knowledge distillation for dimensional emotion recognition in the wild.
\newblock In \emph{Proceedings of the IEEE/CVF conference on computer vision and pattern recognition}, pp.\  3338--3347, 2023.

\bibitem[Aslam et~al.(2024)Aslam, Pedersoli, Koerich, and Granger]{aslam2024multiteacherprivilegedknowledge}
Aslam, M.~H., Pedersoli, M., Koerich, A.~L., and Granger, E.
\newblock Multi teacher privileged knowledge distillation for multimodal expression recognition, 2024.
\newblock URL \url{https://arxiv.org/abs/2408.09035}.

\bibitem[Axelrod \& G{\'o}mez-Bombarelli(2022)Axelrod and G{\'o}mez-Bombarelli]{axelrod2022geom}
Axelrod, S. and G{\'o}mez-Bombarelli, R.
\newblock Geom, energy-annotated molecular conformations for property prediction and molecular generation.
\newblock \emph{Scientific Data}, 9\penalty0 (1):\penalty0 185, 2022.
\newblock \doi{10.1038/s41597-022-01288-4}.
\newblock URL \url{https://doi.org/10.1038/s41597-022-01288-4}.

\bibitem[Bao et~al.(2022)Bao, Dong, Piao, and Wei]{bao2022beitbertpretrainingimage}
Bao, H., Dong, L., Piao, S., and Wei, F.
\newblock Beit: Bert pre-training of image transformers, 2022.
\newblock URL \url{https://arxiv.org/abs/2106.08254}.

\bibitem[Bellemare et~al.(2017)Bellemare, Dabney, and Munos]{bellemare2017distributionalperspectivereinforcementlearning}
Bellemare, M.~G., Dabney, W., and Munos, R.
\newblock A distributional perspective on reinforcement learning, 2017.
\newblock URL \url{https://arxiv.org/abs/1707.06887}.

\bibitem[Bhalla et~al.(2024)Bhalla, Oesterling, Srinivas, Calmon, and Lakkaraju]{bhalla2024interpreting}
Bhalla, U., Oesterling, A., Srinivas, S., Calmon, F.~P., and Lakkaraju, H.
\newblock Interpreting clip with sparse linear concept embeddings (splice), 2024.

\bibitem[Borza et~al.(2022)Borza, Darabant, Ileni, and Marinescu]{att2022}
Borza, D.-L., Darabant, A., Ileni, T., and Marinescu, A.-I.
\newblock Effective online knowledge distillation via attention-based model ensembling.
\newblock \emph{Mathematics}, 10:\penalty0 4285, 11 2022.
\newblock \doi{10.3390/math10224285}.

\bibitem[Bossard et~al.(2014)Bossard, Guillaumin, and Van~Gool]{10.1007/978-3-319-10599-4_29}
Bossard, L., Guillaumin, M., and Van~Gool, L.
\newblock Food-101 -- mining discriminative components with random forests.
\newblock In Fleet, D., Pajdla, T., Schiele, B., and Tuytelaars, T. (eds.), \emph{Computer Vision -- ECCV 2014}, pp.\  446--461, Cham, 2014. Springer International Publishing.
\newblock ISBN 978-3-319-10599-4.

\bibitem[Bowman et~al.(2015)Bowman, Angeli, Potts, and Manning]{bowman-etal-2015-large}
Bowman, S.~R., Angeli, G., Potts, C., and Manning, C.~D.
\newblock A large annotated corpus for learning natural language inference.
\newblock In M{\`a}rquez, L., Callison-Burch, C., and Su, J. (eds.), \emph{Proceedings of the 2015 Conference on Empirical Methods in Natural Language Processing}, pp.\  632--642, Lisbon, Portugal, September 2015. Association for Computational Linguistics.
\newblock \doi{10.18653/v1/D15-1075}.
\newblock URL \url{https://aclanthology.org/D15-1075}.

\bibitem[Brody et~al.(2022)Brody, Alon, and Yahav]{brody2022attentivegraphattentionnetworks}
Brody, S., Alon, U., and Yahav, E.
\newblock How attentive are graph attention networks?, 2022.
\newblock URL \url{https://arxiv.org/abs/2105.14491}.

\bibitem[Carion et~al.(2020)Carion, Massa, Synnaeve, Usunier, Kirillov, and Zagoruyko]{carion2020end}
Carion, N., Massa, F., Synnaeve, G., Usunier, N., Kirillov, A., and Zagoruyko, S.
\newblock End-to-end object detection with transformers.
\newblock In \emph{European conference on computer vision}, pp.\  213--229. Springer, 2020.

\bibitem[Cimpoi et~al.(2014)Cimpoi, Maji, Kokkinos, Mohamed, , and Vedaldi]{cimpoi14describing}
Cimpoi, M., Maji, S., Kokkinos, I., Mohamed, S., , and Vedaldi, A.
\newblock Describing textures in the wild.
\newblock In \emph{Proceedings of the {IEEE} Conf. on Computer Vision and Pattern Recognition ({CVPR})}, 2014.

\bibitem[Coates et~al.(2011)Coates, Ng, and Lee]{coates2011analysis}
Coates, A., Ng, A., and Lee, H.
\newblock An analysis of single-layer networks in unsupervised feature learning.
\newblock In \emph{Proceedings of the fourteenth international conference on artificial intelligence and statistics}, pp.\  215--223. JMLR Workshop and Conference Proceedings, 2011.

\bibitem[Cohan et~al.(2020)Cohan, Feldman, Beltagy, Downey, and Weld]{specter2020cohan}
Cohan, A., Feldman, S., Beltagy, I., Downey, D., and Weld, D.~S.
\newblock {SPECTER: Document-level Representation Learning using Citation-informed Transformers}.
\newblock In \emph{ACL}, 2020.

\bibitem[Darrin et~al.(2024)Darrin, Formont, Cheung, and Piantanida]{darrin2024textttcosmicmutualinformationtaskagnostic}
Darrin, M., Formont, P., Cheung, J. C.~K., and Piantanida, P.
\newblock $\texttt{COSMIC}$: Mutual information for task-agnostic summarization evaluation, 2024.
\newblock URL \url{https://arxiv.org/abs/2402.19457}.

\bibitem[Dietterich(2000)]{dietterich-iwmcs-00}
Dietterich, T.~G.
\newblock Ensemble methods in machine learning.
\newblock In \emph{International workshop on multiple classifier systems}, pp.\  1--15. Springer, 2000.

\bibitem[Dong et~al.(2021)Dong, Li, Shen, and Qiu]{dong-etal-2021-hrkd}
Dong, C., Li, Y., Shen, Y., and Qiu, M.
\newblock {HRKD}: Hierarchical relational knowledge distillation for cross-domain language model compression.
\newblock In Moens, M.-F., Huang, X., Specia, L., and Yih, S. W.-t. (eds.), \emph{Proceedings of the 2021 Conference on Empirical Methods in Natural Language Processing}, pp.\  3126--3136, Online and Punta Cana, Dominican Republic, November 2021. Association for Computational Linguistics.
\newblock \doi{10.18653/v1/2021.emnlp-main.250}.
\newblock URL \url{https://aclanthology.org/2021.emnlp-main.250}.

\bibitem[Dong et~al.(2024)Dong, Miller, Lei, and Ward]{dong2024cluster}
Dong, Y., Miller, K., Lei, Q., and Ward, R.
\newblock Cluster-aware semi-supervised learning: relational knowledge distillation provably learns clustering.
\newblock \emph{Advances in Neural Information Processing Systems}, 36, 2024.

\bibitem[Dosovitskiy et~al.(2021)Dosovitskiy, Beyer, Kolesnikov, Weissenborn, Zhai, Unterthiner, Dehghani, Minderer, Heigold, Gelly, Uszkoreit, and Houlsby]{dosovitskiy2021imageworth16x16words}
Dosovitskiy, A., Beyer, L., Kolesnikov, A., Weissenborn, D., Zhai, X., Unterthiner, T., Dehghani, M., Minderer, M., Heigold, G., Gelly, S., Uszkoreit, J., and Houlsby, N.
\newblock An image is worth 16x16 words: Transformers for image recognition at scale, 2021.
\newblock URL \url{https://arxiv.org/abs/2010.11929}.

\bibitem[Duval et~al.(2023)Duval, Misra, and Ballas]{duval2023simplerecipecompetitivelowcompute}
Duval, Q., Misra, I., and Ballas, N.
\newblock A simple recipe for competitive low-compute self supervised vision models, 2023.
\newblock URL \url{https://arxiv.org/abs/2301.09451}.

\bibitem[Dvornik et~al.(2019)Dvornik, Schmid, and Mairal]{dvornik2019diversity}
Dvornik, N., Schmid, C., and Mairal, J.
\newblock Diversity with cooperation: Ensemble methods for few-shot classification.
\newblock In \emph{Proceedings of the IEEE/CVF international conference on computer vision}, pp.\  3723--3731, 2019.

\bibitem[Fang et~al.(2021)Fang, Wang, Wang, Zhang, Yang, and Liu]{fang2021seedselfsuperviseddistillationvisual}
Fang, Z., Wang, J., Wang, L., Zhang, L., Yang, Y., and Liu, Z.
\newblock Seed: Self-supervised distillation for visual representation, 2021.
\newblock URL \url{https://arxiv.org/abs/2101.04731}.

\bibitem[Farebrother et~al.(2024)Farebrother, Orbay, Vuong, Taïga, Chebotar, Xiao, Irpan, Levine, Castro, Faust, Kumar, and Agarwal]{farebrother2024stopregressingtrainingvalue}
Farebrother, J., Orbay, J., Vuong, Q., Taïga, A.~A., Chebotar, Y., Xiao, T., Irpan, A., Levine, S., Castro, P.~S., Faust, A., Kumar, A., and Agarwal, R.
\newblock Stop regressing: Training value functions via classification for scalable deep rl, 2024.
\newblock URL \url{https://arxiv.org/abs/2403.03950}.

\bibitem[Feng et~al.(2023)Feng, Ni, Lan, Ma, and Ma]{pmlr-v202-feng23c}
Feng, S., Ni, Y., Lan, Y., Ma, Z.-M., and Ma, W.-Y.
\newblock Fractional denoising for 3{D} molecular pre-training.
\newblock In \emph{Proceedings of the 40th International Conference on Machine Learning}, volume 202 of \emph{Proceedings of Machine Learning Research}, pp.\  9938--9961. PMLR, 23--29 Jul 2023.
\newblock URL \url{https://proceedings.mlr.press/v202/feng23c.html}.

\bibitem[Feng et~al.(2024)Feng, Ni, Li, Huang, Ma, Ma, and Lan]{feng2024unicornunifiedcontrastivelearning}
Feng, S., Ni, Y., Li, M., Huang, Y., Ma, Z.-M., Ma, W.-Y., and Lan, Y.
\newblock Unicorn: A unified contrastive learning approach for multi-view molecular representation learning, 2024.
\newblock URL \url{https://arxiv.org/abs/2405.10343}.

\bibitem[Frosst \& Hinton(2017)Frosst and Hinton]{frosst2017distilling}
Frosst, N. and Hinton, G.
\newblock Distilling a neural network into a soft decision tree, 2017.

\bibitem[Gao et~al.(2022)Gao, Zhuang, Lin, Cheng, Sun, Li, and Shen]{gao2022discoremedyselfsupervisedlearning}
Gao, Y., Zhuang, J.-X., Lin, S., Cheng, H., Sun, X., Li, K., and Shen, C.
\newblock Disco: Remedy self-supervised learning on lightweight models with distilled contrastive learning, 2022.
\newblock URL \url{https://arxiv.org/abs/2104.09124}.

\bibitem[Gou et~al.(2021)Gou, Yu, Maybank, and Tao]{Gou_2021}
Gou, J., Yu, B., Maybank, S.~J., and Tao, D.
\newblock Knowledge distillation: A survey.
\newblock \emph{International Journal of Computer Vision}, 129\penalty0 (6):\penalty0 1789–1819, March 2021.
\newblock ISSN 1573-1405.
\newblock \doi{10.1007/s11263-021-01453-z}.
\newblock URL \url{http://dx.doi.org/10.1007/s11263-021-01453-z}.

\bibitem[Hamilton et~al.(2018)Hamilton, Ying, and Leskovec]{hamilton2018inductiverepresentationlearninglarge}
Hamilton, W.~L., Ying, R., and Leskovec, J.
\newblock Inductive representation learning on large graphs, 2018.
\newblock URL \url{https://arxiv.org/abs/1706.02216}.

\bibitem[He et~al.(2016)He, Zhang, Ren, and Sun]{he2016deep}
He, K., Zhang, X., Ren, S., and Sun, J.
\newblock Deep residual learning for image recognition.
\newblock In \emph{Proceedings of the IEEE conference on computer vision and pattern recognition}, pp.\  770--778, 2016.

\bibitem[Herbold(2020)]{Herbold2020}
Herbold, S.
\newblock Autorank: A python package for automated ranking of classifiers.
\newblock \emph{Journal of Open Source Software}, 5\penalty0 (48):\penalty0 2173, 2020.
\newblock \doi{10.21105/joss.02173}.
\newblock URL \url{https://doi.org/10.21105/joss.02173}.

\bibitem[Hinton et~al.(2015)Hinton, Vinyals, and Dean]{hinton2015distilling}
Hinton, G., Vinyals, O., and Dean, J.
\newblock Distilling the knowledge in a neural network.
\newblock \emph{arXiv preprint arXiv:1503.02531}, 2015.

\bibitem[Hu et~al.(2020)Hu, Liu, Gomes, Zitnik, Liang, Pande, and Leskovec]{hu2020strategiespretraininggraphneural}
Hu, W., Liu, B., Gomes, J., Zitnik, M., Liang, P., Pande, V., and Leskovec, J.
\newblock Strategies for pre-training graph neural networks, 2020.
\newblock URL \url{https://arxiv.org/abs/1905.12265}.

\bibitem[Hu et~al.(2021)Hu, Fey, Ren, Nakata, Dong, and Leskovec]{hu2021ogblsc}
Hu, W., Fey, M., Ren, H., Nakata, M., Dong, Y., and Leskovec, J.
\newblock Ogb-lsc: A large-scale challenge for machine learning on graphs, 2021.

\bibitem[Huang et~al.(2017)Huang, Liu, Van Der~Maaten, and Weinberger]{huang2017densely}
Huang, G., Liu, Z., Van Der~Maaten, L., and Weinberger, K.~Q.
\newblock Densely connected convolutional networks.
\newblock In \emph{Proceedings of the IEEE conference on computer vision and pattern recognition}, pp.\  4700--4708, 2017.

\bibitem[Huang et~al.(2021)Huang, Fu, Gao, Zhao, Roohani, Leskovec, Coley, Xiao, Sun, and Zitnik]{Huang2021tdc}
Huang, K., Fu, T., Gao, W., Zhao, Y., Roohani, Y., Leskovec, J., Coley, C.~W., Xiao, C., Sun, J., and Zitnik, M.
\newblock Therapeutics data commons: Machine learning datasets and tasks for drug discovery and development.
\newblock \emph{Proceedings of Neural Information Processing Systems, NeurIPS Datasets and Benchmarks}, 2021.

\bibitem[Iandola et~al.(2016)Iandola, Han, Moskewicz, Ashraf, Dally, and Keutzer]{iandola2016squeezenetalexnetlevelaccuracy50x}
Iandola, F.~N., Han, S., Moskewicz, M.~W., Ashraf, K., Dally, W.~J., and Keutzer, K.
\newblock Squeezenet: Alexnet-level accuracy with 50x fewer parameters and <0.5mb model size, 2016.
\newblock URL \url{https://arxiv.org/abs/1602.07360}.

\bibitem[Irwin \& Shoichet(2005)Irwin and Shoichet]{irwin_zinc_2005}
Irwin, J.~J. and Shoichet, B.~K.
\newblock {ZINC} – {A} {Free} {Database} of {Commercially} {Available} {Compounds} for {Virtual} {Screening}.
\newblock \emph{Journal of chemical information and modeling}, 45\penalty0 (1):\penalty0 177--182, 2005.
\newblock ISSN 1549-9596.
\newblock \doi{10.1021/ci049714}.
\newblock URL \url{https://www.ncbi.nlm.nih.gov/pmc/articles/PMC1360656/}.

\bibitem[Isert et~al.(2021)Isert, Atz, Jiménez-Luna, and Schneider]{isert2021qmugs}
Isert, C., Atz, K., Jiménez-Luna, J., and Schneider, G.
\newblock Qmugs: Quantum mechanical properties of drug-like molecules, 2021.

\bibitem[Jiang et~al.(2023)Jiang, Sablayrolles, Mensch, Bamford, Chaplot, de~las Casas, Bressand, Lengyel, Lample, Saulnier, Lavaud, Lachaux, Stock, Scao, Lavril, Wang, Lacroix, and Sayed]{jiang2023mistral}
Jiang, A.~Q., Sablayrolles, A., Mensch, A., Bamford, C., Chaplot, D.~S., de~las Casas, D., Bressand, F., Lengyel, G., Lample, G., Saulnier, L., Lavaud, L.~R., Lachaux, M.-A., Stock, P., Scao, T.~L., Lavril, T., Wang, T., Lacroix, T., and Sayed, W.~E.
\newblock Mistral 7b, 2023.

\bibitem[Kim et~al.(2022)Kim, Chen, Cheng, Gindulyte, He, He, Li, Shoemaker, Thiessen, Yu, Zaslavsky, Zhang, and Bolton]{pubchem}
Kim, S., Chen, J., Cheng, T., Gindulyte, A., He, J., He, S., Li, Q., Shoemaker, B.~A., Thiessen, P.~A., Yu, B., Zaslavsky, L., Zhang, J., and Bolton, E.~E.
\newblock {PubChem 2023 update}.
\newblock \emph{Nucleic Acids Research}, 51\penalty0 (D1):\penalty0 D1373--D1380, 10 2022.
\newblock ISSN 0305-1048.
\newblock \doi{10.1093/nar/gkac956}.
\newblock URL \url{https://doi.org/10.1093/nar/gkac956}.

\bibitem[Krause et~al.(2013)Krause, Stark, Deng, and Fei-Fei]{6755945}
Krause, J., Stark, M., Deng, J., and Fei-Fei, L.
\newblock 3d object representations for fine-grained categorization.
\newblock In \emph{2013 IEEE International Conference on Computer Vision Workshops}, pp.\  554--561, 2013.
\newblock \doi{10.1109/ICCVW.2013.77}.

\bibitem[Krizhevsky et~al.(2009)Krizhevsky, Hinton, et~al.]{krizhevsky2009learning}
Krizhevsky, A., Hinton, G., et~al.
\newblock Learning multiple layers of features from tiny images, 2009.

\bibitem[Kubota et~al.(2024)Kubota, Haraguchi, and Uchida]{kubota2024impressionclip}
Kubota, Y., Haraguchi, D., and Uchida, S.
\newblock Impression-clip: Contrastive shape-impression embedding for fonts, 2024.

\bibitem[Li et~al.(2024)Li, Yang, Cheng, Liu, and Hu]{Li2024}
Li, S., Yang, X., Cheng, G., Liu, W., and Hu, H.
\newblock Sa-mdrad: sample-adaptive multi-teacher dynamic rectification adversarial distillation.
\newblock \emph{Multimedia Systems}, 30\penalty0 (4), July 2024.
\newblock ISSN 1432-1882.
\newblock \doi{10.1007/s00530-024-01416-7}.
\newblock URL \url{http://dx.doi.org/10.1007/s00530-024-01416-7}.

\bibitem[Li \& Li(2023)Li and Li]{li2023angleoptimized}
Li, X. and Li, J.
\newblock Angle-optimized text embeddings, 2023.

\bibitem[Liang et~al.(2023)Liang, Jiang, Li, Tang, Yin, and Zhao]{lianghomodistil}
Liang, C., Jiang, H., Li, Z., Tang, X., Yin, B., and Zhao, T.
\newblock Homodistil: Homotopic task-agnostic distillation of pre-trained transformers.
\newblock In \emph{The Eleventh International Conference on Learning Representations}, 2023.

\bibitem[Liu et~al.(2021{\natexlab{a}})Liu, Huang, Lin, Xie, Wang, Chang, and Liang]{liu2021exploring}
Liu, L., Huang, Q., Lin, S., Xie, H., Wang, B., Chang, X., and Liang, X.
\newblock Exploring inter-channel correlation for diversity-preserved knowledge distillation.
\newblock In \emph{Proceedings of the IEEE/CVF international conference on computer vision}, pp.\  8271--8280, 2021{\natexlab{a}}.

\bibitem[Liu et~al.(2022)Liu, Wang, Liu, Lasenby, Guo, and Tang]{liu2022pretraining}
Liu, S., Wang, H., Liu, W., Lasenby, J., Guo, H., and Tang, J.
\newblock Pre-training molecular graph representation with 3d geometry.
\newblock In \emph{International Conference on Learning Representations}, 2022.
\newblock URL \url{https://openreview.net/forum?id=xQUe1pOKPam}.

\bibitem[Liu et~al.(2023)Liu, Chen, Liu, Feng, Sun, Tian, and Wu]{liu2023ernie}
Liu, W., Chen, X., Liu, J., Feng, S., Sun, Y., Tian, H., and Wu, H.
\newblock Ernie 3.0 tiny: Frustratingly simple method to improve task-agnostic distillation generalization.
\newblock \emph{arXiv preprint arXiv:2301.03416}, 2023.

\bibitem[Liu et~al.(2024)Liu, Blodgett, Cheung, Liao, Olteanu, and Xiao]{ecbd-bench}
Liu, Y.~L., Blodgett, S.~L., Cheung, J., Liao, Q.~V., Olteanu, A., and Xiao, Z.
\newblock {ECBD}: Evidence-centered benchmark design for {NLP}.
\newblock In Ku, L.-W., Martins, A., and Srikumar, V. (eds.), \emph{Proceedings of the 62nd Annual Meeting of the Association for Computational Linguistics (Volume 1: Long Papers)}, pp.\  16349--16365, Bangkok, Thailand, August 2024. Association for Computational Linguistics.
\newblock \doi{10.18653/v1/2024.acl-long.861}.
\newblock URL \url{https://aclanthology.org/2024.acl-long.861}.

\bibitem[Liu et~al.(2021{\natexlab{b}})Liu, Lin, Cao, Hu, Wei, Zhang, Lin, and Guo]{liu2021swin}
Liu, Z., Lin, Y., Cao, Y., Hu, H., Wei, Y., Zhang, Z., Lin, S., and Guo, B.
\newblock Swin transformer: Hierarchical vision transformer using shifted windows.
\newblock In \emph{Proceedings of the IEEE/CVF international conference on computer vision}, pp.\  10012--10022, 2021{\natexlab{b}}.

\bibitem[Ma et~al.(2024{\natexlab{a}})Ma, Zhang, Cao, Li, and Gao]{MA2024123892}
Ma, D., Zhang, K., Cao, Q., Li, J., and Gao, X.
\newblock Coordinate attention guided dual-teacher adaptive knowledge distillation for image classification.
\newblock \emph{Expert Systems with Applications}, 250:\penalty0 123892, 2024{\natexlab{a}}.
\newblock ISSN 0957-4174.
\newblock \doi{https://doi.org/10.1016/j.eswa.2024.123892}.
\newblock URL \url{https://www.sciencedirect.com/science/article/pii/S0957417424007589}.

\bibitem[Ma et~al.(2018)Ma, Zhang, Zheng, and Sun]{ma2018shufflenet}
Ma, N., Zhang, X., Zheng, H.-T., and Sun, J.
\newblock Shufflenet v2: Practical guidelines for efficient cnn architecture design.
\newblock In \emph{Proceedings of the European conference on computer vision (ECCV)}, pp.\  116--131, 2018.

\bibitem[Ma et~al.(2024{\natexlab{b}})Ma, Dong, Ji, Liu, Zhang, Wang, He, Qian, Zhang, and Yang]{Ma2024}
Ma, Z., Dong, J., Ji, S., Liu, Z., Zhang, X., Wang, Z., He, S., Qian, F., Zhang, X., and Yang, L.
\newblock Let all be whitened: Multi-teacher distillation for efficient visual retrieval.
\newblock \emph{Proceedings of the AAAI Conference on Artificial Intelligence}, 38\penalty0 (5):\penalty0 4126–4135, March 2024{\natexlab{b}}.
\newblock ISSN 2159-5399.
\newblock \doi{10.1609/aaai.v38i5.28207}.
\newblock URL \url{http://dx.doi.org/10.1609/aaai.v38i5.28207}.

\bibitem[Maas et~al.(2011)Maas, Daly, Pham, Huang, Ng, and Potts]{maas-EtAl:2011:ACL-HLT2011}
Maas, A.~L., Daly, R.~E., Pham, P.~T., Huang, D., Ng, A.~Y., and Potts, C.
\newblock Learning word vectors for sentiment analysis.
\newblock In \emph{Proceedings of the 49th Annual Meeting of the Association for Computational Linguistics: Human Language Technologies}, pp.\  142--150, Portland, Oregon, USA, June 2011. Association for Computational Linguistics.
\newblock URL \url{http://www.aclweb.org/anthology/P11-1015}.

\bibitem[Maji et~al.(2013)Maji, Rahtu, Kannala, Blaschko, and Vedaldi]{maji13fine-grained}
Maji, S., Rahtu, E., Kannala, J., Blaschko, M., and Vedaldi, A.
\newblock Fine-grained visual classification of aircraft.
\newblock \emph{arXiv preprint arXiv:1306.5151}, 2013.

\bibitem[McAuley \& Leskovec(2013)McAuley and Leskovec]{10.1145/2507157.2507163}
McAuley, J. and Leskovec, J.
\newblock Hidden factors and hidden topics: understanding rating dimensions with review text.
\newblock In \emph{Proceedings of the 7th ACM Conference on Recommender Systems}, RecSys '13, pp.\  165–172, New York, NY, USA, 2013. Association for Computing Machinery.
\newblock ISBN 9781450324090.
\newblock \doi{10.1145/2507157.2507163}.
\newblock URL \url{https://doi.org/10.1145/2507157.2507163}.

\bibitem[Merrick(2024)]{merrick2024embeddingclusteringdataimprove}
Merrick, L.
\newblock Embedding and clustering your data can improve contrastive pretraining, 2024.
\newblock URL \url{https://arxiv.org/abs/2407.18887}.

\bibitem[Merrick et~al.(2024)Merrick, Xu, Nuti, and Campos]{merrick2024arcticembedscalableefficientaccurate}
Merrick, L., Xu, D., Nuti, G., and Campos, D.
\newblock Arctic-embed: Scalable, efficient, and accurate text embedding models, 2024.
\newblock URL \url{https://arxiv.org/abs/2405.05374}.

\bibitem[Mobley \& Guthrie(2014)Mobley and Guthrie]{mobley_freesolv_2014}
Mobley, D.~L. and Guthrie, J.~P.
\newblock {FreeSolv}: a database of experimental and calculated hydration free energies, with input files.
\newblock \emph{Journal of Computer-Aided Molecular Design}, 28\penalty0 (7):\penalty0 711--720, July 2014.
\newblock ISSN 1573-4951.
\newblock \doi{10.1007/s10822-014-9747-x}.
\newblock URL \url{https://doi.org/10.1007/s10822-014-9747-x}.

\bibitem[Morgan(1965)]{morganfp}
Morgan, H.~L.
\newblock The generation of a unique machine description for chemical structures-a technique developed at chemical abstracts service.
\newblock \emph{Journal of Chemical Documentation}, 5\penalty0 (2):\penalty0 107--113, 1965.
\newblock \doi{10.1021/c160017a018}.
\newblock URL \url{https://doi.org/10.1021/c160017a018}.

\bibitem[Morris et~al.(2021)Morris, Ritzert, Fey, Hamilton, Lenssen, Rattan, and Grohe]{morris2021weisfeilerlemanneuralhigherorder}
Morris, C., Ritzert, M., Fey, M., Hamilton, W.~L., Lenssen, J.~E., Rattan, G., and Grohe, M.
\newblock Weisfeiler and leman go neural: Higher-order graph neural networks, 2021.
\newblock URL \url{https://arxiv.org/abs/1810.02244}.

\bibitem[Muennighoff et~al.(2023)Muennighoff, Tazi, Magne, and Reimers]{muennighoff2023mteb}
Muennighoff, N., Tazi, N., Magne, L., and Reimers, N.
\newblock Mteb: Massive text embedding benchmark, 2023.

\bibitem[Muralidharan et~al.(2024)Muralidharan, Sreenivas, Joshi, Chochowski, Patwary, Shoeybi, Catanzaro, Kautz, and Molchanov]{muralidharan2024compact}
Muralidharan, S., Sreenivas, S.~T., Joshi, R., Chochowski, M., Patwary, M., Shoeybi, M., Catanzaro, B., Kautz, J., and Molchanov, P.
\newblock Compact language models via pruning and knowledge distillation.
\newblock \emph{arXiv preprint arXiv:2407.14679}, 2024.

\bibitem[Murphy(2013)]{murphy2013machine}
Murphy, K.~P.
\newblock \emph{Machine learning : a probabilistic perspective}.
\newblock MIT Press, Cambridge, Mass. [u.a.], 2013.
\newblock ISBN 9780262018029 0262018020.
\newblock URL \url{https://www.amazon.com/Machine-Learning-Probabilistic-Perspective-Computation/dp/0262018020/ref=sr_1_2?ie=UTF8&qid=1336857747&sr=8-2}.

\bibitem[Navaneet et~al.(2022)Navaneet, Koohpayegani, Tejankar, and Pirsiavash]{navaneet2022simregregressionsimpleeffective}
Navaneet, K.~L., Koohpayegani, S.~A., Tejankar, A., and Pirsiavash, H.
\newblock Simreg: Regression as a simple yet effective tool for self-supervised knowledge distillation, 2022.
\newblock URL \url{https://arxiv.org/abs/2201.05131}.

\bibitem[Netzer et~al.(2011)Netzer, Wang, Coates, Bissacco, Wu, Ng, et~al.]{netzer2011reading}
Netzer, Y., Wang, T., Coates, A., Bissacco, A., Wu, B., Ng, A.~Y., et~al.
\newblock Reading digits in natural images with unsupervised feature learning.
\newblock In \emph{NIPS workshop on deep learning and unsupervised feature learning}, volume 2011, pp.\ ~4. Granada, 2011.

\bibitem[Ni et~al.(2021)Ni, Ábrego, Constant, Ma, Hall, Cer, and Yang]{ni2021sentencet5}
Ni, J., Ábrego, G.~H., Constant, N., Ma, J., Hall, K.~B., Cer, D., and Yang, Y.
\newblock Sentence-t5: Scalable sentence encoders from pre-trained text-to-text models, 2021.

\bibitem[Noroozi et~al.(2018)Noroozi, Vinjimoor, Favaro, and Pirsiavash]{noroozi2018boosting}
Noroozi, M., Vinjimoor, A., Favaro, P., and Pirsiavash, H.
\newblock Boosting self-supervised learning via knowledge transfer.
\newblock In \emph{Proceedings of the IEEE conference on computer vision and pattern recognition}, pp.\  9359--9367, 2018.

\bibitem[Oquab et~al.(2023)Oquab, Darcet, Moutakanni, Vo, Szafraniec, Khalidov, Fernandez, Haziza, Massa, El-Nouby, et~al.]{oquab2023dinov2}
Oquab, M., Darcet, T., Moutakanni, T., Vo, H., Szafraniec, M., Khalidov, V., Fernandez, P., Haziza, D., Massa, F., El-Nouby, A., et~al.
\newblock Dinov2: Learning robust visual features without supervision.
\newblock \emph{arXiv preprint arXiv:2304.07193}, 2023.

\bibitem[Pan et~al.(2022)Pan, Zhou, and Tian]{pan2022extreme}
Pan, Z., Zhou, X., and Tian, H.
\newblock Extreme generative image compression by learning text embedding from diffusion models.
\newblock \emph{arXiv preprint arXiv:2211.07793}, 2022.

\bibitem[Park et~al.(2019{\natexlab{a}})Park, Kim, Lu, and Cho]{park2019relational}
Park, W., Kim, D., Lu, Y., and Cho, M.
\newblock Relational knowledge distillation.
\newblock In \emph{Proceedings of the IEEE/CVF conference on computer vision and pattern recognition}, pp.\  3967--3976, 2019{\natexlab{a}}.

\bibitem[Park et~al.(2019{\natexlab{b}})Park, Kim, Lu, and Cho]{rkd}
Park, W., Kim, D., Lu, Y., and Cho, M.
\newblock Relational knowledge distillation, 2019{\natexlab{b}}.
\newblock URL \url{https://arxiv.org/abs/1904.05068}.

\bibitem[Parkhi et~al.(2012)Parkhi, Vedaldi, Zisserman, and Jawahar]{6248092}
Parkhi, O.~M., Vedaldi, A., Zisserman, A., and Jawahar, C.~V.
\newblock Cats and dogs.
\newblock In \emph{2012 IEEE Conference on Computer Vision and Pattern Recognition}, pp.\  3498--3505, 2012.
\newblock \doi{10.1109/CVPR.2012.6248092}.

\bibitem[Peng et~al.(2019)Peng, Jin, Liu, Li, Wu, Liu, Zhou, and Zhang]{cc}
Peng, B., Jin, X., Liu, J., Li, D., Wu, Y., Liu, Y., Zhou, S., and Zhang, Z.
\newblock Correlation congruence for knowledge distillation.
\newblock In \emph{Proceedings of the IEEE/CVF International Conference on Computer Vision (ICCV)}, October 2019.

\bibitem[Pichler et~al.(2022)Pichler, Colombo, Boudiaf, Koliander, and Piantanida]{pichler2022differential}
Pichler, G., Colombo, P., Boudiaf, M., Koliander, G., and Piantanida, P.
\newblock A differential entropy estimator for training neural networks, 2022.

\bibitem[Pimentel et~al.(2023)Pimentel, Meister, and Cotterell]{pimentel2023usefulness}
Pimentel, T., Meister, C., and Cotterell, R.
\newblock On the usefulness of embeddings, clusters and strings for text generator evaluation, 2023.

\bibitem[Polykovskiy et~al.(2018)Polykovskiy, Zhebrak, S{\'{a}}nchez{-}Lengeling, Golovanov, Tatanov, Belyaev, Kurbanov, Artamonov, Aladinskiy, Veselov, Kadurin, Nikolenko, Aspuru{-}Guzik, and Zhavoronkov]{moses}
Polykovskiy, D., Zhebrak, A., S{\'{a}}nchez{-}Lengeling, B., Golovanov, S., Tatanov, O., Belyaev, S., Kurbanov, R., Artamonov, A., Aladinskiy, V., Veselov, M., Kadurin, A., Nikolenko, S.~I., Aspuru{-}Guzik, A., and Zhavoronkov, A.
\newblock Molecular sets {(MOSES):} {A} benchmarking platform for molecular generation models.
\newblock \emph{CoRR}, abs/1811.12823, 2018.
\newblock URL \url{http://arxiv.org/abs/1811.12823}.

\bibitem[Qiu et~al.(2024)Qiu, Han, Maddix, Zhang, Wang, and Wilson]{qiu2024transferring}
Qiu, S., Han, B., Maddix, D.~C., Zhang, S., Wang, Y., and Wilson, A.~G.
\newblock Transferring knowledge from large foundation models to small downstream models.
\newblock \emph{arXiv preprint arXiv:2406.07337}, 2024.

\bibitem[Sandler et~al.(2018)Sandler, Howard, Zhu, Zhmoginov, and Chen]{sandler2018mobilenetv2}
Sandler, M., Howard, A., Zhu, M., Zhmoginov, A., and Chen, L.-C.
\newblock Mobilenetv2: Inverted residuals and linear bottlenecks.
\newblock In \emph{Proceedings of the IEEE conference on computer vision and pattern recognition}, pp.\  4510--4520, 2018.

\bibitem[Saravia et~al.(2018)Saravia, Liu, Huang, Wu, and Chen]{saravia-etal-2018-carer}
Saravia, E., Liu, H.-C.~T., Huang, Y.-H., Wu, J., and Chen, Y.-S.
\newblock {CARER}: Contextualized affect representations for emotion recognition.
\newblock In \emph{Proceedings of the 2018 Conference on Empirical Methods in Natural Language Processing}, pp.\  3687--3697, Brussels, Belgium, October-November 2018. Association for Computational Linguistics.
\newblock \doi{10.18653/v1/D18-1404}.
\newblock URL \url{https://www.aclweb.org/anthology/D18-1404}.

\bibitem[Sarkar \& Etemad(2024)Sarkar and Etemad]{sarkar2024xkd}
Sarkar, P. and Etemad, A.
\newblock Xkd: Cross-modal knowledge distillation with domain alignment for video representation learning.
\newblock In \emph{Proceedings of the AAAI Conference on Artificial Intelligence}, volume~38, pp.\  14875--14885, 2024.

\bibitem[Scimeca et~al.(2023)Scimeca, Rubinstein, Teney, Oh, Nicolicioiu, and Bengio]{scimeca2023shortcut}
Scimeca, L., Rubinstein, A., Teney, D., Oh, S.~J., Nicolicioiu, A.~M., and Bengio, Y.
\newblock Shortcut bias mitigation via ensemble diversity using diffusion probabilistic models.
\newblock \emph{arXiv preprint arXiv:2311.16176}, 2023.

\bibitem[Shayegh et~al.(2024)Shayegh, Cao, Zhu, Cheung, and Mou]{shayegh2024ensemble}
Shayegh, B., Cao, Y., Zhu, X., Cheung, J.~C., and Mou, L.
\newblock Ensemble distillation for unsupervised constituency parsing.
\newblock In \emph{International Conference on Learning Representations (ICLR)}, 2024.

\bibitem[Solatorio(2024)]{solatorio2024gistembed}
Solatorio, A.~V.
\newblock Gistembed: Guided in-sample selection of training negatives for text embedding fine-tuning.
\newblock \emph{arXiv preprint arXiv:2402.16829}, 2024.
\newblock URL \url{https://arxiv.org/abs/2402.16829}.

\bibitem[Song et~al.(2023)Song, Xie, Zhang, and Luo]{song2023multi}
Song, K., Xie, J., Zhang, S., and Luo, Z.
\newblock Multi-mode online knowledge distillation for self-supervised visual representation learning.
\newblock In \emph{Proceedings of the IEEE/CVF Conference on Computer Vision and Pattern Recognition}, pp.\  11848--11857, 2023.

\bibitem[Stewart et~al.(2023)Stewart, Bach, Berthet, and Vert]{stewart2023regressionclassificationinfluencetask}
Stewart, L., Bach, F., Berthet, Q., and Vert, J.-P.
\newblock Regression as classification: Influence of task formulation on neural network features, 2023.
\newblock URL \url{https://arxiv.org/abs/2211.05641}.

\bibitem[Stärk et~al.(2021)Stärk, Beaini, Corso, Tossou, Dallago, Günnemann, and Liò]{stark2021_3dinfomax}
Stärk, H., Beaini, D., Corso, G., Tossou, P., Dallago, C., Günnemann, S., and Liò, P.
\newblock 3d infomax improves gnns for molecular property prediction.
\newblock \emph{arXiv preprint arXiv:2110.04126}, 2021.

\bibitem[Su et~al.(2023)Su, Shi, Kasai, Wang, Hu, Ostendorf, tau Yih, Smith, Zettlemoyer, and Yu]{su2023embeddertaskinstructionfinetunedtext}
Su, H., Shi, W., Kasai, J., Wang, Y., Hu, Y., Ostendorf, M., tau Yih, W., Smith, N.~A., Zettlemoyer, L., and Yu, T.
\newblock One embedder, any task: Instruction-finetuned text embeddings, 2023.
\newblock URL \url{https://arxiv.org/abs/2212.09741}.

\bibitem[Sun et~al.(2024)Sun, Ren, Li, Wang, and Cao]{sun2024logit}
Sun, S., Ren, W., Li, J., Wang, R., and Cao, X.
\newblock Logit standardization in knowledge distillation.
\newblock In \emph{Proceedings of the IEEE/CVF Conference on Computer Vision and Pattern Recognition}, pp.\  15731--15740, 2024.

\bibitem[Szegedy et~al.(2015)Szegedy, Liu, Jia, Sermanet, Reed, Anguelov, Erhan, Vanhoucke, and Rabinovich]{szegedy2015going}
Szegedy, C., Liu, W., Jia, Y., Sermanet, P., Reed, S., Anguelov, D., Erhan, D., Vanhoucke, V., and Rabinovich, A.
\newblock Going deeper with convolutions.
\newblock In \emph{Proceedings of the IEEE conference on computer vision and pattern recognition}, pp.\  1--9, 2015.

\bibitem[Tan et~al.(2019)Tan, Chen, Pang, Vasudevan, Sandler, Howard, and Le]{tan2019mnasnet}
Tan, M., Chen, B., Pang, R., Vasudevan, V., Sandler, M., Howard, A., and Le, Q.~V.
\newblock Mnasnet: Platform-aware neural architecture search for mobile.
\newblock In \emph{Proceedings of the IEEE/CVF conference on computer vision and pattern recognition}, pp.\  2820--2828, 2019.

\bibitem[Theisen et~al.(2024)Theisen, Kim, Yang, Hodgkinson, and Mahoney]{theisen2024ensembles}
Theisen, R., Kim, H., Yang, Y., Hodgkinson, L., and Mahoney, M.~W.
\newblock When are ensembles really effective?
\newblock \emph{Advances in Neural Information Processing Systems}, 36, 2024.

\bibitem[Touvron et~al.(2023)Touvron, Martin, Stone, Albert, Almahairi, Babaei, Bashlykov, Batra, Bhargava, Bhosale, Bikel, Blecher, Ferrer, Chen, Cucurull, Esiobu, Fernandes, Fu, Fu, Fuller, Gao, Goswami, Goyal, Hartshorn, Hosseini, Hou, Inan, Kardas, Kerkez, Khabsa, Kloumann, Korenev, Koura, Lachaux, Lavril, Lee, Liskovich, Lu, Mao, Martinet, Mihaylov, Mishra, Molybog, Nie, Poulton, Reizenstein, Rungta, Saladi, Schelten, Silva, Smith, Subramanian, Tan, Tang, Taylor, Williams, Kuan, Xu, Yan, Zarov, Zhang, Fan, Kambadur, Narang, Rodriguez, Stojnic, Edunov, and Scialom]{touvron2023llama}
Touvron, H., Martin, L., Stone, K., Albert, P., Almahairi, A., Babaei, Y., Bashlykov, N., Batra, S., Bhargava, P., Bhosale, S., Bikel, D., Blecher, L., Ferrer, C.~C., Chen, M., Cucurull, G., Esiobu, D., Fernandes, J., Fu, J., Fu, W., Fuller, B., Gao, C., Goswami, V., Goyal, N., Hartshorn, A., Hosseini, S., Hou, R., Inan, H., Kardas, M., Kerkez, V., Khabsa, M., Kloumann, I., Korenev, A., Koura, P.~S., Lachaux, M.-A., Lavril, T., Lee, J., Liskovich, D., Lu, Y., Mao, Y., Martinet, X., Mihaylov, T., Mishra, P., Molybog, I., Nie, Y., Poulton, A., Reizenstein, J., Rungta, R., Saladi, K., Schelten, A., Silva, R., Smith, E.~M., Subramanian, R., Tan, X.~E., Tang, B., Taylor, R., Williams, A., Kuan, J.~X., Xu, P., Yan, Z., Zarov, I., Zhang, Y., Fan, A., Kambadur, M., Narang, S., Rodriguez, A., Stojnic, R., Edunov, S., and Scialom, T.
\newblock Llama 2: Open foundation and fine-tuned chat models, 2023.

\bibitem[Vilnis \& McCallum(2015)Vilnis and McCallum]{DBLP:journals/corr/VilnisM14}
Vilnis, L. and McCallum, A.
\newblock Word representations via gaussian embedding.
\newblock In Bengio, Y. and LeCun, Y. (eds.), \emph{3rd International Conference on Learning Representations, {ICLR} 2015, San Diego, CA, USA, May 7-9, 2015, Conference Track Proceedings}, 2015.
\newblock URL \url{http://arxiv.org/abs/1412.6623}.

\bibitem[Wang et~al.(2022{\natexlab{a}})Wang, Li, Jin, Cho, Ji, Han, and Burke]{wang2022chemicalreactionaware}
Wang, H., Li, W., Jin, X., Cho, K., Ji, H., Han, J., and Burke, M.~D.
\newblock Chemical-reaction-aware molecule representation learning.
\newblock In \emph{International Conference on Learning Representations}, 2022{\natexlab{a}}.
\newblock URL \url{https://openreview.net/forum?id=6sh3pIzKS-}.

\bibitem[Wang et~al.(2023)Wang, Ma, Zhang, Zhang, Avery, Hull, and Carneiro]{wang2023learnable}
Wang, H., Ma, C., Zhang, J., Zhang, Y., Avery, J., Hull, L., and Carneiro, G.
\newblock Learnable cross-modal knowledge distillation for multi-modal learning with missing modality.
\newblock In \emph{International Conference on Medical Image Computing and Computer-Assisted Intervention}, pp.\  216--226. Springer, 2023.

\bibitem[Wang et~al.(2022{\natexlab{b}})Wang, Yang, and van~de Weijer]{wang2022attention}
Wang, K., Yang, F., and van~de Weijer, J.
\newblock Attention distillation: self-supervised vision transformer students need more guidance.
\newblock \emph{arXiv preprint arXiv:2210.00944}, 2022{\natexlab{b}}.

\bibitem[Wang et~al.(2022{\natexlab{c}})Wang, Xie, Li, Fan, Song, Liang, Lu, Luo, and Shao]{wang2022pvt}
Wang, W., Xie, E., Li, X., Fan, D.-P., Song, K., Liang, D., Lu, T., Luo, P., and Shao, L.
\newblock Pvt v2: Improved baselines with pyramid vision transformer.
\newblock \emph{Computational Visual Media}, 8\penalty0 (3):\penalty0 415--424, 2022{\natexlab{c}}.

\bibitem[Welinder et~al.(2010)Welinder, Branson, Mita, Wah, Schroff, Belongie, and Perona]{WelinderEtal2010}
Welinder, P., Branson, S., Mita, T., Wah, C., Schroff, F., Belongie, S., and Perona, P.
\newblock {Caltech-UCSD Birds 200}.
\newblock Technical Report CNS-TR-2010-001, California Institute of Technology, 2010.

\bibitem[Wenlock \& Tomkinson(2021)Wenlock and Tomkinson]{lipo}
Wenlock, M. and Tomkinson, N.
\newblock Experimental in vitro dmpk and physicochemical data on a set of publicly disclosed compounds, 2021.
\newblock URL \url{https://www.ebi.ac.uk/chembl/document_report_card/CHEMBL3301361/}.

\bibitem[Wolf et~al.(2020)Wolf, Debut, Sanh, Chaumond, Delangue, Moi, Cistac, Rault, Louf, Funtowicz, Davison, Shleifer, von Platen, Ma, Jernite, Plu, Xu, Scao, Gugger, Drame, Lhoest, and Rush]{wolf2020huggingfacestransformersstateoftheartnatural}
Wolf, T., Debut, L., Sanh, V., Chaumond, J., Delangue, C., Moi, A., Cistac, P., Rault, T., Louf, R., Funtowicz, M., Davison, J., Shleifer, S., von Platen, P., Ma, C., Jernite, Y., Plu, J., Xu, C., Scao, T.~L., Gugger, S., Drame, M., Lhoest, Q., and Rush, A.~M.
\newblock Huggingface's transformers: State-of-the-art natural language processing, 2020.
\newblock URL \url{https://arxiv.org/abs/1910.03771}.

\bibitem[Wu et~al.(2018)Wu, Ramsundar, Feinberg, Gomes, Geniesse, Pappu, Leswing, and Pande]{wu_moleculenet_2018}
Wu, Z., Ramsundar, B., Feinberg, E.~N., Gomes, J., Geniesse, C., Pappu, A.~S., Leswing, K., and Pande, V.
\newblock {MoleculeNet}: {A} {Benchmark} for {Molecular} {Machine} {Learning}, October 2018.
\newblock URL \url{http://arxiv.org/abs/1703.00564}.

\bibitem[Xie et~al.(2021)Xie, Wang, Yu, Anandkumar, Alvarez, and Luo]{xie2021segformer}
Xie, E., Wang, W., Yu, Z., Anandkumar, A., Alvarez, J.~M., and Luo, P.
\newblock Segformer: Simple and efficient design for semantic segmentation with transformers.
\newblock \emph{Advances in neural information processing systems}, 34:\penalty0 12077--12090, 2021.

\bibitem[Xie et~al.(2017)Xie, Girshick, Doll{\'a}r, Tu, and He]{xie2017aggregated}
Xie, S., Girshick, R., Doll{\'a}r, P., Tu, Z., and He, K.
\newblock Aggregated residual transformations for deep neural networks.
\newblock In \emph{Proceedings of the IEEE conference on computer vision and pattern recognition}, pp.\  1492--1500, 2017.

\bibitem[Xu et~al.(2022{\natexlab{a}})Xu, Fang, Zhang, Xie, Wang, Dai, Xiong, and Tian]{xu2022baginstancesaggregationboosts}
Xu, H., Fang, J., Zhang, X., Xie, L., Wang, X., Dai, W., Xiong, H., and Tian, Q.
\newblock Bag of instances aggregation boosts self-supervised distillation, 2022{\natexlab{a}}.
\newblock URL \url{https://arxiv.org/abs/2107.01691}.

\bibitem[Xu et~al.(2022{\natexlab{b}})Xu, Koehn, and Murray]{xu-etal-2022-importance}
Xu, H., Koehn, P., and Murray, K.
\newblock The importance of being parameters: An intra-distillation method for serious gains.
\newblock In Goldberg, Y., Kozareva, Z., and Zhang, Y. (eds.), \emph{Proceedings of the 2022 Conference on Empirical Methods in Natural Language Processing}, pp.\  170--183, Abu Dhabi, United Arab Emirates, December 2022{\natexlab{b}}. Association for Computational Linguistics.
\newblock \doi{10.18653/v1/2022.emnlp-main.13}.
\newblock URL \url{https://aclanthology.org/2022.emnlp-main.13}.

\bibitem[Xu et~al.(2019)Xu, Hu, Leskovec, and Jegelka]{xu2018how}
Xu, K., Hu, W., Leskovec, J., and Jegelka, S.
\newblock How powerful are graph neural networks?
\newblock In \emph{International Conference on Learning Representations}, 2019.
\newblock URL \url{https://openreview.net/forum?id=ryGs6iA5Km}.

\bibitem[Xu et~al.(2021)Xu, Wang, Ni, Guo, and Tang]{xu2021selfsupervised}
Xu, M., Wang, H., Ni, B., Guo, H., and Tang, J.
\newblock Self-supervised graph-level representation learning with local and global structure.
\newblock \emph{arXiv preprint arXiv:2106.04113}, 2021.

\bibitem[Ye et~al.(2024)Ye, Jiang, Tian, Zhang, and Chen]{ye2024knowledge}
Ye, X., Jiang, R., Tian, X., Zhang, R., and Chen, Y.
\newblock Knowledge distillation via multi-teacher feature ensemble.
\newblock \emph{IEEE Signal Processing Letters}, 2024.

\bibitem[Yim et~al.(2017)Yim, Joo, Bae, and Kim]{8100237}
Yim, J., Joo, D., Bae, J., and Kim, J.
\newblock A gift from knowledge distillation: Fast optimization, network minimization and transfer learning.
\newblock In \emph{2017 IEEE Conference on Computer Vision and Pattern Recognition (CVPR)}, pp.\  7130--7138, 2017.
\newblock \doi{10.1109/CVPR.2017.754}.

\bibitem[You et~al.(2020)You, Chen, Sui, Chen, Wang, and Shen]{You2020GraphCL}
You, Y., Chen, T., Sui, Y., Chen, T., Wang, Z., and Shen, Y.
\newblock Graph contrastive learning with augmentations.
\newblock In Larochelle, H., Ranzato, M., Hadsell, R., Balcan, M.~F., and Lin, H. (eds.), \emph{Advances in Neural Information Processing Systems}, volume~33, pp.\  5812--5823. Curran Associates, Inc., 2020.
\newblock URL \url{https://proceedings.neurips.cc/paper/2020/file/3fe230348e9a12c13120749e3f9fa4cd-Paper.pdf}.

\bibitem[Yuan et~al.(2020)Yuan, Shou, Pei, Lin, Gong, Fu, and Jiang]{yuan2020reinforced}
Yuan, F., Shou, L., Pei, J., Lin, W., Gong, M., Fu, Y., and Jiang, D.
\newblock Reinforced multi-teacher selection for knowledge distillation, 2020.

\bibitem[Zagoruyko \& Komodakis(2017)Zagoruyko and Komodakis]{zagoruyko2017wideresidualnetworks}
Zagoruyko, S. and Komodakis, N.
\newblock Wide residual networks, 2017.
\newblock URL \url{https://arxiv.org/abs/1605.07146}.

\bibitem[Zhang et~al.(2023)Zhang, Chen, and Wang]{zhang2023adaptive}
Zhang, H., Chen, D., and Wang, C.
\newblock Adaptive multi-teacher knowledge distillation with meta-learning.
\newblock In \emph{2023 IEEE International Conference on Multimedia and Expo (ICME)}, pp.\  1943--1948. IEEE, 2023.

\bibitem[Zhang et~al.(2019)Zhang, Song, Gao, Chen, Bao, and Ma]{zhang2019your}
Zhang, L., Song, J., Gao, A., Chen, J., Bao, C., and Ma, K.
\newblock Be your own teacher: Improve the performance of convolutional neural networks via self distillation.
\newblock In \emph{Proceedings of the IEEE/CVF international conference on computer vision}, pp.\  3713--3722, 2019.

\bibitem[Zhang et~al.(2015)Zhang, Zhao, and LeCun]{Zhang2015CharacterlevelCN}
Zhang, X., Zhao, J.~J., and LeCun, Y.
\newblock Character-level convolutional networks for text classification.
\newblock In \emph{NIPS}, 2015.

\bibitem[Zhu et~al.(2020)Zhu, Liu, Li, Lai, He, Chen, and Zheng]{zhu2020ensembled}
Zhu, J., Liu, J., Li, W., Lai, J., He, X., Chen, L., and Zheng, Z.
\newblock Ensembled ctr prediction via knowledge distillation.
\newblock In \emph{Proceedings of the 29th ACM International Conference on Information \& Knowledge Management}, pp.\  2941--2958, 2020.

\end{thebibliography}
\bibliographystyle{icml2025}
\appendix
\onecolumn
    \addcontentsline{toc}{section}{Appendix}
    \part{Appendix}
    \parttoc
    \newpage

\section{Proofs}
\label{sec:appendix_theoretical_result}
We denote $\mathbf{X}$ as the random variable over $\Xrond$ that describes the input distribution. We suppose we have access to a dataset $\mathcal{D} = \set{\xbold_i}_{i=1}^n \subset \Xrond$ of inputs drawn following $p_\mathbf{X}$ and different embedders $\teachk : \Xrond \rightarrow \R^{d_k}$, $k \in \set{1, \ldots, K}$, that map the inputs to different embedding spaces. We denote $\mathbf{Z_k} = \teachk(\mathbf{X})$ as the random variable over $\R^{d_k}$ that describes the embedding of the input distribution in the $k$-th embedding space and by $\mathbf{U} = \stud(\mathbf{X})$ the random variable over $\R^{d}$ that describe the embedding of the input distribution in the student embedding space. We denote by $\zbold_i^k = \teachk(\xbold_i)$ the embedding of $\xbold_i$ in the $k$-th embedding space. We are interested in learning a representation that captures the information contained in all the embeddings.

Let us consider a task characterized by a target set $\Yrond$ of discrete concepts and the feature space $\mathcal{X}$ with joint probability measure $P_{\rmY\rmX} \in \mathcal{P}(\mathcal{Y}\times \mathcal{X})$. 
For every projection of the features through the different teachers, the Bayes decision rule  
$c_{\teachk}^* \triangleq  \argmax\limits_{c: \R^{d_k} \rightarrow \Yrond } \Esp_{\rmX \rmY}  \big[\mathbb{1} \left[c(\teachk(\rmX)) = \rmY \right]\big]$
and similarly for the student: $c_{\stud}^* \triangleq \argmax\limits_{c : \R^d \rightarrow \Yrond } \Esp_{\rmX \rmY}\big[\mathbb{1}[c(\stud(\rmX)) = \rmY]\big]$.

We leverage the following recent result from \citep{darrin2024textttcosmicmutualinformationtaskagnostic}:

\begin{proposition}
\label{prop:cosmic_b}
    Let $C_\teachk$ = $c_{\teachk}^*(\teachk(\rmX))$ and $C_\stud$ = $c_{\stud}^*(\stud(\rmX))$ denote the outcome of the Bayes classifier observing the output of the teacher $\teachk$ and the student $\stud$, respectively
        \begin{align}
           \label{eq:term_upper_bound}
           \Pr \left( C_\stud \neq C_\teachk \right) &\leq  1 - \exp\big(  - h(\teachk(\rmX) | \stud(\rmX)) \big).
        \end{align}
\end{proposition}

\subsection{Proof of \autoref{prop:loss_upper_bound}}
By applying the above proposition to all the terms in \autoref{eq:ideal_loss}, we obtain the following bound on the loss function:

\begin{prop}[Upper bound]
    \begin{align}
        \mathcal{L}^*(\mathbf{X}\mathbf{Y}, \stud, \teach_1, \ldots, \teach_K) &\leq \frac1K\sum_{k=1}^K \left( 1 -  \exp\big( -  h(\teachk(\mathbf{X}) | \stud(\mathbf{X})) \big) \right) \\
        &\leq 1 - \exp\left(  - \underbrace{\frac{1}{K} \sum_{k=1}^K h(\teachk(\mathbf{X}) | \stud(\mathbf{X}))}_{\text{Negative log likelihood}} \right).
    \end{align}
    \begin{proof}
        \begin{align*}
            \mathcal{L}^*(\mathbf{X}\mathbf{Y}, \stud, \teach_1, \ldots, \teach_K) &\leq \frac1K\sum_{k=1}^K \left( 1 -  \exp\big( -  h(\teachk(\mathbf{X}) | \stud(\mathbf{X})) \big) \right) \\
            &\leq 1 -  \frac1K \sum_{k=1}^K \exp\big(  - h(\teachk(\mathbf{X}) | \stud(\mathbf{X})) \big) \\
            &\leq 1 +  \frac1K \sum_{k=1}^K - \exp\big(  - h(\teachk(\mathbf{X}) | \stud(\mathbf{X})) \big) \\
            &\leq 1 - \exp\left(-  \frac1K \sum_{k=1}^K  h(\teachk(\mathbf{X}) | \stud(\mathbf{X})) \right). 
            \end{align*}
        We simply rearrange the terms and use the fact that $x \mapsto - \exp(-x)$ is concave to interchange the sum and the exponential.
    \end{proof}
\end{prop}

    \newpage

\section{Molecular Modelling}

\subsection{Model architecture}

\label{sec:appendix_mol_model_model_architecture}

We trained a 10-layer GINE~\citep{hu2020strategiespretraininggraphneural} neural network with a 512 hidden dimension, using a 2-layer network for the message passing process.
We use the atomic number of each node as input, as well as possible chirality information, and the nature of the bond between each pair of nodes.
We use a batch size of 256 and a learning rate of $1e-4$ to train the model for 400 epochs on the 250k dataset and 200 epochs on the 2M dataset.
For the teacher-specific kernels, we used a 3-layer MLP with a hidden size of 1024.

\subsubsection{Chosen Teachers}

The teachers used to train our molecular modeling students are summed up in~\autoref{tab:teach}.
We gathered various representation models for molecular modeling, with different pre-training objectives, input modalities, architectures, and training datasets.

\begin{table}[h]
    \centering
    \caption{Description of all teachers used in our experiments.}
    
\centering
    \resizebox{\textwidth}{!}{
        \begin{tabular}{c| c | c |  c | c c c}
            \toprule
            Model name & \small SMILES & \small 2D-GNN &  \small 3D-GNN & Architecture & Out size & Dataset (size)\\
            \midrule
            GraphCL\tiny\citep{You2020GraphCL} & & \checkmark & & \small GIN  & 300 & GEOM~\tiny\citep{axelrod2022geom} \small (50k)\\
            GraphLog\tiny\citep{xu2021selfsupervised} & & \checkmark & & \small GIN  & 300 & GEOM~\tiny\citep{axelrod2022geom} \small (50k)\\
            GraphMVP\tiny\citep{liu2022pretraining}\footnotemark[1] & & \checkmark & & \small GIN  & 300 & GEOM~\tiny\citep{axelrod2022geom} \small (50k)\\
            3D-infomax\tiny\citep{stark2021_3dinfomax}\footnotemark[1] & & \checkmark & & \small PNA & 800 & QMugs~\tiny\citep{isert2021qmugs} \small (620k)\\
            \midrule
            ChemBERT \small MTR\tiny\citep{ahmad2022chemberta2}\footnotemark[2] & \checkmark &  & & \small RoBERTa & 384 & PubChem~\tiny\citep{pubchem} \small (5M, 10M, 77M)\\
            \midrule
            3D-fractional\tiny\citep{pmlr-v202-feng23c} & & & \checkmark & TorchMD-net & 256 & PCQM4Mv2\citep{hu2021ogblsc} \small (3.7M)\\
            \bottomrule
        \end{tabular}
    }
    \vspace{0.2cm}

    \label{tab:teach}
\end{table}

\footnotetext[1]{Models aiming at incorporating 3D information into 2D-GNNs models.}
\footnotetext[2]{We used the three versions of ChemBERT-MTR models trained on 5M, 10M, and 77M.}

\subsubsection{Architecture influence}

\begin{figure}[h]
    \centering
    \includegraphics[width=0.7\linewidth]{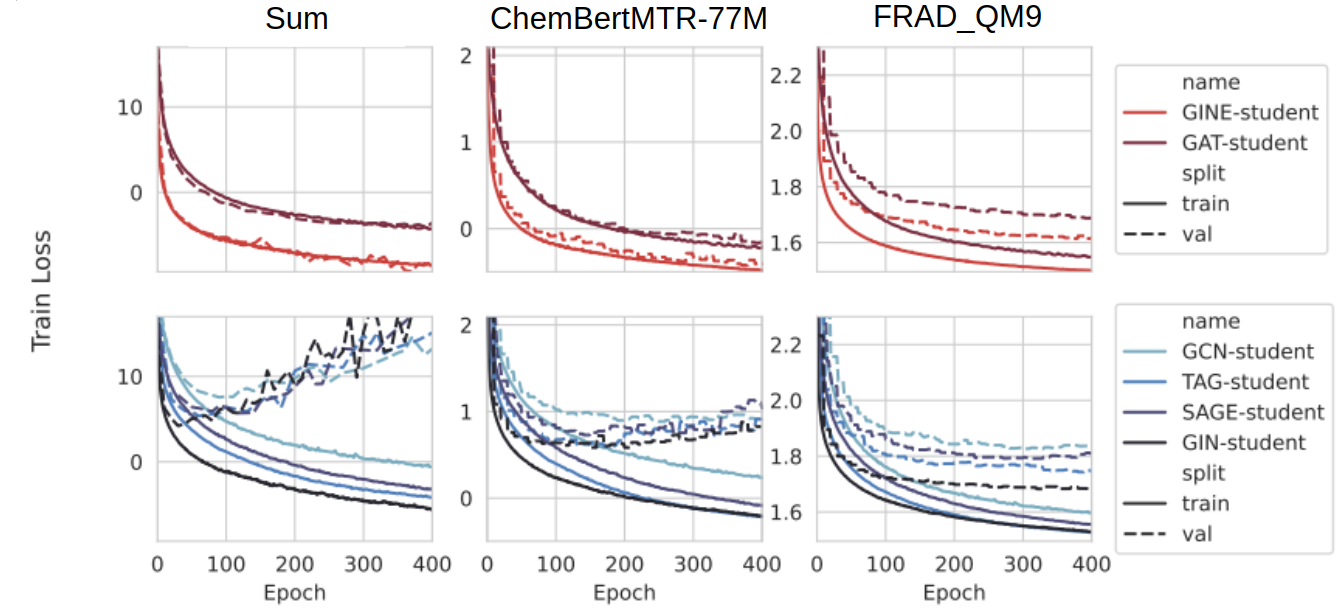}
    \caption{
        Training loss of different students using different GNN architectures on the ZINC-250k dataset.
    }
    \label{fig:gnn_train_curve}
\end{figure}

\autoref{fig:gnn_train_curve} shows the training loss of the student model with different GNN architectures on the ZINC-250k dataset.
In particular, we compared the GINE architecture with a Graph Convolutional Network (GCN)~\citep{morris2021weisfeilerlemanneuralhigherorder}, a Graph Attention Network (GAT)~\citep{brody2022attentivegraphattentionnetworks}, a GraphSAGE (SAGE)~\citep{hamilton2018inductiverepresentationlearninglarge}, a Toplogy Adaptative Graph Convolutional Network (TAG)~\citep{brody2022attentivegraphattentionnetworks}, and a GIN Network, that separates from the GINE architecture by the fact that it does not take edge features into account~\citep{xu2018how}.
We observe that the GINE architectures outperform the other architectures, with a lower training loss, a faster convergence, and a lower validation loss.
The Graph attention network (GAT) is the second best performing architecture, but it is still outperformed by the GINE architecture.
These two architectures are the only ones to use the edge embeddings in the message passing process, which could explain their better performance.

Indeed, all other architectures perform worse, especially when considering their validation loss computed on 10\% of the training set.
Specifically, the GIN architecture, not using edge feature, performs significantly worse than the GINE architecture, while having a similar architecture.

For our experiments, we decided to use the GINE architecture, as it performs the best during training and converges faster than the other architectures.

\subsubsection{Additional results on the TDC datasets}
\label{sec:appendix_mol_model_results_TDC}

\begin{table*}[ht]
    \centering
    \caption{Average rank of each model on the ADMET and HTS downstream tasks from the TDC~\citep{Huang2021tdc} platform. Our student outperforms all baselines, including teachers, on average.}
    \resizebox{\textwidth}{!}{\begin{tabular}{r|cccccc|c}
{} & {Absorption} & {Distribution} & {Metabolism} & {Excretion} & {Tox} & {HTS} & {Avg} \\
\midrule
InfoGraph & 13.50 & 13.27 & 13.32 & 11.40 & 11.98 & 9.40 & 12.14 \\
ChemBertMLM-10M & 10.65 & 11.00 & 10.70 & 13.80 & 11.11 & 14.60 & 11.98 \\
FRAD QM9${}^{(t)}$ & 10.57 & 11.13 & 10.38 & 8.33 & 10.04 & 7.80 & 9.71 \\
ChemGPT-1.2B & 9.55 & 11.73 & 11.75 & 10.73 & 10.86 & 11.20 & 10.97 \\
GROVER & 10.43 & 8.33 & 11.25 & 8.53 & 10.38 & 11.00 & 9.99 \\
GraphCL${}^{(t)}$ & 10.89 & 8.53 & 9.45 & 10.13 & 8.70 & 9.80 & 9.58 \\
GraphLog${}^{(t)}$ & 11.05 & 7.80 & 9.07 & 10.53 & 8.93 & 14.00 & 10.23 \\
GraphMVP${}^{(t)}$ & 7.20 & 6.20 & 7.85 & 9.80 & 7.49 & 8.80 & 7.89 \\
MolR gat & 6.95 & 7.60 & 8.30 & 8.53 & 6.49 & \underline{3.40} & 6.88 \\
ThreeDInfomax${}^{(t)}$ & \underline{4.17} & \underline{6.00} & 7.58 & 7.13 & 6.16 & 10.40 & 6.91 \\
ChemBertMTR-77M${}^{(t)}$ & \textbf{\underline{3.50}} & \textbf{\underline{4.27}} & 5.75 & 5.00 & 6.03 & 4.20 & \underline{4.79} \\
\midrule
MSE & 8.07 & 6.40 & 5.55 & 6.33 & 7.55 & \textbf{3.00} & 6.15 \\
\midrule
Cosine & 5.51 & 6.13 & \underline{3.60} & \underline{4.33} & \textbf{4.97} & 6.20 & 5.13 \\
\midrule
student-250k & \textbf{3.55} & 6.20 & \textbf{\underline{2.70}} & \textbf{\underline{2.40}} & \underline{4.99} & 3.80 & \textbf{3.94} \\
student-2M & 4.40 & \textbf{5.40} & \textbf{2.75} & \textbf{3.00} & \textbf{\underline{4.34}} & \textbf{\underline{2.40}} & \textbf{\underline{3.72}} \\
\end{tabular}
}
    \label{tab:molecule_results_all_rank}
\end{table*}

The average rank of each model in each task category can be found in~\autoref{tab:molecule_results_all_rank}.
Surprisingly, the performances of the "student-250k" and "student-2M" models are similar on average.
Specifically, the student-250k model outperforms the student-2M model on regression datasets notably, by achieving the best performances on the FreeSolv~\citep{mobley_freesolv_2014} and Lipophilicity~\citep{lipo} tasks.
This suggests that our method can leverage the diversity of the teachers to learn more informative representations, even when trained on a smaller dataset of 250k datapoints.

\FloatBarrier

\subsection{Kernel's predictive power}

\begin{figure}
    \centering
    \includegraphics[width=0.8\linewidth]{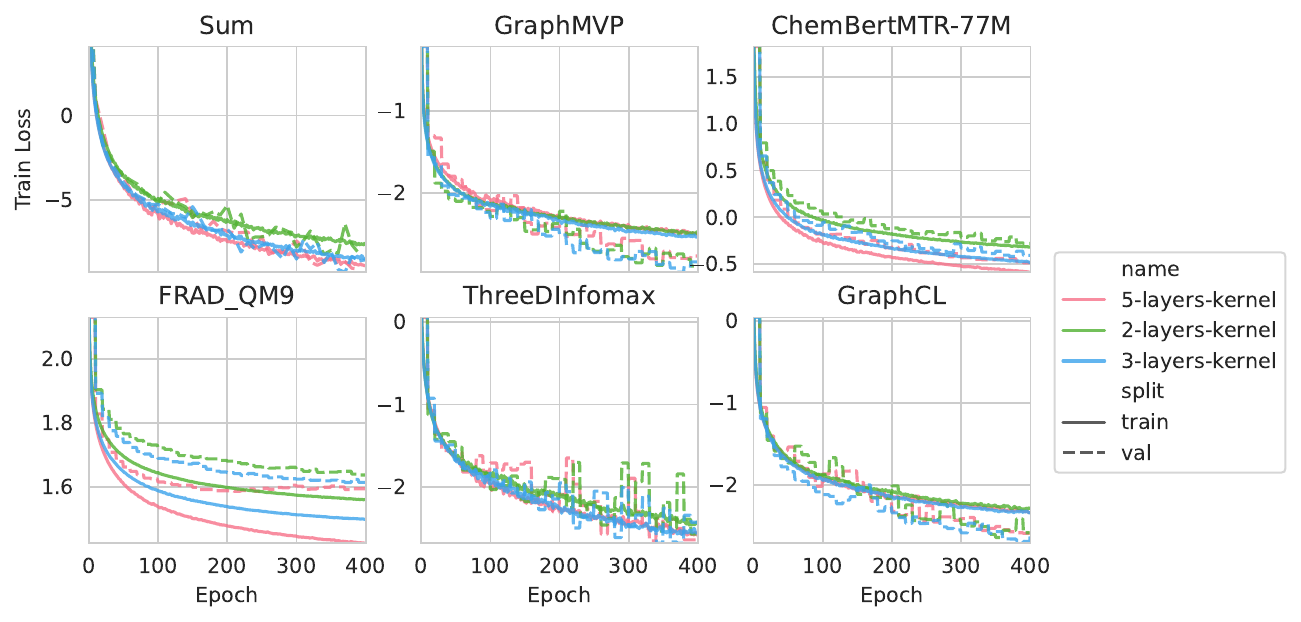}
    \caption{
        Training loss of the student model along the training with different kernel-size on the ZINC-250k dataset.
    }
    \label{fig:kernel_train_curve}
\end{figure}

\begin{figure}[ht]
\vskip 0.2in
\begin{center}
\centerline{\includegraphics[width=0.8\columnwidth]{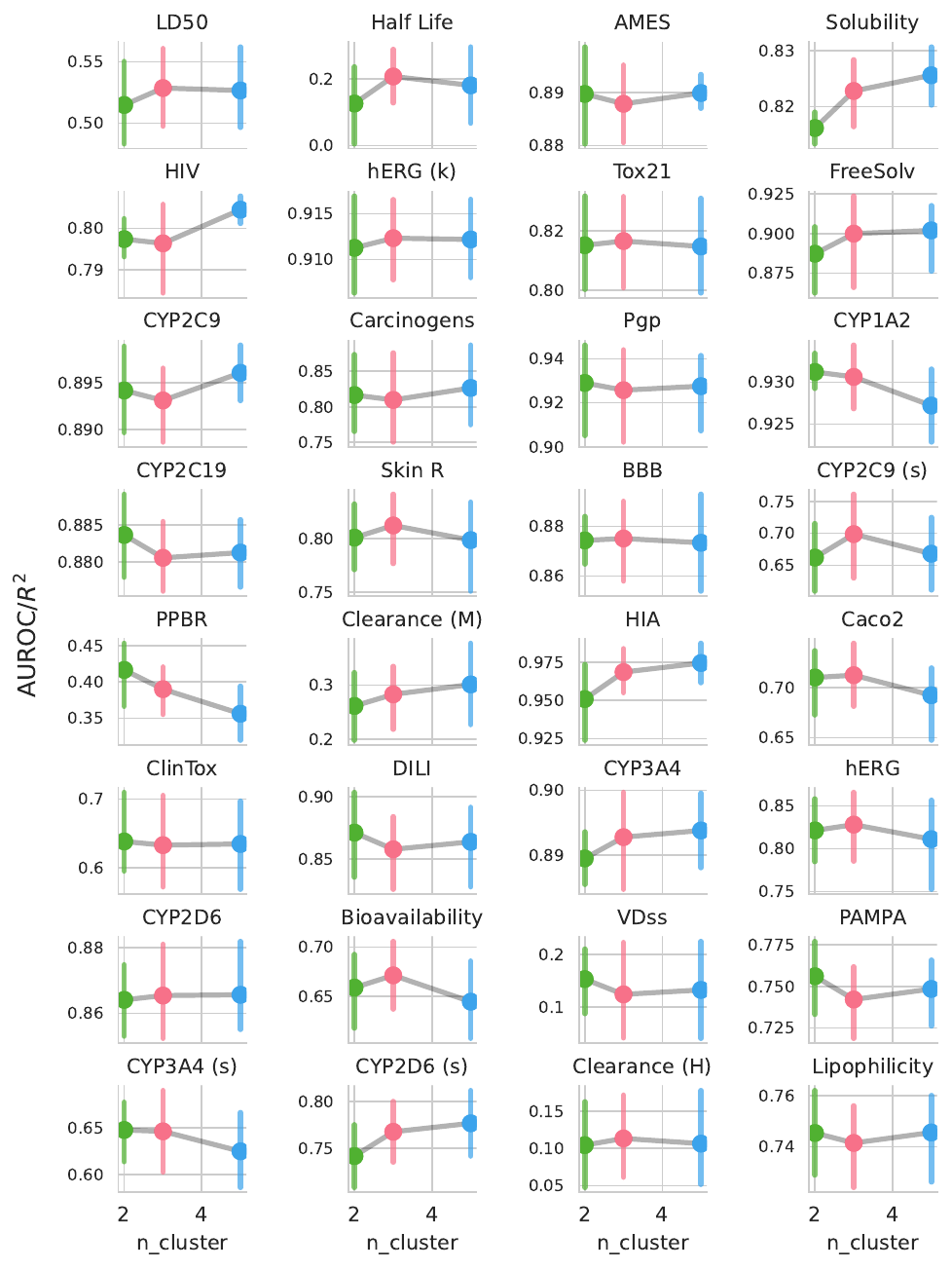}}
\caption{Test AUROC/$R^2$ score of the students on the classification/regression tasks, trained with different kernel-size on the ZINC-250k dataset.}
\label{fig:kernel}
\end{center}
\vskip -0.2in
\end{figure}

Our method relies on teacher-specific heads to distill the knowledge of each teacher.
In this section, we wish to evaluate the impact of the choice of these kernels and their predictive power (in terms of depth) on the performance and training of the student model.

We performed this experiment with kernels of depth 2, 3, and 5, and we trained the student model with these kernels on the ZINC-250k dataset and evaluated the performance of the student model on the ADMET and HTS downstream tasks.

First, during the training, as expected, the more powerful the kernel, the lower the training loss is (see~\autoref{fig:kernel_train_curve}), even though the difference is significant, especially between the students using kernels of depth 3 and 5.
Overall, the performances of each student on the downstream tasks are similar, underlining the robustness of our method regarding the choice of the kernel's depth (see~\autoref{fig:kernel}).
For our experiments in the main paper, we used a kernel of depth 3, as it enables the best trade-off between computational complexity, and training convergence while providing competitive results on the downstream tasks.

\FloatBarrier

\subsection{Evaluation details}
\label{sec:appendix_mol_model_eval_details}

\begin{wraptable}{R}{0.5\linewidth}
    \vskip -0.5cm
    \centering
    \caption{
        Tasks extracted from the Therapeutic Data Commons platform considered in our experiments.
    }

    \resizebox{\linewidth}{!}{\definecolor{r2}{rgb}{0.1, 0.0, 0.5}
\definecolor{aur}{rgb}{0.1, 0.3, 0.0}

\begin{tabular}{c c c | c c }
    \toprule
    Category & Model & Task & cls & reg \\
    \midrule
    \multirow{8}{*}{Absorption}
    & P-glycoprotein Inhibition & 1212 & \checkmark \\
    & AqSolDB & 9982 & & \checkmark \\
    & Lipophilicity & 4200 & & \checkmark \\
     & Caco-2 Permeability & 906 & & \checkmark \\
    & Human Intestinal Absorption & 578 & \checkmark &  \\
    & FreeSolv & 642 & & \checkmark \\
    & PAMPA Permeability & 2035 & \checkmark & \\
    & Oral Bioavailability & 640 & \checkmark & \\

    \midrule

    \multirow{3}{*}{Distribution}
    & Plasma-Protein BDR & 1614 &  & \checkmark \\
    & Blood-Brain barrier & 1975 & \checkmark &  \\
    & VDss & 1130 & & \checkmark \\

    \midrule

    \multirow{8}{*}{Metabolism}
    & CYPP450 3A4 Inhib. & 12328 & \checkmark & \\
    & CYPP450 1A2 Inhib. & 12579 & \checkmark & \\
    & CYPP450 2C19 Inhib. & 12665 & \checkmark & \\
    & CYPP450 2C9 Inhib. & 12092 & \checkmark & \\
    & CYPP450 2D6 Inhib. & 13130 & \checkmark & \\
    & CYPP450 2D6 Substrate & 664 & \checkmark & \\
    & CYPP450 3A4 Substrate & 667 & \checkmark & \\
    & CYPP450 2C9 Substrate & 666 & \checkmark & \\

    \midrule

    \multirow{3}{*}{Excretion}
    & Clearance hepatocyte & 1020 & & \checkmark \\
    & Half Life & 667 & & \checkmark \\
    & Clearance microsome & 1102 & & \checkmark \\

    \midrule

    \multirow{9}{*}{Toxicity}
            & Tox21 & 7831 & \checkmark & \\
    & \multirow{2}{*}{hERG} & 13445 & \checkmark & \\
    &                        & 648 & \checkmark & \\
    & Acute Toxicity LD50 & 7385 & & \checkmark \\
    & Ames Mutagenicity & 7255 & \checkmark & \\
    & ClinTox & 1484 & \checkmark & \\
    & Carcinogens & 278 & \checkmark & \\
    & Drug Induced Liver Injury & 475 & \checkmark & \\
    & Skin Reaction & 404 & \checkmark & \\

    \midrule

    \multirow{1}{*}{HTS}
            & HIV & 40000 & \checkmark & \\

    \bottomrule

\end{tabular}

}

    \label{tab:admet}
    \vskip -1cm
\end{wraptable}

\subsubsection{Benchmark Choice}
We selected a total of 32 tasks, extracted from the Therapeutic Data Commons~\citep{Huang2021tdc} platform, 8 absorption tasks, 3 distribution tasks, 8 metabolism tasks, 3 excretion tasks, 9 toxicity tasks and 1 high-throughput screening task.
A summary of the tasks considered can be found in~\autoref{tab:admet}, with their corresponding size (total number of samples) and type (classification or regression).
For all tasks, we computed 5 conformations for each molecule, and used the least energetic as an input of our 3D models.

\subsubsection{Evaluation Procedure}
For every task, we opted for a random split since we obtained similar results to a scaffold split, with a faster computation time, with a ratio of 70/10/20 for the train/validation/test sets.
For all tasks, we compute the embeddings generated by each model on the task.
We then train a 2 layer perceptron with a hidden size of 128 on the task for $\min(100,200 * \frac{5000}{\text{task size}})$ epochs (to limit the compute time on large tasks) with a learning rate of $1e-3$.
We then select the best checkpoint according to the validation performances and report the test metrics of this checkpoint.

\subsubsection{Evaluation Metrics}
We repeat this process five times with different seeds in the train-val-test splits in order to enable the establishment of robust rankings using autorank~\citep{Herbold2020}.
We decided to report the ranks of the models to enable the comparison of the models on both classification and regression by simply averaging the rank.
To compute the rank on all tasks, we rely on the AUROC score for classification tasks and the $R^2$ score for regression tasks.
For the excretion tasks, since the regression labels have a large variance, we decided to apply the regression on the log-values and report the $R^2$ score on the log-values.

\FloatBarrier

\subsection{Single-Teacher setting}
\label{sec:appendix_mol_single_teach}

To assess the impact of the multi-teacher setting on the performance of the student model, we trained students to distill the knowledge of a single teacher.
We used only the two best performing teachers, 3D-infomax~\citep{stark2021_3dinfomax} and ChemBERTaMTR~\citep{ahmad2022chemberta2}, to train the student model on the 2M datapoints dataset.
We also train a student with both teachers, to see if those two teachers are sufficient to achieve the same performance as the models we presented in the core of the paper.

\begin{figure}[ht]
    \centering
    \includegraphics[width=\textwidth]{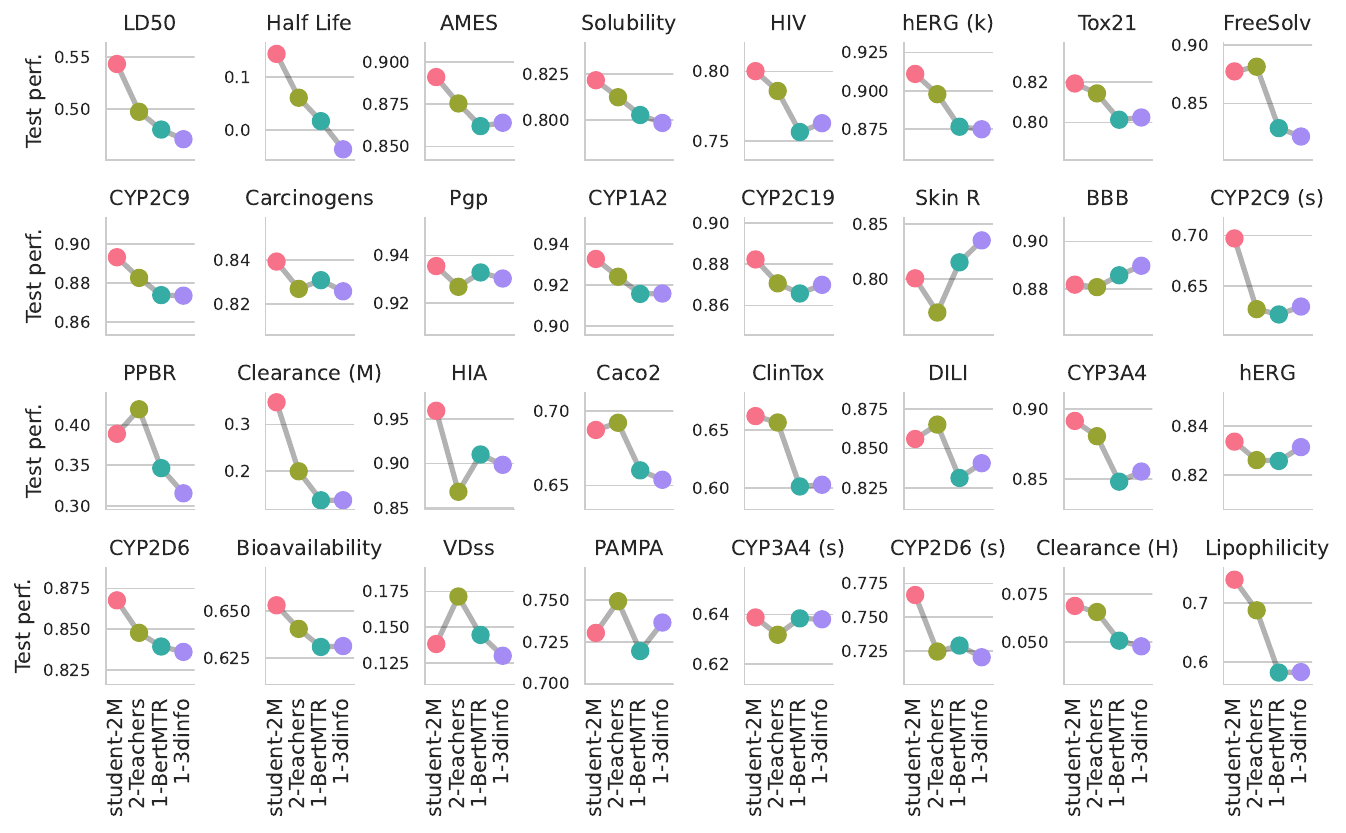}
    \caption{
        Test AUROC/$R^2$ score of the students on the classification/regression tasks, trained with all teachers (student-2M), two teachers (2-Teachers) and one teacher (1-ChemBertMTR for the model trained with ChemBertMTR-77M and 1-teacher-3dinfomax for the model trained with 3D-infomax).
    }
    \label{fig:multi_vs_single}
\end{figure}

~\autoref{fig:multi_vs_single} shows how these students underperform compared to a student trained with all teachers, in terms of AUROC for classification tasks and $R^2$ for regression tasks respectively.
These tables also show that the student trained with both teachers performs better than each student trained with only one teacher.
All results are aggregated in~\autoref{tab:sgl_cls} and~\autoref{tab:sgl_reg}.

\begin{table}
    \caption{Performance of the student models trained with only the best teacher ("1-ChemBertMTR"), the second-best teacher ("1-3dinfo"), both teachers together ("2-teachers"), and "student-2M" on regression tasks (R2).}
    \label{tab:sgl_reg}
    \begin{tabular}{r||c|cccc|c|}
{} & {\rotatebox{90}{\shortstack{  \\ avg}}} & {\rotatebox{90}{\shortstack{Absorption \\ Caco2}}} & {\rotatebox{90}{\shortstack{Absorption \\ FreeSolv}}} & {\rotatebox{90}{\shortstack{Absorption \\ Lipophilicity}}} & {\rotatebox{90}{\shortstack{Absorption \\ Solubility}}} & {\rotatebox{90}{\shortstack{Tox \\ LD50}}} \\
\midrule
1-3dinfo & 0.392$\pm$ \tiny 0.317 & 0.654$\pm$ \tiny 0.041 & 0.822$\pm$ \tiny 0.044 & 0.583$\pm$ \tiny 0.047 & 0.798$\pm$ \tiny 0.010 & 0.471$\pm$ \tiny 0.048 \\
1-BertMTR & 0.405$\pm$ \tiny 0.309 & 0.660$\pm$ \tiny 0.026 & 0.829$\pm$ \tiny 0.031 & 0.582$\pm$ \tiny 0.044 & 0.803$\pm$ \tiny 0.010 & 0.480$\pm$ \tiny 0.023 \\
2-Teachers & \textbf{0.449$\pm$ \tiny 0.312} & \textbf{\underline{0.692$\pm$ \tiny 0.043}} & \textbf{\underline{0.882$\pm$ \tiny 0.034}} & \textbf{0.688$\pm$ \tiny 0.028} & \textbf{0.812$\pm$ \tiny 0.012} & \textbf{0.497$\pm$ \tiny 0.033} \\
\midrule
student-2M & \textbf{\underline{0.476$\pm$ \tiny 0.301}} & \textbf{0.687$\pm$ \tiny 0.045} & \textbf{0.878$\pm$ \tiny 0.036} & \textbf{\underline{0.739$\pm$ \tiny 0.021}} & \textbf{\underline{0.822$\pm$ \tiny 0.005}} & \textbf{\underline{0.543$\pm$ \tiny 0.041}} \\
\end{tabular}

    \begin{tabular}{r||cc|ccc|}
{} & {\rotatebox{90}{\shortstack{Distribution \\ PPBR}}} & {\rotatebox{90}{\shortstack{Distribution \\ VDss}}} & {\rotatebox{90}{\shortstack{Excretion \\ Clearance (H)}}} & {\rotatebox{90}{\shortstack{Excretion \\ Clearance (M)}}} & {\rotatebox{90}{\shortstack{Excretion \\ Half Life}}} \\
\midrule
1-3dinfo & 0.316$\pm$ \tiny 0.062 & 0.130$\pm$ \tiny 0.146 & 0.048$\pm$ \tiny 0.095 & 0.137$\pm$ \tiny 0.083 & -0.037$\pm$ \tiny 0.254 \\
1-BertMTR & 0.347$\pm$ \tiny 0.070 & \textbf{0.145$\pm$ \tiny 0.072} & 0.051$\pm$ \tiny 0.148 & 0.136$\pm$ \tiny 0.110 & 0.017$\pm$ \tiny 0.195 \\
2-Teachers & \textbf{\underline{0.419$\pm$ \tiny 0.032}} & \textbf{\underline{0.172$\pm$ \tiny 0.098}} & \textbf{0.066$\pm$ \tiny 0.075} & \textbf{0.199$\pm$ \tiny 0.075} & \textbf{0.061$\pm$ \tiny 0.135} \\
\midrule
student-2M & \textbf{0.389$\pm$ \tiny 0.050} & 0.138$\pm$ \tiny 0.115 & \textbf{\underline{0.069$\pm$ \tiny 0.060}} & \textbf{\underline{0.348$\pm$ \tiny 0.062}} & \textbf{\underline{0.144$\pm$ \tiny 0.205}} \\
\end{tabular}

\end{table}

\begin{table}
    \caption{Performance of the student models trained with only the best teacher ("1-ChemBertMTR"), the second-best teacher ("1-3dinfo"), both teachers together ("2-teachers"), and "student-2M" on classification tasks (AUROC).}
    \label{tab:sgl_cls}
    \resizebox{\textwidth}{!}{
    \begin{tabular}{r||c|cccc|c|c|}
{} & {\rotatebox{90}{\shortstack{  \\ avg}}} & {\rotatebox{90}{\shortstack{Absorption \\ Bioavailability}}} & {\rotatebox{90}{\shortstack{Absorption \\ HIA}}} & {\rotatebox{90}{\shortstack{Absorption \\ PAMPA}}} & {\rotatebox{90}{\shortstack{Absorption \\ Pgp}}} & {\rotatebox{90}{\shortstack{Distribution \\ BBB}}} & {\rotatebox{90}{\shortstack{HTS \\ HIV}}} \\
\midrule
1-BertMTR & 0.801$\pm$ \tiny 0.101 & 0.631$\pm$ \tiny 0.059 & \textbf{0.910$\pm$ \tiny 0.054} & 0.719$\pm$ \tiny 0.020 & \textbf{0.933$\pm$ \tiny 0.021} & \textbf{0.886$\pm$ \tiny 0.022} & 0.756$\pm$ \tiny 0.005 \\
1-3dinfo & 0.803$\pm$ \tiny 0.100 & 0.631$\pm$ \tiny 0.050 & 0.899$\pm$ \tiny 0.056 & \textbf{0.737$\pm$ \tiny 0.021} & 0.930$\pm$ \tiny 0.024 & \textbf{\underline{0.890$\pm$ \tiny 0.024}} & 0.763$\pm$ \tiny 0.010 \\
2-Teachers & \textbf{0.808$\pm$ \tiny 0.097} & \textbf{0.641$\pm$ \tiny 0.055} & 0.868$\pm$ \tiny 0.051 & \textbf{\underline{0.749$\pm$ \tiny 0.009}} & 0.927$\pm$ \tiny 0.026 & 0.881$\pm$ \tiny 0.026 & \textbf{0.786$\pm$ \tiny 0.012} \\
\midrule
student-2M & \textbf{\underline{0.825$\pm$ \tiny 0.096}} & \textbf{\underline{0.653$\pm$ \tiny 0.055}} & \textbf{\underline{0.959$\pm$ \tiny 0.026}} & 0.730$\pm$ \tiny 0.024 & \textbf{\underline{0.936$\pm$ \tiny 0.024}} & 0.882$\pm$ \tiny 0.020 & \textbf{\underline{0.800$\pm$ \tiny 0.014}} \\
\end{tabular}
}
    \resizebox{\textwidth}{!}{
    \begin{tabular}{r||cccccccc|}
{} & {\rotatebox{90}{\shortstack{Metabolism \\ CYP1A2}}} & {\rotatebox{90}{\shortstack{Metabolism \\ CYP2C19}}} & {\rotatebox{90}{\shortstack{Metabolism \\ CYP2C9 (s)}}} & {\rotatebox{90}{\shortstack{Metabolism \\ CYP2C9}}} & {\rotatebox{90}{\shortstack{Metabolism \\ CYP2D6 (s)}}} & {\rotatebox{90}{\shortstack{Metabolism \\ CYP2D6}}} & {\rotatebox{90}{\shortstack{Metabolism \\ CYP3A4 (s)}}} & {\rotatebox{90}{\shortstack{Metabolism \\ CYP3A4}}} \\
\midrule
1-BertMTR & 0.916$\pm$ \tiny 0.008 & 0.866$\pm$ \tiny 0.007 & 0.622$\pm$ \tiny 0.088 & 0.874$\pm$ \tiny 0.006 & \textbf{0.729$\pm$ \tiny 0.021} & 0.839$\pm$ \tiny 0.010 & \textbf{0.638$\pm$ \tiny 0.017} & 0.848$\pm$ \tiny 0.014 \\
1-3dinfo & 0.916$\pm$ \tiny 0.005 & 0.870$\pm$ \tiny 0.006 & \textbf{0.630$\pm$ \tiny 0.069} & 0.874$\pm$ \tiny 0.005 & 0.721$\pm$ \tiny 0.028 & 0.836$\pm$ \tiny 0.008 & \textbf{0.638$\pm$ \tiny 0.015} & 0.855$\pm$ \tiny 0.007 \\
2-Teachers & \textbf{0.924$\pm$ \tiny 0.006} & \textbf{0.871$\pm$ \tiny 0.009} & 0.627$\pm$ \tiny 0.086 & \textbf{0.883$\pm$ \tiny 0.005} & 0.725$\pm$ \tiny 0.043 & \textbf{0.848$\pm$ \tiny 0.010} & 0.632$\pm$ \tiny 0.047 & \textbf{0.881$\pm$ \tiny 0.006} \\
\midrule
student-2M & \textbf{\underline{0.933$\pm$ \tiny 0.006}} & \textbf{\underline{0.882$\pm$ \tiny 0.007}} & \textbf{\underline{0.697$\pm$ \tiny 0.093}} & \textbf{\underline{0.893$\pm$ \tiny 0.002}} & \textbf{\underline{0.766$\pm$ \tiny 0.058}} & \textbf{\underline{0.868$\pm$ \tiny 0.010}} & \textbf{\underline{0.639$\pm$ \tiny 0.054}} & \textbf{\underline{0.892$\pm$ \tiny 0.004}} \\
\end{tabular}
}
    \resizebox{\textwidth}{!}{
    \begin{tabular}{r||cccccccc|}
{} & {\rotatebox{90}{\shortstack{Tox \\ AMES}}} & {\rotatebox{90}{\shortstack{Tox \\ Carcinogens}}} & {\rotatebox{90}{\shortstack{Tox \\ ClinTox}}} & {\rotatebox{90}{\shortstack{Tox \\ DILI}}} & {\rotatebox{90}{\shortstack{Tox \\ Skin R}}} & {\rotatebox{90}{\shortstack{Tox \\ Tox21}}} & {\rotatebox{90}{\shortstack{Tox \\ hERG}}} & {\rotatebox{90}{\shortstack{Tox \\ hERG (k)}}} \\
\midrule
1-BertMTR & 0.862$\pm$ \tiny 0.014 & \textbf{0.831$\pm$ \tiny 0.074} & 0.601$\pm$ \tiny 0.069 & 0.831$\pm$ \tiny 0.060 & \textbf{0.815$\pm$ \tiny 0.057} & 0.801$\pm$ \tiny 0.056 & 0.826$\pm$ \tiny 0.023 & 0.877$\pm$ \tiny 0.012 \\
1-3dinfo & 0.864$\pm$ \tiny 0.011 & 0.826$\pm$ \tiny 0.083 & 0.603$\pm$ \tiny 0.074 & 0.841$\pm$ \tiny 0.049 & \textbf{\underline{0.835$\pm$ \tiny 0.051}} & 0.802$\pm$ \tiny 0.055 & \textbf{0.831$\pm$ \tiny 0.019} & 0.875$\pm$ \tiny 0.007 \\
2-Teachers & \textbf{0.875$\pm$ \tiny 0.007} & 0.827$\pm$ \tiny 0.062 & \textbf{0.657$\pm$ \tiny 0.102} & \textbf{\underline{0.865$\pm$ \tiny 0.024}} & 0.770$\pm$ \tiny 0.053 & \textbf{0.814$\pm$ \tiny 0.057} & 0.826$\pm$ \tiny 0.029 & \textbf{0.898$\pm$ \tiny 0.006} \\
\midrule
student-2M & \textbf{\underline{0.891$\pm$ \tiny 0.014}} & \textbf{\underline{0.839$\pm$ \tiny 0.095}} & \textbf{\underline{0.662$\pm$ \tiny 0.072}} & \textbf{0.856$\pm$ \tiny 0.045} & 0.801$\pm$ \tiny 0.026 & \textbf{\underline{0.819$\pm$ \tiny 0.054}} & \textbf{\underline{0.834$\pm$ \tiny 0.019}} & \textbf{\underline{0.911$\pm$ \tiny 0.005}} \\
\end{tabular}

    }
\end{table}

\FloatBarrier

\subsection{Comprehensive results}
\label{sec:appendix_mol_model_comprehensive_results}

The following tables provide the raw results of the different evaluated models on the ADMET and HTS downstream tasks.
~\autoref{tab:molecule_all_cls} and~\autoref{tab:molecule_all_reg} display the test performances of the models on the classification and regression tasks respectively.
All regression tasks are evaluated using the $R^2$ score, while the classification tasks are evaluated using the AUROC score.
We report the mean values of the metrics over 5 runs for each task, as well as the standard deviation.

\begin{figure}
    \centering
    \includegraphics[width=0.8\linewidth]{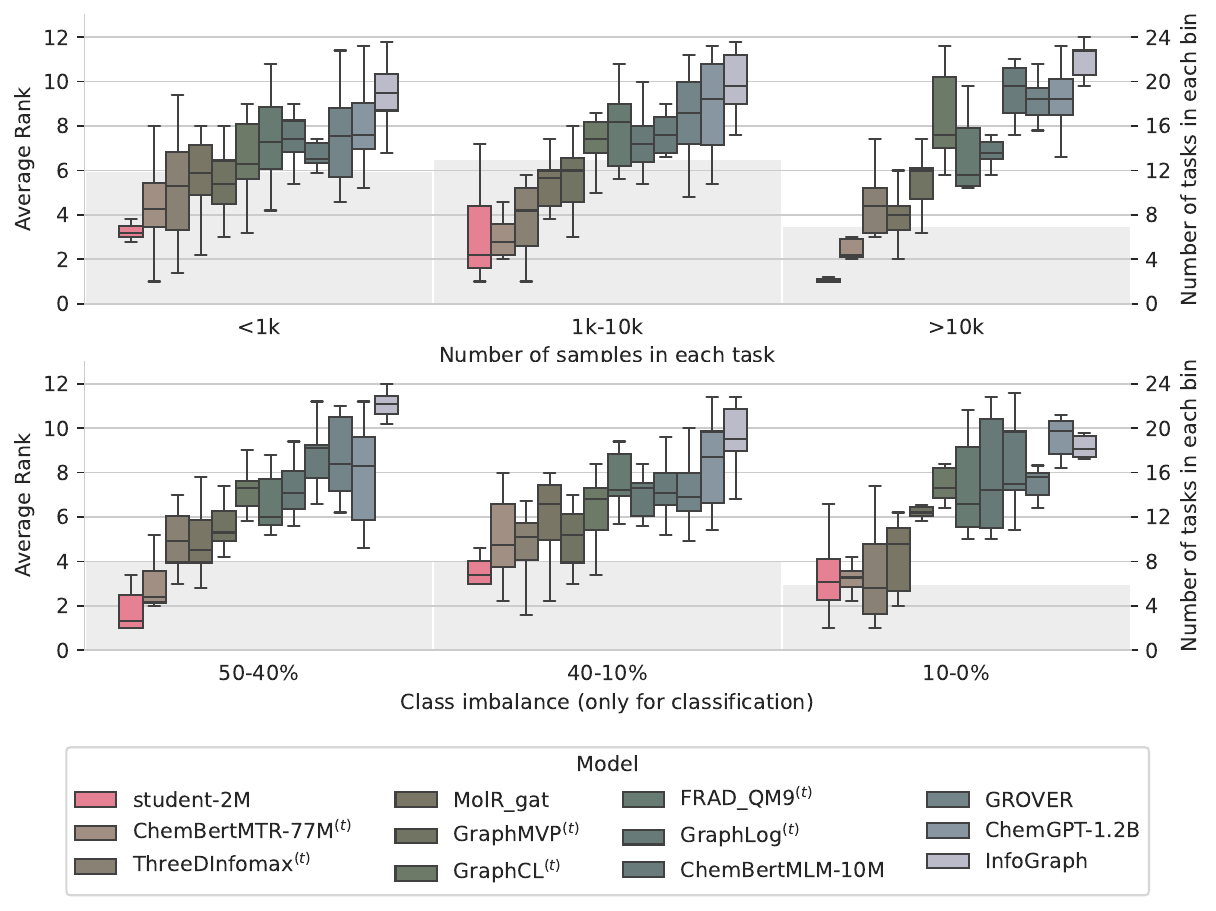}
    \caption{
        Average ranking of our models when grouping tasks based on the number of samples in the task and the class imbalance (for classification tasks).
    }
    \label{fig:splittask}
\end{figure}

    We display in~\autoref{fig:splittask} the evolution of the average rank of the embedders when separating the tasks based on the amount of samples, and the class imbalance (for classification tasks).
    Our student appears robust in both setups, even though as the class imbalance becomes more important, or as the amount of samples in the task decreases, the difference between the top-performing embedders becomes less significant.

\begin{table}
    \centering
    \caption{AUROC of each model on the ADMET and HTS downstream classification tasks. The best embedder for each task is highlighted in bold and underlined, and the second best is highlighted in bold.}
    \begin{minipage}{0.4\textwidth}
        \resizebox{\textwidth}{!}{
            \rotatebox{90}{\begin{tabular}{r||c|c|c|cccccccc|}
{} & {\rotatebox{-90}{\shortstack{  \\ avg}}} & {\rotatebox{-90}{\shortstack{Distribution \\ BBB}}} & {\rotatebox{-90}{\shortstack{HTS \\ HIV}}} & {\rotatebox{-90}{\shortstack{Metabolism \\ CYP1A2}}} & {\rotatebox{-90}{\shortstack{Metabolism \\ CYP2C19}}} & {\rotatebox{-90}{\shortstack{Metabolism \\ CYP2C9 (s)}}} & {\rotatebox{-90}{\shortstack{Metabolism \\ CYP2C9}}} & {\rotatebox{-90}{\shortstack{Metabolism \\ CYP2D6 (s)}}} & {\rotatebox{-90}{\shortstack{Metabolism \\ CYP2D6}}} & {\rotatebox{-90}{\shortstack{Metabolism \\ CYP3A4 (s)}}} & {\rotatebox{-90}{\shortstack{Metabolism \\ CYP3A4}}} \\
\midrule
InfoGraph & 0.768$\pm$ \tiny 0.097 & 0.843$\pm$ \tiny 0.022 & 0.769$\pm$ \tiny 0.018 & 0.878$\pm$ \tiny 0.003 & 0.832$\pm$ \tiny 0.004 & 0.624$\pm$ \tiny 0.085 & 0.840$\pm$ \tiny 0.007 & 0.674$\pm$ \tiny 0.031 & 0.813$\pm$ \tiny 0.013 & 0.567$\pm$ \tiny 0.042 & 0.817$\pm$ \tiny 0.003 \\
ChemGPT-1.2B & 0.779$\pm$ \tiny 0.094 & 0.853$\pm$ \tiny 0.012 & 0.760$\pm$ \tiny 0.014 & 0.886$\pm$ \tiny 0.008 & 0.835$\pm$ \tiny 0.004 & 0.638$\pm$ \tiny 0.034 & 0.852$\pm$ \tiny 0.008 & 0.670$\pm$ \tiny 0.014 & 0.816$\pm$ \tiny 0.020 & 0.608$\pm$ \tiny 0.026 & 0.844$\pm$ \tiny 0.006 \\
FRAD QM9${}^{(t)}$ & 0.785$\pm$ \tiny 0.111 & 0.869$\pm$ \tiny 0.013 & 0.779$\pm$ \tiny 0.005 & 0.906$\pm$ \tiny 0.004 & 0.845$\pm$ \tiny 0.010 & 0.609$\pm$ \tiny 0.049 & 0.860$\pm$ \tiny 0.006 & 0.711$\pm$ \tiny 0.025 & 0.815$\pm$ \tiny 0.018 & 0.607$\pm$ \tiny 0.060 & 0.855$\pm$ \tiny 0.003 \\
ChemBertMLM-10M & 0.785$\pm$ \tiny 0.089 & 0.868$\pm$ \tiny 0.009 & 0.733$\pm$ \tiny 0.012 & 0.886$\pm$ \tiny 0.007 & 0.843$\pm$ \tiny 0.007 & 0.643$\pm$ \tiny 0.020 & 0.851$\pm$ \tiny 0.009 & 0.733$\pm$ \tiny 0.017 & 0.813$\pm$ \tiny 0.014 & 0.606$\pm$ \tiny 0.029 & 0.846$\pm$ \tiny 0.005 \\
GROVER & 0.787$\pm$ \tiny 0.096 & 0.869$\pm$ \tiny 0.016 & 0.760$\pm$ \tiny 0.014 & 0.880$\pm$ \tiny 0.006 & 0.843$\pm$ \tiny 0.008 & 0.614$\pm$ \tiny 0.042 & 0.853$\pm$ \tiny 0.005 & 0.750$\pm$ \tiny 0.040 & 0.824$\pm$ \tiny 0.013 & 0.599$\pm$ \tiny 0.028 & 0.827$\pm$ \tiny 0.007 \\
GraphCL${}^{(t)}$ & 0.792$\pm$ \tiny 0.093 & 0.865$\pm$ \tiny 0.011 & 0.765$\pm$ \tiny 0.019 & 0.898$\pm$ \tiny 0.004 & 0.858$\pm$ \tiny 0.007 & 0.645$\pm$ \tiny 0.032 & 0.862$\pm$ \tiny 0.006 & 0.723$\pm$ \tiny 0.030 & 0.828$\pm$ \tiny 0.009 & 0.607$\pm$ \tiny 0.044 & 0.846$\pm$ \tiny 0.008 \\
GraphLog${}^{(t)}$ & 0.790$\pm$ \tiny 0.096 & 0.867$\pm$ \tiny 0.017 & 0.748$\pm$ \tiny 0.008 & 0.886$\pm$ \tiny 0.006 & 0.855$\pm$ \tiny 0.006 & 0.616$\pm$ \tiny 0.034 & 0.860$\pm$ \tiny 0.009 & \textbf{\underline{0.768$\pm$ \tiny 0.046}} & 0.832$\pm$ \tiny 0.017 & 0.624$\pm$ \tiny 0.049 & 0.847$\pm$ \tiny 0.009 \\
GraphMVP${}^{(t)}$ & 0.800$\pm$ \tiny 0.097 & 0.874$\pm$ \tiny 0.016 & 0.771$\pm$ \tiny 0.016 & 0.899$\pm$ \tiny 0.004 & 0.865$\pm$ \tiny 0.006 & 0.622$\pm$ \tiny 0.082 & 0.870$\pm$ \tiny 0.005 & 0.747$\pm$ \tiny 0.033 & 0.830$\pm$ \tiny 0.016 & 0.619$\pm$ \tiny 0.034 & 0.847$\pm$ \tiny 0.011 \\
MolR gat & 0.808$\pm$ \tiny 0.098 & 0.867$\pm$ \tiny 0.025 & \textbf{0.797$\pm$ \tiny 0.012} & 0.909$\pm$ \tiny 0.004 & 0.859$\pm$ \tiny 0.010 & 0.615$\pm$ \tiny 0.037 & 0.869$\pm$ \tiny 0.004 & 0.717$\pm$ \tiny 0.017 & 0.841$\pm$ \tiny 0.017 & 0.598$\pm$ \tiny 0.045 & 0.873$\pm$ \tiny 0.005 \\
ThreeDInfomax${}^{(t)}$ & 0.815$\pm$ \tiny 0.100 & \textbf{0.885$\pm$ \tiny 0.019} & 0.762$\pm$ \tiny 0.010 & 0.917$\pm$ \tiny 0.005 & 0.868$\pm$ \tiny 0.008 & 0.571$\pm$ \tiny 0.085 & 0.865$\pm$ \tiny 0.009 & 0.747$\pm$ \tiny 0.039 & 0.842$\pm$ \tiny 0.010 & 0.604$\pm$ \tiny 0.015 & 0.858$\pm$ \tiny 0.009 \\
ChemBertMTR-77M${}^{(t)}$ & 0.816$\pm$ \tiny 0.096 & \textbf{\underline{0.894$\pm$ \tiny 0.026}} & \textbf{0.797$\pm$ \tiny 0.014} & 0.919$\pm$ \tiny 0.007 & 0.873$\pm$ \tiny 0.005 & 0.600$\pm$ \tiny 0.114 & 0.877$\pm$ \tiny 0.007 & 0.734$\pm$ \tiny 0.024 & 0.845$\pm$ \tiny 0.015 & \textbf{0.645$\pm$ \tiny 0.051} & 0.883$\pm$ \tiny 0.008 \\
\midrule
MSE & 0.806$\pm$ \tiny 0.094 & 0.878$\pm$ \tiny 0.022 & \textbf{0.797$\pm$ \tiny 0.005} & 0.917$\pm$ \tiny 0.003 & 0.864$\pm$ \tiny 0.008 & 0.663$\pm$ \tiny 0.049 & 0.873$\pm$ \tiny 0.006 & 0.742$\pm$ \tiny 0.034 & 0.849$\pm$ \tiny 0.012 & 0.641$\pm$ \tiny 0.063 & 0.871$\pm$ \tiny 0.008 \\
\midrule
Cosine & 0.816$\pm$ \tiny 0.096 & 0.881$\pm$ \tiny 0.014 & 0.786$\pm$ \tiny 0.014 & 0.925$\pm$ \tiny 0.003 & 0.879$\pm$ \tiny 0.005 & 0.673$\pm$ \tiny 0.070 & \textbf{0.887$\pm$ \tiny 0.008} & 0.758$\pm$ \tiny 0.045 & 0.855$\pm$ \tiny 0.015 & 0.637$\pm$ \tiny 0.025 & \textbf{0.892$\pm$ \tiny 0.005} \\
\midrule
student-250k & \textbf{0.823$\pm$ \tiny 0.095} & 0.875$\pm$ \tiny 0.020 & 0.796$\pm$ \tiny 0.013 & \textbf{0.931$\pm$ \tiny 0.005} & \textbf{0.881$\pm$ \tiny 0.006} & \textbf{\underline{0.699$\pm$ \tiny 0.080}} & \textbf{\underline{0.893$\pm$ \tiny 0.005}} & \textbf{0.767$\pm$ \tiny 0.040} & \textbf{0.865$\pm$ \tiny 0.017} & \textbf{\underline{0.646$\pm$ \tiny 0.052}} & \textbf{\underline{0.893$\pm$ \tiny 0.009}} \\
student-2M & \textbf{\underline{0.825$\pm$ \tiny 0.096}} & 0.882$\pm$ \tiny 0.020 & \textbf{\underline{0.800$\pm$ \tiny 0.014}} & \textbf{\underline{0.933$\pm$ \tiny 0.006}} & \textbf{\underline{0.882$\pm$ \tiny 0.007}} & \textbf{0.697$\pm$ \tiny 0.093} & \textbf{\underline{0.893$\pm$ \tiny 0.002}} & 0.766$\pm$ \tiny 0.058 & \textbf{\underline{0.868$\pm$ \tiny 0.010}} & 0.639$\pm$ \tiny 0.054 & \textbf{0.892$\pm$ \tiny 0.004} \\
\end{tabular}
}
        }
    \end{minipage}
    \begin{minipage}{0.4\textwidth}
        \resizebox{0.94\textwidth}{!}{
            \rotatebox{90}{\begin{tabular}{r||cccc|cccccccc|}
{} & {\rotatebox{-90}{\shortstack{Absorption \\ Bioavailability}}} & {\rotatebox{-90}{\shortstack{Absorption \\ HIA}}} & {\rotatebox{-90}{\shortstack{Absorption \\ PAMPA}}} & {\rotatebox{-90}{\shortstack{Absorption \\ Pgp}}} & {\rotatebox{-90}{\shortstack{Tox \\ AMES}}} & {\rotatebox{-90}{\shortstack{Tox \\ Carcinogens}}} & {\rotatebox{-90}{\shortstack{Tox \\ ClinTox}}} & {\rotatebox{-90}{\shortstack{Tox \\ DILI}}} & {\rotatebox{-90}{\shortstack{Tox \\ Skin R}}} & {\rotatebox{-90}{\shortstack{Tox \\ Tox21}}} & {\rotatebox{-90}{\shortstack{Tox \\ hERG}}} & {\rotatebox{-90}{\shortstack{Tox \\ hERG (k)}}} \\
\midrule
InfoGraph & 0.631$\pm$ \tiny 0.015 & 0.872$\pm$ \tiny 0.085 & 0.685$\pm$ \tiny 0.031 & 0.896$\pm$ \tiny 0.022 & 0.853$\pm$ \tiny 0.009 & 0.728$\pm$ \tiny 0.042 & 0.621$\pm$ \tiny 0.086 & 0.837$\pm$ \tiny 0.056 & 0.714$\pm$ \tiny 0.030 & 0.770$\pm$ \tiny 0.058 & 0.778$\pm$ \tiny 0.027 & 0.849$\pm$ \tiny 0.009 \\
ChemGPT-1.2B & 0.668$\pm$ \tiny 0.046 & 0.859$\pm$ \tiny 0.055 & 0.665$\pm$ \tiny 0.050 & 0.926$\pm$ \tiny 0.027 & 0.843$\pm$ \tiny 0.012 & 0.785$\pm$ \tiny 0.017 & 0.641$\pm$ \tiny 0.022 & 0.857$\pm$ \tiny 0.031 & 0.721$\pm$ \tiny 0.077 & 0.762$\pm$ \tiny 0.066 & 0.789$\pm$ \tiny 0.049 & 0.867$\pm$ \tiny 0.007 \\
FRAD QM9${}^{(t)}$ & 0.626$\pm$ \tiny 0.022 & 0.945$\pm$ \tiny 0.034 & 0.699$\pm$ \tiny 0.043 & 0.914$\pm$ \tiny 0.024 & 0.871$\pm$ \tiny 0.009 & 0.772$\pm$ \tiny 0.057 & 0.553$\pm$ \tiny 0.054 & 0.843$\pm$ \tiny 0.044 & 0.713$\pm$ \tiny 0.043 & 0.797$\pm$ \tiny 0.060 & 0.817$\pm$ \tiny 0.032 & 0.873$\pm$ \tiny 0.004 \\
ChemBertMLM-10M & 0.664$\pm$ \tiny 0.069 & 0.892$\pm$ \tiny 0.082 & 0.715$\pm$ \tiny 0.056 & 0.911$\pm$ \tiny 0.029 & 0.858$\pm$ \tiny 0.013 & 0.776$\pm$ \tiny 0.063 & 0.648$\pm$ \tiny 0.082 & 0.791$\pm$ \tiny 0.031 & 0.747$\pm$ \tiny 0.057 & 0.789$\pm$ \tiny 0.055 & 0.779$\pm$ \tiny 0.017 & 0.867$\pm$ \tiny 0.007 \\
GROVER & 0.663$\pm$ \tiny 0.038 & 0.931$\pm$ \tiny 0.038 & 0.703$\pm$ \tiny 0.027 & 0.918$\pm$ \tiny 0.029 & 0.867$\pm$ \tiny 0.012 & 0.779$\pm$ \tiny 0.084 & 0.637$\pm$ \tiny 0.053 & 0.844$\pm$ \tiny 0.036 & 0.749$\pm$ \tiny 0.081 & 0.780$\pm$ \tiny 0.059 & 0.774$\pm$ \tiny 0.034 & 0.856$\pm$ \tiny 0.004 \\
GraphCL${}^{(t)}$ & 0.643$\pm$ \tiny 0.027 & 0.863$\pm$ \tiny 0.052 & 0.709$\pm$ \tiny 0.034 & 0.920$\pm$ \tiny 0.030 & 0.869$\pm$ \tiny 0.016 & \textbf{\underline{0.847$\pm$ \tiny 0.064}} & 0.639$\pm$ \tiny 0.078 & 0.827$\pm$ \tiny 0.035 & 0.770$\pm$ \tiny 0.038 & 0.787$\pm$ \tiny 0.058 & 0.799$\pm$ \tiny 0.038 & 0.864$\pm$ \tiny 0.005 \\
GraphLog${}^{(t)}$ & 0.622$\pm$ \tiny 0.071 & 0.897$\pm$ \tiny 0.035 & 0.637$\pm$ \tiny 0.024 & 0.920$\pm$ \tiny 0.026 & 0.869$\pm$ \tiny 0.006 & 0.793$\pm$ \tiny 0.076 & 0.696$\pm$ \tiny 0.081 & 0.853$\pm$ \tiny 0.035 & 0.751$\pm$ \tiny 0.101 & 0.801$\pm$ \tiny 0.052 & 0.797$\pm$ \tiny 0.049 & 0.849$\pm$ \tiny 0.006 \\
GraphMVP${}^{(t)}$ & \textbf{\underline{0.694$\pm$ \tiny 0.055}} & 0.944$\pm$ \tiny 0.025 & 0.718$\pm$ \tiny 0.009 & 0.918$\pm$ \tiny 0.038 & 0.874$\pm$ \tiny 0.011 & 0.779$\pm$ \tiny 0.095 & 0.624$\pm$ \tiny 0.046 & \textbf{0.867$\pm$ \tiny 0.049} & 0.750$\pm$ \tiny 0.087 & 0.793$\pm$ \tiny 0.055 & 0.823$\pm$ \tiny 0.045 & 0.872$\pm$ \tiny 0.005 \\
MolR gat & 0.672$\pm$ \tiny 0.049 & 0.957$\pm$ \tiny 0.020 & 0.705$\pm$ \tiny 0.061 & 0.928$\pm$ \tiny 0.028 & 0.871$\pm$ \tiny 0.014 & 0.760$\pm$ \tiny 0.057 & \textbf{0.810$\pm$ \tiny 0.048} & 0.858$\pm$ \tiny 0.042 & 0.748$\pm$ \tiny 0.047 & 0.800$\pm$ \tiny 0.061 & \textbf{\underline{0.844$\pm$ \tiny 0.022}} & 0.881$\pm$ \tiny 0.011 \\
ThreeDInfomax${}^{(t)}$ & 0.670$\pm$ \tiny 0.033 & \textbf{\underline{0.986$\pm$ \tiny 0.014}} & 0.745$\pm$ \tiny 0.026 & \textbf{0.929$\pm$ \tiny 0.030} & 0.872$\pm$ \tiny 0.018 & 0.791$\pm$ \tiny 0.074 & \textbf{\underline{0.837$\pm$ \tiny 0.043}} & 0.842$\pm$ \tiny 0.037 & \textbf{\underline{0.833$\pm$ \tiny 0.041}} & 0.804$\pm$ \tiny 0.059 & 0.829$\pm$ \tiny 0.035 & 0.874$\pm$ \tiny 0.010 \\
ChemBertMTR-77M${}^{(t)}$ & \textbf{0.683$\pm$ \tiny 0.027} & 0.960$\pm$ \tiny 0.034 & \textbf{\underline{0.763$\pm$ \tiny 0.026}} & \textbf{\underline{0.936$\pm$ \tiny 0.030}} & 0.881$\pm$ \tiny 0.005 & 0.776$\pm$ \tiny 0.033 & 0.734$\pm$ \tiny 0.068 & 0.858$\pm$ \tiny 0.037 & 0.758$\pm$ \tiny 0.069 & \textbf{0.818$\pm$ \tiny 0.063} & 0.832$\pm$ \tiny 0.034 & 0.897$\pm$ \tiny 0.009 \\
\midrule
MSE & 0.626$\pm$ \tiny 0.076 & 0.914$\pm$ \tiny 0.040 & 0.735$\pm$ \tiny 0.027 & 0.914$\pm$ \tiny 0.030 & 0.871$\pm$ \tiny 0.010 & 0.783$\pm$ \tiny 0.039 & 0.654$\pm$ \tiny 0.094 & 0.856$\pm$ \tiny 0.033 & 0.770$\pm$ \tiny 0.039 & 0.807$\pm$ \tiny 0.061 & 0.824$\pm$ \tiny 0.018 & 0.895$\pm$ \tiny 0.007 \\
\midrule
Cosine & 0.629$\pm$ \tiny 0.043 & 0.908$\pm$ \tiny 0.062 & \textbf{0.755$\pm$ \tiny 0.024} & 0.926$\pm$ \tiny 0.021 & 0.884$\pm$ \tiny 0.008 & 0.822$\pm$ \tiny 0.084 & 0.650$\pm$ \tiny 0.092 & \textbf{\underline{0.879$\pm$ \tiny 0.030}} & 0.780$\pm$ \tiny 0.033 & 0.814$\pm$ \tiny 0.056 & 0.830$\pm$ \tiny 0.038 & 0.908$\pm$ \tiny 0.006 \\
\midrule
student-250k & 0.671$\pm$ \tiny 0.043 & \textbf{0.969$\pm$ \tiny 0.018} & 0.742$\pm$ \tiny 0.025 & 0.926$\pm$ \tiny 0.025 & \textbf{0.888$\pm$ \tiny 0.009} & 0.810$\pm$ \tiny 0.079 & 0.633$\pm$ \tiny 0.082 & 0.858$\pm$ \tiny 0.036 & \textbf{0.812$\pm$ \tiny 0.040} & 0.817$\pm$ \tiny 0.062 & 0.828$\pm$ \tiny 0.050 & \textbf{\underline{0.912$\pm$ \tiny 0.006}} \\
student-2M & 0.653$\pm$ \tiny 0.055 & 0.959$\pm$ \tiny 0.026 & 0.730$\pm$ \tiny 0.024 & \textbf{\underline{0.936$\pm$ \tiny 0.024}} & \textbf{\underline{0.891$\pm$ \tiny 0.014}} & \textbf{0.839$\pm$ \tiny 0.095} & 0.662$\pm$ \tiny 0.072 & 0.856$\pm$ \tiny 0.045 & 0.801$\pm$ \tiny 0.026 & \textbf{\underline{0.819$\pm$ \tiny 0.054}} & \textbf{0.834$\pm$ \tiny 0.019} & \textbf{0.911$\pm$ \tiny 0.005} \\
\end{tabular}
}
        }
    \end{minipage}
    \label{tab:molecule_all_cls}
\end{table}

\begin{table}
    \centering
    \caption{$R^2$ score of each model on the ADMET downstream regression tasks. The best embedder for each task is highlighted in bold and underlined, and the second best is highlighted in bold.}
    \resizebox{\textwidth}{!}{\begin{tabular}{r||c|cccc|}
{} & {\rotatebox{-90}{\shortstack{  \\ avg}}} & {\rotatebox{-90}{\shortstack{Absorption \\ Caco2}}} & {\rotatebox{-90}{\shortstack{Absorption \\ FreeSolv}}} & {\rotatebox{-90}{\shortstack{Absorption \\ Lipophilicity}}} & {\rotatebox{-90}{\shortstack{Absorption \\ Solubility}}} \\
\midrule
InfoGraph & 0.275$\pm$ \tiny 0.284 & 0.491$\pm$ \tiny 0.031 & 0.639$\pm$ \tiny 0.058 & 0.341$\pm$ \tiny 0.035 & 0.700$\pm$ \tiny 0.007 \\
ChemBertMLM-10M & 0.264$\pm$ \tiny 0.364 & 0.543$\pm$ \tiny 0.076 & 0.776$\pm$ \tiny 0.038 & 0.363$\pm$ \tiny 0.063 & 0.774$\pm$ \tiny 0.007 \\
FRAD QM9${}^{(t)}$ & 0.332$\pm$ \tiny 0.284 & 0.564$\pm$ \tiny 0.051 & 0.686$\pm$ \tiny 0.082 & 0.483$\pm$ \tiny 0.029 & 0.758$\pm$ \tiny 0.011 \\
ChemGPT-1.2B & 0.340$\pm$ \tiny 0.329 & 0.567$\pm$ \tiny 0.079 & 0.831$\pm$ \tiny 0.048 & 0.487$\pm$ \tiny 0.020 & 0.798$\pm$ \tiny 0.009 \\
GROVER & 0.350$\pm$ \tiny 0.274 & 0.575$\pm$ \tiny 0.058 & 0.708$\pm$ \tiny 0.024 & 0.470$\pm$ \tiny 0.043 & 0.733$\pm$ \tiny 0.027 \\
GraphLog${}^{(t)}$ & 0.350$\pm$ \tiny 0.311 & 0.545$\pm$ \tiny 0.055 & 0.811$\pm$ \tiny 0.017 & 0.486$\pm$ \tiny 0.037 & 0.765$\pm$ \tiny 0.010 \\
GraphCL${}^{(t)}$ & 0.355$\pm$ \tiny 0.292 & 0.559$\pm$ \tiny 0.051 & 0.764$\pm$ \tiny 0.038 & 0.467$\pm$ \tiny 0.067 & 0.745$\pm$ \tiny 0.021 \\
GraphMVP${}^{(t)}$ & 0.397$\pm$ \tiny 0.320 & 0.592$\pm$ \tiny 0.064 & 0.861$\pm$ \tiny 0.036 & 0.590$\pm$ \tiny 0.064 & 0.791$\pm$ \tiny 0.009 \\
MolR gat & 0.394$\pm$ \tiny 0.307 & 0.651$\pm$ \tiny 0.089 & 0.804$\pm$ \tiny 0.075 & 0.518$\pm$ \tiny 0.037 & 0.822$\pm$ \tiny 0.010 \\
ThreeDInfomax${}^{(t)}$ & 0.425$\pm$ \tiny 0.322 & 0.700$\pm$ \tiny 0.038 & 0.852$\pm$ \tiny 0.055 & 0.624$\pm$ \tiny 0.031 & \textbf{\underline{0.848$\pm$ \tiny 0.004}} \\
ChemBertMTR-77M${}^{(t)}$ & 0.459$\pm$ \tiny 0.308 & \textbf{\underline{0.725$\pm$ \tiny 0.027}} & 0.874$\pm$ \tiny 0.037 & 0.670$\pm$ \tiny 0.025 & \textbf{0.839$\pm$ \tiny 0.007} \\
\midrule
MSE & 0.420$\pm$ \tiny 0.299 & 0.642$\pm$ \tiny 0.060 & 0.851$\pm$ \tiny 0.063 & 0.605$\pm$ \tiny 0.021 & 0.792$\pm$ \tiny 0.018 \\
\midrule
Cosine & 0.460$\pm$ \tiny 0.311 & 0.699$\pm$ \tiny 0.056 & \textbf{0.893$\pm$ \tiny 0.034} & 0.721$\pm$ \tiny 0.028 & 0.815$\pm$ \tiny 0.009 \\
\midrule
student-250k & \textbf{\underline{0.482$\pm$ \tiny 0.298}} & \textbf{0.712$\pm$ \tiny 0.040} & \textbf{\underline{0.900$\pm$ \tiny 0.035}} & \textbf{\underline{0.742$\pm$ \tiny 0.019}} & 0.823$\pm$ \tiny 0.007 \\
student-2M & \textbf{0.476$\pm$ \tiny 0.301} & 0.687$\pm$ \tiny 0.045 & 0.878$\pm$ \tiny 0.036 & \textbf{0.739$\pm$ \tiny 0.021} & 0.822$\pm$ \tiny 0.005 \\
\end{tabular}
}
    \vskip 0.5cm
    \resizebox{\textwidth}{!}{\begin{tabular}{r||cc|ccc|c|}
{} & {\rotatebox{-90}{\shortstack{Distribution \\ PPBR}}} & {\rotatebox{-90}{\shortstack{Distribution \\ VDss}}} & {\rotatebox{-90}{\shortstack{Excretion \\ Clearance (H)}}} & {\rotatebox{-90}{\shortstack{Excretion \\ Clearance (M)}}} & {\rotatebox{-90}{\shortstack{Excretion \\ Half Life}}} & {\rotatebox{-90}{\shortstack{Tox \\ LD50}}} \\
\midrule
InfoGraph & 0.093$\pm$ \tiny 0.073 & 0.018$\pm$ \tiny 0.190 & -0.048$\pm$ \tiny 0.133 & 0.070$\pm$ \tiny 0.046 & -0.011$\pm$ \tiny 0.161 & 0.458$\pm$ \tiny 0.039 \\
ChemBertMLM-10M & 0.112$\pm$ \tiny 0.035 & 0.066$\pm$ \tiny 0.091 & -0.185$\pm$ \tiny 0.122 & 0.040$\pm$ \tiny 0.178 & -0.240$\pm$ \tiny 0.279 & 0.390$\pm$ \tiny 0.044 \\
FRAD QM9${}^{(t)}$ & 0.180$\pm$ \tiny 0.031 & -0.004$\pm$ \tiny 0.050 & 0.006$\pm$ \tiny 0.095 & 0.124$\pm$ \tiny 0.059 & 0.104$\pm$ \tiny 0.129 & 0.415$\pm$ \tiny 0.039 \\
ChemGPT-1.2B & 0.175$\pm$ \tiny 0.036 & 0.046$\pm$ \tiny 0.173 & -0.018$\pm$ \tiny 0.071 & 0.117$\pm$ \tiny 0.099 & -0.047$\pm$ \tiny 0.182 & 0.442$\pm$ \tiny 0.043 \\
GROVER & 0.185$\pm$ \tiny 0.056 & \textbf{0.186$\pm$ \tiny 0.079} & -0.034$\pm$ \tiny 0.095 & 0.197$\pm$ \tiny 0.082 & 0.035$\pm$ \tiny 0.161 & 0.447$\pm$ \tiny 0.058 \\
GraphLog${}^{(t)}$ & 0.240$\pm$ \tiny 0.082 & \textbf{\underline{0.202$\pm$ \tiny 0.111}} & -0.094$\pm$ \tiny 0.053 & 0.068$\pm$ \tiny 0.120 & 0.018$\pm$ \tiny 0.192 & 0.457$\pm$ \tiny 0.054 \\
GraphCL${}^{(t)}$ & 0.237$\pm$ \tiny 0.048 & 0.158$\pm$ \tiny 0.075 & -0.022$\pm$ \tiny 0.127 & 0.123$\pm$ \tiny 0.108 & 0.007$\pm$ \tiny 0.165 & 0.508$\pm$ \tiny 0.026 \\
GraphMVP${}^{(t)}$ & 0.327$\pm$ \tiny 0.036 & 0.168$\pm$ \tiny 0.081 & -0.009$\pm$ \tiny 0.135 & 0.144$\pm$ \tiny 0.071 & -0.017$\pm$ \tiny 0.226 & 0.527$\pm$ \tiny 0.042 \\
MolR gat & 0.284$\pm$ \tiny 0.093 & 0.155$\pm$ \tiny 0.180 & -0.024$\pm$ \tiny 0.091 & 0.174$\pm$ \tiny 0.050 & 0.059$\pm$ \tiny 0.232 & 0.496$\pm$ \tiny 0.040 \\
ThreeDInfomax${}^{(t)}$ & 0.314$\pm$ \tiny 0.053 & 0.152$\pm$ \tiny 0.061 & 0.071$\pm$ \tiny 0.049 & 0.195$\pm$ \tiny 0.114 & -0.004$\pm$ \tiny 0.264 & 0.500$\pm$ \tiny 0.040 \\
ChemBertMTR-77M${}^{(t)}$ & \textbf{\underline{0.393$\pm$ \tiny 0.055}} & 0.138$\pm$ \tiny 0.127 & 0.011$\pm$ \tiny 0.048 & 0.250$\pm$ \tiny 0.078 & \textbf{0.196$\pm$ \tiny 0.190} & 0.491$\pm$ \tiny 0.031 \\
\midrule
MSE & 0.362$\pm$ \tiny 0.077 & 0.135$\pm$ \tiny 0.097 & 0.034$\pm$ \tiny 0.097 & 0.244$\pm$ \tiny 0.062 & 0.060$\pm$ \tiny 0.116 & 0.470$\pm$ \tiny 0.030 \\
\midrule
Cosine & 0.382$\pm$ \tiny 0.032 & 0.108$\pm$ \tiny 0.084 & \textbf{0.079$\pm$ \tiny 0.102} & 0.275$\pm$ \tiny 0.054 & 0.111$\pm$ \tiny 0.158 & 0.515$\pm$ \tiny 0.039 \\
\midrule
student-250k & \textbf{0.390$\pm$ \tiny 0.042} & 0.125$\pm$ \tiny 0.111 & \textbf{\underline{0.113$\pm$ \tiny 0.070}} & \textbf{0.283$\pm$ \tiny 0.076} & \textbf{\underline{0.207$\pm$ \tiny 0.101}} & \textbf{0.529$\pm$ \tiny 0.039} \\
student-2M & 0.389$\pm$ \tiny 0.050 & 0.138$\pm$ \tiny 0.115 & 0.069$\pm$ \tiny 0.060 & \textbf{\underline{0.348$\pm$ \tiny 0.062}} & 0.144$\pm$ \tiny 0.205 & \textbf{\underline{0.543$\pm$ \tiny 0.041}} \\
\end{tabular}
}
    \label{tab:molecule_all_reg}
\end{table}

\FloatBarrier

\FloatBarrier

    \newpage

\section{Natural Language Processing}

\subsection{Training set and hyperparameters}
\label{sec:appendix_training_params}

\subsubsection{Training set}
\label{sec:appendix_nlp_model_dataset_statistics}

\paragraph{Dataset sources.} We ran experiments with two training sets a home-made dataset combining different training sets of different embedders and the GISTEmbed dataset. We provide the statistics of our dataset in \autoref{tab:nlp_training_datasets_ours} and the GISTEmbed dataset is described in~\citep{solatorio2024gistembed}.

\paragraph{Dataset construction.} Most embedding datasets consists of positive and negative samples, questions and answers, or sentences and their labels. We flattened the datasets to have only one column of sentences and deduplicated the dataset. For the MEDI~\citep{} dataset for example, given query, positive and negative samples we build a dataset with three times the number of entries, one for each sentence. We then deduplicated the dataset to remove any duplicate entries.

 \begin{table}
\caption{Number of samples in each dataset}
\label{tab:nlp_training_datasets_ours}
\centering \resizebox{\textwidth}{!}{ \begin{tabular}{lr}
\toprule
 & Number of samples \\
URL &  \\
\midrule
\url{https://huggingface.co/datasets/embedding-data/SPECTER} & 190872 \\
\url{https://huggingface.co/datasets/embedding-data/Amazon-QA} & 3264474 \\
\url{https://huggingface.co/datasets/embedding-data/simple-wiki} & 203755 \\
\url{https://huggingface.co/datasets/embedding-data/QQP_triplets} & 328188 \\
\url{https://huggingface.co/datasets/embedding-data/sentence-compression} & 356409 \\
\url{https://huggingface.co/datasets/embedding-data/altlex} & 223901 \\
\url{https://huggingface.co/datasets/fancyzhx/ag_news} & 120000 \\
\url{https://huggingface.co/datasets/stanfordnlp/sst2} & 67349 \\
\url{https://huggingface.co/datasets/dair-ai/emotion} & 416809 \\
\url{https://huggingface.co/datasets/stanfordnlp/snli} & 1100304 \\
\url{https://huggingface.co/datasets/cardiffnlp/tweet_eval} & 45000 \\
\url{https://huggingface.co/datasets/stanfordnlp/imdb} & 25000 \\
 & 6342061 \\
\bottomrule
\end{tabular}
}
\end{table}

\subsubsection{Teachers and based students performance}
\label{sec:appendix_nlp_teachers_list}

\paragraph{Teachers.} We selected $4$ teachers from the MTEB benchmark~\citep{muennighoff2023mteb} as teachers for our distillation method. We provide the list of the teachers and their performance in \autoref{tab:nlp:base_student_teacher_perfs}. The $4$ teachers of widely different sizes ($335$M, $435$M and $7$B) have display strong but different performances on the MTEB benchmark.

 \begin{table}
\caption{Performance of the 4 teachers we used and of the base students. Experiments with single teacher distillation were performed with the stronger teacher SFR-Embedding-2\_R.}
\label{tab:nlp:base_student_teacher_perfs}
\resizebox{\textwidth}{!}{\begin{tabular}{llc|cccccccccccc|c}
\toprule
 &  & \rotatebox{90}{\shortstack{Size}} & \rotatebox{90}{\shortstack{Amazon \\ Counterfactual}} & \rotatebox{90}{\shortstack{Amazon \\ Polarity}} & \rotatebox{90}{\shortstack{Amazon \\ Reviews}} & \rotatebox{90}{\shortstack{Banking77}} & \rotatebox{90}{\shortstack{Emotion}} & \rotatebox{90}{\shortstack{Imdb}} & \rotatebox{90}{\shortstack{MTOPDomain}} & \rotatebox{90}{\shortstack{MTOPIntent}} & \rotatebox{90}{\shortstack{Massive \\ Intent}} & \rotatebox{90}{\shortstack{Massive \\ Scenario}} & \rotatebox{90}{\shortstack{Toxic \\ Conversations}} & \rotatebox{90}{\shortstack{Tweet \\ Sentiment \\ Extraction}} & \rotatebox{90}{\shortstack{Avg.}} \\
\midrule
\multirow[c]{4}{*}{Teacher} & SFR-Embedding-2\_R & 7111.0 & 92.7 & 97.3 & 61.0 & 90.0 & 93.4 & 96.8 & 98.6 & 91.3 & 86.0 & 90.6 & 91.1 & 79.7 & 89.0 \\
 & stella\_en\_400M\_v5 & 435.0 & 92.4 & 97.2 & 59.5 & 89.3 & 78.8 & 96.5 & 98.8 & 92.3 & 85.2 & 89.6 & 86.9 & 73.6 & 86.7 \\
 & UAE-Large-V1 & 335.0 & 75.5 & 92.8 & 48.3 & 87.7 & 51.8 & 92.8 & 94.0 & 76.9 & 76.5 & 79.8 & 71.1 & 59.8 & 75.6 \\
 & sf\_model\_e5 & 335.0 & 70.8 & 91.8 & 48.9 & 84.6 & 54.9 & 93.1 & 93.6 & 66.0 & 73.5 & 77.4 & 71.2 & 61.5 & 74.0 \\
\cline{1-16}
\multirow[c]{3}{*}{Student (Base)} & snowflake-arctic-embed-m & 109.0 & 76.8 & 82.8 & 38.9 & 80.3 & 46.5 & 74.1 & 92.7 & 65.2 & 66.9 & 72.8 & 64.9 & 56.7 & 68.2 \\
 & snowflake-arctic-embed-s & 33.0 & 71.2 & 78.8 & 38.3 & 79.1 & 45.8 & 69.5 & 90.9 & 58.6 & 64.8 & 70.0 & 62.0 & 58.9 & 65.7 \\
 & snowflake-arctic-embed-xs & 23.0 & 65.1 & 70.0 & 35.3 & 76.4 & 41.8 & 62.8 & 90.8 & 58.0 & 63.5 & 71.0 & 64.3 & 56.2 & 62.9 \\
\cline{1-16}
\bottomrule
\end{tabular}}

\end{table}

\subsubsection{Single teacher distillation}

\begin{wrapfigure}[19]{r}{0.4\textwidth}
\centering
\includegraphics[width=0.3\textwidth]{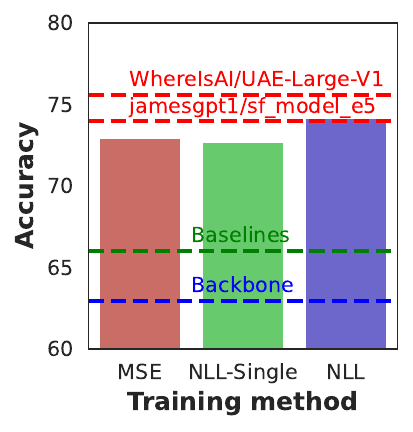}
\caption{Comparison of distilled small model with the performance of the initial backbone, baselines in the MTEB, with our teachers' performance.}
\label{fig:classification_against_single_16_30}
\end{wrapfigure}

\label{sec:appendix_nlp_single}
\paragraph{Single teacher vs. Multi-Teachers.} Since some teachers yield strong performance on their own, distilling only from the strongest could yield similar results as the multi-teacher setting involving weaker teachers.
We applied our method in a single-teacher setting using the strongest teacher by far (SF-Embeddings-R\_2) as a teacher and compared the results to the multi-teacher setting. 
Consistently with results in computer vision and molecular representations, we found that adding weaker teachers did improve our results (\autoref{fig:classification_against_single_16_30}), supporting our hypothesis that enforcing reconstruction capabilities for a diversity of models indeed leads to more informative representations.\vspace{-0.1cm}
\subsubsection{Hyperparameters}
\label{sec:appendix_nlp_hyperparameters}

\paragraph{Training hyperparameters.} We trained our models using the Adam optimizer with a constant learning rate of $5.10^{-5}$ and an effective batch size of $16$ for all our models.

\subsection{Detailed evaluation results}
\label{sec:appendix_nlp_detailed_results}

We ran different parts of the MTEB benchmarks and report the overall results for all our models in this section.

\subsubsection{Evaluation on classification tasks}
\label{sec:appendix_nlp_classification}

\paragraph{Small models' performance.} In \autoref{tab:mteb_classification_per_size_16_30} and \autoref{tab:mteb_classification_per_size_30_50}, we provide the classification accuracy of our distilled models on the MTEB classification benchmark for our smaller models xs ($22$M) and s ($33$M). Our smallest model significantly improves SOTA performance for models of its size by increasing the average score of $2$ points compared to the previous best model.
 \begin{table}
\caption{Performance of our distilled models compared to models of similar sizes 16M to 30M parameters from the MTEB Benchmark on classification tasks.}
\label{tab:mteb_classification_per_size_16_30}
\resizebox{\textwidth}{!}{\begin{tabular}{llc|cccccccccccc|c}
\toprule
 & Task & Size & \rotatebox{90}{\shortstack{Amazon \\ Counterfactual}} & \rotatebox{90}{\shortstack{Amazon \\ Polarity}} & \rotatebox{90}{\shortstack{Amazon \\ Reviews}} & \rotatebox{90}{\shortstack{Banking77}} & \rotatebox{90}{\shortstack{Emotion}} & \rotatebox{90}{\shortstack{Imdb}} & \rotatebox{90}{\shortstack{MTOPDomain}} & \rotatebox{90}{\shortstack{MTOPIntent}} & \rotatebox{90}{\shortstack{Massive \\ Intent}} & \rotatebox{90}{\shortstack{Massive \\ Scenario}} & \rotatebox{90}{\shortstack{Toxic \\ Conversations}} & \rotatebox{90}{\shortstack{Tweet \\ Sentiment \\ Extraction}} & Avg. \\
  & Model &  &  &  &  &  &  &  &  &  &  &  &  &  &  \\
\midrule
\multirow[c]{16}{*}{MTEB} & GIST & 23M & 72.9 & \bfseries \underline{87.2} & 42.6 & \underline{84.2} & 52.1 & 78.5 & \underline{94.8} & \underline{77.7} & 73.2 & 76.7 & \bfseries \underline{72.9} & 59.9 & 72.7 \\
 & Bulbasaur & 17M & 71.9 & 78.8 & 39.3 & 80.6 & 44.8 & 71.5 & 90.8 & 68.7 & 68.8 & 73.8 & 66.3 & 59.5 & 67.9 \\
 & Ivysaur & 23M & 72.1 & \underline{86.7} & \bfseries \underline{42.7} & 81.9 & 45.4 & 80.8 & 92.1 & 71.9 & 70.3 & 74.9 & 65.5 & 58.7 & 70.2 \\
 & Squirtle & 16M & 69.6 & 82.1 & 41.9 & 67.1 & 45.8 & 75.0 & 87.3 & 54.7 & 61.5 & 67.0 & 64.5 & \underline{61.8} & 64.9 \\
 & Venusaur & 16M & \underline{73.2} & 80.0 & 39.7 & 78.0 & 44.4 & 73.0 & 89.9 & 71.0 & 67.8 & 72.4 & 64.4 & 59.7 & 67.8 \\
 & Wartortle & 17M & 70.4 & 82.0 & 42.4 & 71.1 & 46.8 & 74.6 & 88.2 & 54.9 & 62.3 & 68.2 & 65.2 & \bfseries \underline{62.5} & 65.7 \\
 & gte-micro & 17M & 68.8 & 77.1 & 40.9 & 69.6 & 46.2 & 62.2 & 86.7 & 49.7 & 59.0 & 66.6 & 66.1 & 60.8 & 62.8 \\
 & gte-micro-v2 & 17M & 71.4 & 77.7 & 39.0 & 80.4 & 44.5 & 70.6 & 90.5 & 67.5 & 68.5 & 73.5 & 66.7 & 59.3 & 67.5 \\
 & gte-micro-v4 & 19M & 71.8 & 80.0 & 39.8 & 80.9 & 44.9 & 72.0 & 90.9 & 68.5 & 69.1 & 74.2 & 66.0 & 59.4 & 68.1 \\
 & snowflake-arctic-embed-xs & 23M & 65.1 & 70.0 & 35.3 & 76.4 & 41.8 & 62.8 & 90.8 & 58.0 & 63.5 & 71.0 & 64.3 & 56.2 & 62.9 \\
 & bge-micro & 17M & 66.3 & 75.4 & 35.8 & 80.6 & 42.5 & 70.7 & 90.2 & 68.0 & 67.8 & 73.0 & 69.2 & 56.7 & 66.3 \\
 & bge-micro-v2 & 17M & 67.8 & 79.8 & 37.5 & 81.2 & 44.5 & 76.5 & 90.7 & 68.3 & 68.6 & 73.9 & 70.2 & 57.6 & 68.0 \\
 & gte-tiny & 23M & 71.8 & 86.6 & \underline{42.6} & 81.7 & 44.7 & 80.5 & 91.8 & 69.9 & 70.1 & 74.9 & \underline{71.0} & 58.6 & 70.3 \\
 & slx-v0.1 & 23M & 61.5 & 64.3 & 30.3 & 80.0 & 40.5 & 61.8 & 92.0 & 63.3 & 67.9 & 73.9 & 62.1 & 54.0 & 62.6 \\
 & multi-qa-MiniLM-L6-cos-v1 & 23M & 61.8 & 62.4 & 29.6 & 78.6 & 39.6 & 61.2 & 90.0 & 59.6 & 66.8 & 73.8 & 65.1 & 51.6 & 61.7 \\
 & all-MiniLM-L6-v2 & 23M & 63.6 & 64.3 & 30.9 & 80.0 & 40.8 & 61.8 & 91.7 & 61.5 & 66.9 & 73.8 & 62.1 & 54.0 & 62.6 \\
\cline{1-16}
MSE & Student-xs & 23M & 71.6 & 86.2 & 42.3 & 83.6 & \underline{57.5} & \bfseries \underline{83.5} & 94.5 & 75.4 & \underline{74.3} & \bfseries \underline{80.4} & 66.3 & 59.3 & \underline{72.9} \\
\cline{1-16}
NLL & Student-xs & 23M & \bfseries \underline{76.5} & 84.9 & 42.4 & \bfseries \underline{85.8} & \bfseries \underline{58.0} & \underline{81.1} & \bfseries \underline{95.2} & \bfseries \underline{79.9} & \bfseries \underline{75.8} & \underline{80.4} & 68.1 & 60.1 & \bfseries \underline{74.0} \\
\cline{1-16}
\bottomrule
\end{tabular}}
\end{table}

 \begin{table}
\caption{Performance of our distilled models compared to models of similar sizes 30M to 50M parameters from the MTEB Benchmark on classification tasks.}
\label{tab:mteb_classification_per_size_30_50}
\resizebox{\textwidth}{!}{\begin{tabular}{llc|cccccccccccc|c}
\toprule
 & Task & Size & \rotatebox{90}{\shortstack{Amazon \\ Counterfactual}} & \rotatebox{90}{\shortstack{Amazon \\ Polarity}} & \rotatebox{90}{\shortstack{Amazon \\ Reviews}} & \rotatebox{90}{\shortstack{Banking77}} & \rotatebox{90}{\shortstack{Emotion}} & \rotatebox{90}{\shortstack{Imdb}} & \rotatebox{90}{\shortstack{MTOPDomain}} & \rotatebox{90}{\shortstack{MTOPIntent}} & \rotatebox{90}{\shortstack{Massive \\ Intent}} & \rotatebox{90}{\shortstack{Massive \\ Scenario}} & \rotatebox{90}{\shortstack{Toxic \\ Conversations}} & \rotatebox{90}{\shortstack{Tweet \\ Sentiment \\ Extraction}} & Avg. \\
  & Model &  &  &  &  &  &  &  &  &  &  &  &  &  &  \\
\midrule
\multirow[c]{12}{*}{MTEB} & bge-small-en-v1.5 & 33M & 73.8 & 92.8 & 47.0 & 85.7 & 47.8 & \bfseries \underline{90.6} & 93.4 & 74.8 & 74.8 & 78.7 & 69.9 & 60.5 & 74.1 \\
 & GIST & 33M & 75.3 & \underline{93.2} & \underline{49.7} & \underline{86.7} & 55.9 & 89.5 & \underline{95.5} & 79.1 & 75.5 & 79.2 & \bfseries \underline{72.8} & 61.0 & \bfseries \underline{76.1} \\
 & NoInstruct & 33M & 75.8 & \bfseries \underline{93.3} & \bfseries \underline{50.0} & 86.4 & 55.1 & \underline{90.2} & 95.3 & \underline{79.6} & \underline{76.0} & 79.3 & 69.4 & 61.3 & \underline{76.0} \\
 & snowflake-arctic-embed-s & 33M & 71.2 & 78.8 & 38.3 & 79.1 & 45.8 & 69.5 & 90.9 & 58.6 & 64.8 & 70.0 & 62.0 & 58.9 & 65.7 \\
 & bge-small-4096 & 35M & 68.8 & 81.3 & 38.6 & 80.0 & 40.1 & 80.1 & 90.4 & 66.5 & 67.6 & 73.5 & 69.3 & 57.6 & 67.8 \\
 & LASER & 43M & 76.8 & 61.0 & 28.7 & 57.8 & 24.8 & 57.6 & 75.4 & 49.5 & 47.9 & 55.9 & 54.0 & 48.7 & 53.2 \\
 & e5-small & 33M & 76.2 & 87.5 & 42.6 & 81.9 & 46.9 & 75.5 & 92.0 & 73.2 & 72.2 & 75.8 & \underline{72.8} & \bfseries \underline{63.3} & 71.7 \\
 & e5-small-v2 & 33M & \bfseries \underline{77.6} & 91.3 & 45.9 & 81.6 & 47.1 & 86.0 & 92.7 & 72.6 & 71.6 & 76.4 & 71.1 & \underline{61.5} & 72.9 \\
 & jina-embedding-s-en-v1 & 35M & 64.8 & 64.3 & 30.6 & 74.6 & 36.1 & 58.7 & 88.8 & 58.6 & 64.7 & 71.8 & 59.4 & 54.3 & 60.6 \\
 & jina-embeddings-v2-small-en & 33M & 71.4 & 82.9 & 40.9 & 78.2 & 44.0 & 73.6 & 94.0 & 72.5 & 67.6 & 69.8 & 71.5 & 59.4 & 68.8 \\
 & all-MiniLM-L12-v2 & 33M & 65.3 & 63.0 & 30.8 & 80.4 & 41.2 & 59.8 & 91.9 & 62.8 & 67.2 & 74.6 & 67.5 & 54.2 & 63.2 \\
 & gte-small & 33M & 73.2 & 91.8 & 48.0 & 84.1 & 46.6 & 86.8 & 93.0 & 69.7 & 70.3 & 75.6 & 70.3 & 58.2 & 72.3 \\
\cline{1-16}
MSE & Student-s & 33M & 72.6 & 90.3 & 44.3 & 84.2 & \underline{56.5} & 88.8 & 94.9 & 77.2 & 75.4 & \bfseries \underline{81.2} & 64.9 & 60.4 & 74.2 \\
\cline{1-16}
NLL & Student-s & 33M & \underline{77.3} & 89.2 & 43.8 & \bfseries \underline{86.7} & \bfseries \underline{58.0} & 88.3 & \bfseries \underline{95.5} & \bfseries \underline{81.9} & \bfseries \underline{76.7} & \underline{80.7} & 66.1 & 60.6 & 75.4 \\
\cline{1-16}
\bottomrule
\end{tabular}}
\end{table}

 \begin{table}
\caption{Performance of our distilled models compared to models of similar sizes 100M to 120M parameters from the MTEB Benchmark on classification tasks.}
\label{tab:mteb_classification_per_size_100_120}
\resizebox{\textwidth}{!}{\begin{tabular}{llc|cccccccccccc|c}
\toprule
 & Task & Size & \rotatebox{90}{\shortstack{Amazon \\ Counterfactual}} & \rotatebox{90}{\shortstack{Amazon \\ Polarity}} & \rotatebox{90}{\shortstack{Amazon \\ Reviews}} & \rotatebox{90}{\shortstack{Banking77}} & \rotatebox{90}{\shortstack{Emotion}} & \rotatebox{90}{\shortstack{Imdb}} & \rotatebox{90}{\shortstack{MTOPDomain}} & \rotatebox{90}{\shortstack{MTOPIntent}} & \rotatebox{90}{\shortstack{Massive \\ Intent}} & \rotatebox{90}{\shortstack{Massive \\ Scenario}} & \rotatebox{90}{\shortstack{Toxic \\ Conversations}} & \rotatebox{90}{\shortstack{Tweet \\ Sentiment \\ Extraction}} & Avg. \\
  & Model &  &  &  &  &  &  &  &  &  &  &  &  &  &  \\
\midrule
\multirow[c]{26}{*}{MTEB} & bge-base-en-v1.5 & 109M & 76.2 & \underline{93.4} & 48.9 & 87.0 & 51.9 & \bfseries \underline{90.8} & 94.2 & 76.9 & 76.2 & 80.2 & 71.6 & 59.4 & 75.5 \\
 & GIST & 109M & 76.0 & \bfseries \underline{93.5} & \bfseries \underline{50.5} & \underline{87.3} & 54.7 & \underline{89.7} & 95.3 & 78.1 & 76.0 & 79.6 & \underline{72.4} & 59.3 & \underline{76.0} \\
 & bilingual-embedding-small & 118M & 74.3 & 82.2 & 40.2 & 80.3 & 40.8 & 73.7 & 89.7 & 66.5 & 68.9 & 74.5 & 62.5 & 59.6 & 67.8 \\
 & multilingual-e5-small & 118M & 73.8 & 88.7 & 44.7 & 79.4 & 42.5 & 80.8 & 91.1 & 71.1 & 70.3 & 74.5 & 69.4 & 62.6 & 70.7 \\
 & snowflake-arctic-embed-m & 109M & 76.8 & 82.8 & 38.9 & 80.3 & 46.5 & 74.1 & 92.7 & 65.2 & 66.9 & 72.8 & 64.9 & 56.7 & 68.2 \\
 & snowflake-arctic-embed-m-v1.5 & 109M & 68.3 & 90.3 & 46.3 & 80.0 & 43.7 & 84.4 & 91.4 & 60.6 & 66.7 & 73.1 & 66.8 & 53.9 & 68.8 \\
 & ml-nlp-elser.html & 110M & 74.2 & 61.9 & 32.1 & 82.0 & 46.6 & 65.0 & 93.2 & 71.1 & 68.5 & 75.0 & 68.2 & 53.6 & 65.9 \\
 & e5-base-4k & 112M & 77.8 & 92.8 & 46.7 & 83.5 & 47.0 & 86.2 & 93.7 & 75.3 & 73.0 & 77.7 & 72.1 & 60.4 & 73.8 \\
 & instructor-base & 110M & \bfseries \underline{86.2} & 88.4 & 44.6 & 77.0 & 51.8 & 81.2 & 93.7 & 70.3 & 67.5 & 72.6 & 71.8 & \bfseries \underline{63.3} & 72.4 \\
 & bert-base-uncased & 110M & 74.2 & 71.3 & 33.6 & 63.4 & 35.3 & 65.3 & 82.6 & 68.1 & 59.9 & 64.3 & 70.0 & 51.8 & 61.7 \\
 & e5-base & 109M & \underline{79.7} & 88.0 & 42.6 & 83.3 & 49.4 & 76.0 & 93.2 & 74.8 & 72.2 & 76.8 & \bfseries \underline{74.1} & 61.4 & 72.6 \\
 & e5-base-v2 & 110M & 77.8 & 92.8 & 46.7 & 83.5 & 47.0 & 86.2 & 93.7 & 75.3 & 73.0 & 77.7 & 72.1 & 60.4 & 73.8 \\
 & jina-embedding-b-en-v1 & 110M & 66.7 & 67.6 & 31.2 & 84.1 & 44.7 & 63.9 & 91.5 & 72.8 & 71.1 & 76.2 & 66.2 & 56.9 & 66.1 \\
 & contriever-base-msmarco & 110M & 72.2 & 68.6 & 37.4 & 80.0 & 44.8 & 67.0 & 93.2 & 69.3 & 67.8 & 76.0 & 67.8 & 56.1 & 66.7 \\
 & sup-simcse-bert-base-uncased & 110M & 75.8 & 82.5 & 39.6 & 75.8 & 44.8 & 73.5 & 84.3 & 63.1 & 66.0 & 70.8 & 72.0 & 59.7 & 67.3 \\
 & unsup-simcse-bert-base-uncased & 110M & 67.1 & 74.5 & 33.9 & 73.5 & 42.2 & 69.6 & 81.7 & 59.2 & 59.8 & 66.2 & 68.8 & 53.4 & 62.5 \\
 & all-mpnet-base-v2 & 110M & 65.0 & 67.1 & 31.4 & 81.7 & 42.2 & 71.2 & 91.9 & 68.3 & 69.8 & 75.7 & 61.0 & 55.0 & 65.0 \\
 & allenai-specter & 110M & 58.7 & 57.8 & 26.3 & 66.7 & 24.8 & 56.4 & 74.5 & 50.0 & 51.7 & 58.6 & 57.4 & 45.5 & 52.4 \\
 & gtr-t5-base & 110M & 69.3 & 67.8 & 38.5 & 79.3 & 42.2 & 66.0 & 92.4 & 62.4 & 67.0 & 75.4 & 66.6 & 56.0 & 65.3 \\
 & msmarco-bert-co-condensor & 110M & 64.1 & 66.9 & 34.9 & 82.3 & 41.9 & 60.2 & 91.3 & 71.1 & 70.4 & 73.7 & 64.0 & 55.7 & 64.7 \\
 & paraphrase-multilingual-MiniLM-L12-v2 & 118M & 71.5 & 69.2 & 35.1 & 79.8 & 42.3 & 60.5 & 87.0 & 65.5 & 66.9 & 71.5 & 60.1 & 56.1 & 63.8 \\
 & sentence-t5-base & 110M & 75.8 & 85.1 & 44.9 & 76.5 & 51.4 & 77.3 & 90.3 & 63.3 & 69.7 & 72.3 & 68.2 & \underline{62.7} & 69.8 \\
 & text2vec-base-multilingual & 118M & 71.0 & 66.1 & 33.1 & 78.1 & 43.4 & 59.4 & 81.0 & 62.8 & 63.8 & 67.0 & 66.0 & 55.2 & 62.2 \\
 & Angle\_BERT & 109M & 77.9 & 76.0 & 37.2 & 75.5 & 45.2 & 68.8 & 85.4 & 64.5 & 66.3 & 70.6 & 67.1 & 57.6 & 66.0 \\
 & gte-base & 109M & 74.2 & 91.8 & \underline{49.0} & 85.1 & 48.6 & 86.0 & 93.0 & 72.0 & 71.5 & 76.4 & 71.6 & 57.0 & 73.0 \\
 & ALL\_862873 & 118M & 50.8 & 52.6 & 22.6 & 36.4 & 22.8 & 50.8 & 61.0 & 29.7 & 34.3 & 44.1 & 54.9 & 40.8 & 41.7 \\
\cline{1-16}
MSE & Student-m & 109M & 76.6 & 89.1 & 44.7 & 87.2 & \bfseries \underline{60.8} & 88.0 & \underline{95.7} & \underline{81.6} & \underline{77.7} & \underline{82.2} & 67.3 & 60.5 & 76.0 \\
\cline{1-16}
NLL & Student-m & 109M & 79.6 & 89.5 & 45.8 & \bfseries \underline{88.0} & \underline{59.7} & 88.3 & \bfseries \underline{96.2} & \bfseries \underline{83.9} & \bfseries \underline{78.6} & \bfseries \underline{82.7} & 67.1 & 61.3 & \bfseries \underline{76.7} \\
\cline{1-16}
\bottomrule
\end{tabular}}
\end{table}

 \begin{table}
\caption{Performance of our distilled models compared to models of similar sizes 200M to 420M parameters from the MTEB Benchmark on classification tasks.}
\label{tab:mteb_classification_per_size_200_420}
\resizebox{\textwidth}{!}{\begin{tabular}{llc|cccccccccccc|c}
\toprule
 & Task & Size & \rotatebox{90}{\shortstack{Amazon \\ Counterfactual}} & \rotatebox{90}{\shortstack{Amazon \\ Polarity}} & \rotatebox{90}{\shortstack{Amazon \\ Reviews}} & \rotatebox{90}{\shortstack{Banking77}} & \rotatebox{90}{\shortstack{Emotion}} & \rotatebox{90}{\shortstack{Imdb}} & \rotatebox{90}{\shortstack{MTOPDomain}} & \rotatebox{90}{\shortstack{MTOPIntent}} & \rotatebox{90}{\shortstack{Massive \\ Intent}} & \rotatebox{90}{\shortstack{Massive \\ Scenario}} & \rotatebox{90}{\shortstack{Toxic \\ Conversations}} & \rotatebox{90}{\shortstack{Tweet \\ Sentiment \\ Extraction}} & Avg. \\
  & Model &  &  &  &  &  &  &  &  &  &  &  &  &  &  \\
\midrule
\multirow[c]{20}{*}{MTEB} & gte-multilingual-base & 305M & 76.0 & 80.7 & 43.6 & 85.4 & 48.0 & 74.9 & 92.5 & 72.6 & 72.1 & 76.3 & 71.0 & 57.6 & 70.9 \\
 & bge-large-en-v1.5 & 335M & 75.8 & 92.4 & 48.2 & 87.8 & 51.5 & 92.8 & 94.6 & 79.5 & \bfseries \underline{77.6} & 80.5 & 70.9 & 59.9 & 76.0 \\
 & GIST & 335M & 75.6 & 93.4 & 49.1 & \bfseries \underline{88.1} & 54.7 & 91.2 & \underline{95.2} & 78.2 & 76.2 & 79.3 & \bfseries \underline{71.9} & 59.2 & \underline{76.0} \\
 & MUG-B-1.6 & 335M & 72.4 & 93.7 & \bfseries \underline{50.9} & 85.4 & 55.9 & \bfseries \underline{93.6} & 94.2 & 67.5 & 73.9 & 77.4 & 67.3 & 61.8 & 74.5 \\
 & bilingual-embedding-base & 278M & 77.4 & 89.5 & 46.1 & 78.5 & 47.1 & 87.4 & 92.9 & 64.8 & 68.9 & 75.2 & 63.4 & 62.5 & 71.1 \\
 & snowflake-arctic-embed-l & 334M & 74.8 & 78.4 & 36.7 & 80.1 & 46.5 & 72.9 & 92.6 & 64.5 & 65.8 & 71.1 & 64.7 & 56.7 & 67.1 \\
 & UAE-Large-V1 & 335M & 75.5 & 92.8 & 48.3 & 87.7 & 51.8 & 92.8 & 94.0 & 76.9 & 76.5 & 79.8 & 71.1 & 59.8 & 75.6 \\
 & embedder-100p & 278M & 67.1 & 70.4 & 33.2 & 82.7 & 43.5 & 67.3 & 91.8 & 74.7 & 71.8 & 77.8 & 67.5 & 55.6 & 67.0 \\
 & instructor-large & 335M & \bfseries \underline{88.1} & 91.5 & 47.9 & 78.5 & 52.7 & 88.3 & 93.9 & 68.0 & 68.9 & 73.3 & 71.0 & \bfseries \underline{64.1} & 73.9 \\
 & e5-large & 335M & 77.7 & 90.0 & 43.0 & 84.1 & 48.0 & 82.1 & 93.9 & 76.4 & 73.2 & 77.4 & 70.6 & 61.2 & 73.1 \\
 & e5-large-v2 & 335M & 79.2 & \underline{93.8} & 48.6 & 84.5 & 49.5 & 91.7 & 94.6 & 77.1 & 73.8 & 78.1 & 70.9 & 60.9 & 75.2 \\
 & multilingual-e5-base & 278M & 77.4 & 91.8 & 47.5 & 73.5 & 45.7 & 84.3 & 90.9 & 61.6 & 65.7 & 71.6 & 64.3 & \underline{62.8} & 69.8 \\
 & sf\_model\_e5 & 335M & 70.8 & 91.8 & 48.9 & 84.6 & 54.9 & \underline{93.1} & 93.6 & 66.0 & 73.5 & 77.4 & 71.2 & 61.5 & 74.0 \\
 & jina-embedding-l-en-v1 & 335M & 68.9 & 69.1 & 31.4 & 85.3 & 45.8 & 66.4 & 92.8 & 76.1 & 72.7 & 77.1 & 69.1 & 58.2 & 67.8 \\
 & ember-v1 & 335M & 76.1 & 92.0 & 47.9 & 87.9 & 52.0 & 92.8 & 94.6 & 79.3 & 77.4 & 80.5 & 71.4 & 60.0 & 76.0 \\
 & mxbai-embed-2d-large-v1 & 335M & 74.8 & 93.3 & 46.2 & 86.7 & 49.3 & 90.4 & 93.1 & 73.2 & 73.9 & 78.2 & \underline{71.5} & 59.2 & 74.1 \\
 & mxbai-embed-large-v1 & 335M & 75.0 & \bfseries \underline{93.8} & \underline{49.2} & 87.8 & 50.9 & 92.8 & 94.0 & 76.8 & 76.2 & 80.0 & 71.5 & 59.7 & 75.6 \\
 & paraphrase-multilingual-mpnet-base-v2 & 278M & 75.8 & 76.4 & 38.5 & 81.1 & 45.8 & 64.6 & 89.2 & 68.7 & 69.3 & 75.3 & 71.0 & 59.0 & 67.9 \\
 & gte-large & 335M & 72.6 & 92.5 & 49.1 & 86.1 & 47.9 & 88.5 & 93.5 & 73.2 & 72.6 & 76.8 & 70.6 & 56.6 & 73.3 \\
 & b1ade-embed & 335M & 75.2 & 93.1 & 48.4 & \underline{88.0} & 51.9 & 91.9 & 94.3 & 76.6 & 75.9 & 79.4 & 67.9 & 59.2 & 75.2 \\
\cline{1-16}
MSE & Student-l & 335M & 77.3 & 84.5 & 43.4 & 86.0 & \underline{60.0} & 82.7 & 95.1 & \underline{79.8} & 76.3 & \underline{81.3} & 65.8 & 60.2 & 74.4 \\
\cline{1-16}
NLL & Student-l & 335M & \underline{81.5} & 88.1 & 45.9 & 86.9 & \bfseries \underline{60.4} & 88.2 & \bfseries \underline{95.6} & \bfseries \underline{83.2} & \underline{77.5} & \bfseries \underline{81.4} & 67.7 & 62.2 & \bfseries \underline{76.5} \\
\cline{1-16}
\bottomrule
\end{tabular}}
\end{table}

\subsubsection{Evaluation on similarity and clustering tasks}
\label{sec:appendix_nlp_similarity_clustering}

\paragraph{Limited structure of our embedding spaces.} Our method only seeks to pack as much (statistical) information into the embeddings as possible without any constraints on the underlying structure of the embedding space. It is therefore not surprising that methods that relies on metrics on the embedding space such as similarity tasks do not perform as well as the classification tasks. However, our embedder are still competitive on these tasks achieving average performance for their respective size categories.

\paragraph{Clustering with very small model.} In \autoref{tab:mteb_clustering_per_size_16_30}, we show that our very small model actually outperforms baselines and sits on the pareto frontier for clustering tasks. This is a surprising result as we did not optimize our models for clustering tasks and the embeddings are not designed to have a meaningful structure.

\begin{table}
\caption{Performance of our distilled models compared of models of similar sizes 16M to 30M parameters from the MTEB Benchmark on clustering tasks.}
\label{tab:mteb_clustering_per_size_16_30}
\resizebox{\textwidth}{!}{\begin{tabular}{llc|ccccccc|c}
\toprule
 & Task & Size & \rotatebox{90}{\shortstack{Arxiv \\ Clustering \\ P2P}} & \rotatebox{90}{\shortstack{Arxiv \\ Clustering \\ S2S}} & \rotatebox{90}{\shortstack{Reddit \\ Clustering \\ P2P}} & \rotatebox{90}{\shortstack{Reddit \\ Clustering}} & \rotatebox{90}{\shortstack{Stack \\ Exchange \\ Clustering \\ P2P}} & \rotatebox{90}{\shortstack{Stack \\ Exchange \\ Clustering}} & \rotatebox{90}{\shortstack{Twenty \\ Newsgroups \\ Clustering}} & Avg. \\
  & Model &  &  &  &  &  &  &  &  &  \\
\midrule
\multirow[c]{16}{*}{MTEB} & Bulbasaur & 17M & 40.3 & 31.1 & 51.4 & 45.9 & 30.7 & 52.2 & 39.4 & 41.6 \\
 & Ivysaur & 23M & 46.4 & 35.4 & 56.0 & 47.5 & 33.6 & 53.9 & 40.8 & 44.8 \\
 & Squirtle & 16M & 33.0 & 24.7 & 43.7 & 31.4 & 29.2 & 39.2 & 28.2 & 32.8 \\
 & Venusaur & 16M & 31.8 & 21.1 & 44.1 & 26.7 & 27.5 & 32.8 & 26.1 & 30.0 \\
 & Wartortle & 17M & 35.8 & 27.3 & 46.1 & 35.9 & 29.9 & 45.3 & 31.7 & 36.0 \\
 & gte-micro & 17M & 35.2 & 31.1 & 47.9 & 45.6 & 30.1 & 52.6 & 40.8 & 40.5 \\
 & gte-micro-v4 & 19M & 42.9 & 32.5 & 53.6 & 48.3 & 31.9 & 55.1 & 41.4 & 43.6 \\
 & snowflake-arctic-embed-xs & 23M & 43.5 & 32.1 & \underline{57.8} & 48.3 & 34.6 & 57.5 & 36.3 & 44.3 \\
 & bge-micro & 17M & 44.6 & 34.5 & 54.5 & 45.3 & \underline{34.7} & 53.1 & 39.4 & 43.7 \\
 & bge-micro-v2 & 17M & 44.5 & 33.2 & 55.2 & 45.5 & 34.1 & 54.5 & 40.2 & 43.9 \\
 & gte-tiny & 23M & \bfseries \underline{46.6} & 36.0 & 56.5 & 50.2 & \bfseries \underline{35.7} & \underline{57.5} & 43.3 & \underline{46.6} \\
 & GIST-all-MiniLM-L6-v2 & 23M & 45.3 & 35.5 & 48.7 & 44.1 & 33.9 & 53.1 & 41.1 & 43.1 \\
 & slx-v0.1 & 23M & 46.5 & \underline{37.7} & 54.8 & 50.7 & 34.2 & 53.1 & \bfseries \underline{46.5} & 46.2 \\
 & multi-qa-MiniLM-L6-cos-v1 & 23M & 37.8 & 27.7 & 51.0 & 46.3 & 33.4 & 48.1 & 40.8 & 40.7 \\
 & all-MiniLM-L6-v2 & 23M & \underline{46.5} & \bfseries \underline{37.9} & 54.8 & \underline{50.7} & 34.3 & 53.1 & \bfseries \underline{46.5} & 46.3 \\
 & rubert-tiny-turbo & 29M & 24.8 & 16.7 & 40.5 & 26.3 & 28.0 & 33.5 & 19.9 & 27.1 \\
\cline{1-11}
MSE & Student-xs & 23M & 42.4 & 30.9 & 55.2 & 49.2 & 32.7 & 53.5 & 41.9 & 43.7 \\
\cline{1-11}
NLL & Student-xs & 23M & 45.2 & 33.9 & \bfseries \underline{58.1} & \bfseries \underline{52.1} & 33.1 & \bfseries \underline{59.9} & 44.3 & \bfseries \underline{46.7} \\
\cline{1-11}
\bottomrule
\end{tabular}}
\end{table}

\begin{table}
\caption{Performance of our distilled models compared of models of similar sizes 30M to 50M parameters from the MTEB Benchmark on clustering tasks.}
\label{tab:mteb_clustering_per_size_30_50}
\resizebox{\textwidth}{!}{\begin{tabular}{llc|ccccccc|c}
\toprule
 & Task & Size & \rotatebox{90}{\shortstack{Arxiv \\ Clustering \\ P2P}} & \rotatebox{90}{\shortstack{Arxiv \\ Clustering \\ S2S}} & \rotatebox{90}{\shortstack{Reddit \\ Clustering \\ P2P}} & \rotatebox{90}{\shortstack{Reddit \\ Clustering}} & \rotatebox{90}{\shortstack{Stack \\ Exchange \\ Clustering \\ P2P}} & \rotatebox{90}{\shortstack{Stack \\ Exchange \\ Clustering}} & \rotatebox{90}{\shortstack{Twenty \\ Newsgroups \\ Clustering}} & Avg. \\
  & Model &  &  &  &  &  &  &  &  &  \\
\midrule
\multirow[c]{11}{*}{MTEB} & bge-small-en-v1.5 & 33M & 47.4 & 40.0 & 60.6 & 52.3 & 35.3 & 60.8 & 48.5 & 49.3 \\
 & snowflake-arctic-embed-s & 33M & 44.9 & 35.9 & 60.5 & 50.5 & 34.0 & 60.7 & 38.3 & 46.4 \\
 & bge-small-4096 & 35M & 43.9 & 29.6 & 54.3 & 43.7 & 33.3 & 51.8 & 36.6 & 41.9 \\
 & GIST-small-Embedding-v0 & 33M & 47.6 & 39.9 & 60.6 & \underline{55.5} & 36.2 & 61.9 & \bfseries \underline{50.0} & 50.2 \\
 & NoInstruct-small-Embedding-v0 & 33M & \underline{47.8} & \underline{40.1} & \underline{61.2} & 55.4 & \bfseries \underline{36.6} & \underline{62.0} & 49.9 & \underline{50.4} \\
 & e5-small & 33M & 44.1 & 37.1 & 57.2 & 43.3 & 30.8 & 59.6 & 37.6 & 44.3 \\
 & e5-small-v2 & 33M & 42.1 & 34.8 & 59.7 & 45.7 & 32.0 & 58.5 & 41.1 & 44.8 \\
 & jina-embedding-s-en-v1 & 35M & 34.2 & 24.0 & 49.9 & 38.0 & 31.5 & 46.4 & 34.4 & 36.9 \\
 & jina-embeddings-v2-small-en & 33M & 44.0 & 35.2 & 57.1 & 49.3 & 34.4 & 55.4 & 41.6 & 45.3 \\
 & all-MiniLM-L12-v2 & 33M & 46.1 & 37.5 & 54.8 & 51.2 & 33.1 & 53.0 & 47.5 & 46.2 \\
 & gte-small & 33M & \bfseries \underline{47.9} & \bfseries \underline{40.3} & \bfseries \underline{61.4} & \bfseries \underline{55.6} & \underline{36.3} & \bfseries \underline{62.6} & \underline{50.0} & \bfseries \underline{50.6} \\
\cline{1-11}
MSE & Student-s & 33M & 43.1 & 33.3 & 57.1 & 50.8 & 32.3 & 55.7 & 42.8 & 45.0 \\
\cline{1-11}
NLL & Student-s & 33M & 45.9 & 35.2 & 60.3 & 51.9 & 32.3 & 61.5 & 45.1 & 47.4 \\
\cline{1-11}
\bottomrule
\end{tabular}}
\end{table}

\begin{table}
\caption{Performance of our distilled models compared of models of similar sizes 100M to 120M parameters from the MTEB Benchmark on clustering tasks.}
\label{tab:mteb_clustering_per_size_100_120}
\resizebox{\textwidth}{!}{\begin{tabular}{llc|ccccccc|c}
\toprule
 & Task & Size & \rotatebox{90}{\shortstack{Arxiv \\ Clustering \\ P2P}} & \rotatebox{90}{\shortstack{Arxiv \\ Clustering \\ S2S}} & \rotatebox{90}{\shortstack{Reddit \\ Clustering \\ P2P}} & \rotatebox{90}{\shortstack{Reddit \\ Clustering}} & \rotatebox{90}{\shortstack{Stack \\ Exchange \\ Clustering \\ P2P}} & \rotatebox{90}{\shortstack{Stack \\ Exchange \\ Clustering}} & \rotatebox{90}{\shortstack{Twenty \\ Newsgroups \\ Clustering}} & Avg. \\
  & Model &  &  &  &  &  &  &  &  &  \\
\midrule
\multirow[c]{26}{*}{MTEB} & bge-base-en-v1.5 & 109M & \bfseries \underline{48.8} & \underline{42.8} & 62.7 & 56.6 & 35.2 & 66.1 & 50.8 & 51.8 \\
 & bilingual-embedding-small & 118M & 41.8 & 31.6 & 58.4 & 47.4 & 33.6 & 52.5 & 40.5 & 43.7 \\
 & multilingual-e5-small & 118M & 39.2 & 30.8 & 59.0 & 39.1 & 32.1 & 53.5 & 33.2 & 41.0 \\
 & snowflake-arctic-embed-m & 109M & 47.2 & 37.4 & 62.8 & 47.5 & \bfseries \underline{39.4} & 59.5 & 37.7 & 47.4 \\
 & snowflake-arctic-embed-m-v1.5 & 109M & 45.0 & 34.1 & 61.8 & 51.9 & 33.8 & 61.2 & 38.1 & 46.6 \\
 & GIST-Embedding-v0 & 109M & 48.3 & 42.7 & 62.4 & 59.1 & 35.6 & \underline{66.1} & \underline{52.2} & \underline{52.4} \\
 & ml-nlp-elser.html & 110M & 35.3 & 23.2 & 51.9 & 38.7 & 28.7 & 42.7 & 27.8 & 35.5 \\
 & e5-base-4k & 112M & 46.1 & 39.7 & \bfseries \underline{63.4} & 56.2 & 32.5 & 65.2 & 48.2 & 50.2 \\
 & instructor-base & 110M & 39.7 & 29.2 & 63.2 & \underline{59.3} & 35.3 & 65.0 & 51.3 & 49.0 \\
 & bert-base-uncased & 110M & 35.2 & 27.5 & 43.3 & 27.2 & 26.6 & 43.6 & 23.4 & 32.4 \\
 & e5-base & 109M & 44.6 & 40.5 & 62.2 & 48.2 & 32.6 & 63.9 & 42.6 & 47.8 \\
 & e5-base-v2 & 110M & 46.1 & 39.7 & \underline{63.2} & 56.5 & 33.0 & 64.6 & 49.9 & 50.4 \\
 & jina-embedding-b-en-v1 & 110M & 39.2 & 29.1 & 52.5 & 42.9 & 31.4 & 48.1 & 38.1 & 40.2 \\
 & contriever-base-msmarco & 110M & 42.6 & 32.3 & 57.6 & 54.9 & 32.2 & 63.1 & 46.8 & 47.1 \\
 & sup-simcse-bert-base-uncased & 110M & 35.2 & 27.5 & 47.7 & 40.2 & 29.4 & 47.5 & 34.9 & 37.5 \\
 & unsup-simcse-bert-base-uncased & 110M & 32.6 & 24.7 & 45.1 & 32.2 & 28.5 & 43.1 & 23.2 & 32.8 \\
 & all-mpnet-base-v2 & 110M & 48.4 & 39.7 & 56.8 & 54.8 & 34.3 & 53.8 & 49.7 & 48.2 \\
 & allenai-specter & 110M & 44.8 & 35.3 & 35.1 & 24.1 & 31.5 & 39.0 & 24.2 & 33.4 \\
 & gtr-t5-base & 110M & 35.5 & 27.2 & 58.5 & 56.1 & 33.0 & 64.2 & 46.7 & 45.9 \\
 & msmarco-bert-co-condensor & 110M & 36.9 & 29.0 & 53.5 & 48.0 & 30.5 & 59.5 & 38.7 & 42.3 \\
 & paraphrase-multilingual-MiniLM-L12-v2 & 118M & 38.3 & 31.6 & 50.1 & 42.6 & 31.7 & 49.3 & 40.0 & 40.5 \\
 & sentence-t5-base & 110M & 39.3 & 27.3 & 59.7 & 52.9 & 35.7 & 63.1 & 48.1 & 46.6 \\
 & text2vec-base-multilingual & 118M & 32.3 & 25.5 & 43.3 & 31.2 & 30.6 & 34.4 & 31.6 & 32.7 \\
 & Angle\_BERT & 109M & 35.3 & 27.7 & 46.0 & 40.3 & 28.9 & 48.3 & 33.1 & 37.1 \\
 & gte-base & 109M & \underline{48.6} & \bfseries \underline{43.0} & 62.6 & \bfseries \underline{59.3} & \underline{36.0} & \bfseries \underline{66.6} & \bfseries \underline{52.3} & \bfseries \underline{52.6} \\
 & ALL\_862873 & 118M & 14.8 & 12.2 & 27.1 & 18.4 & 27.3 & 23.7 & 20.2 & 20.5 \\
\cline{1-11}
MSE & Student-m & 109M & 46.5 & 37.1 & 60.4 & 54.5 & 33.4 & 62.0 & 46.1 & 48.6 \\
\cline{1-11}
NLL & Student-m & 109M & 47.7 & 38.7 & 61.5 & 56.3 & 33.8 & 64.7 & 46.6 & 49.9 \\
\cline{1-11}
\bottomrule
\end{tabular}}
\end{table}

\begin{table}
\caption{Performance of our distilled models compared of models of similar sizes 16M to 30M parameters from the MTEB Benchmark on STS tasks.}
\label{tab:mteb_sts_per_size_16_30}
\resizebox{\textwidth}{!}{\begin{tabular}{llc|cccccccccc|c}
\toprule
 & Task & Size & \rotatebox{90}{\shortstack{BIOSSES}} & \rotatebox{90}{\shortstack{SICK-R}} & \rotatebox{90}{\shortstack{STS12}} & \rotatebox{90}{\shortstack{STS13}} & \rotatebox{90}{\shortstack{STS14}} & \rotatebox{90}{\shortstack{STS15}} & \rotatebox{90}{\shortstack{STS16}} & \rotatebox{90}{\shortstack{STS17}} & \rotatebox{90}{\shortstack{STS22}} & \rotatebox{90}{\shortstack{STSBenchmark}} & Avg. \\
  & Model &  &  &  &  &  &  &  &  &  &  &  &  \\
\midrule
\multirow[c]{12}{*}{MTEB} & Bulbasaur & 17M & 85.0 & 76.0 & 69.5 & 81.0 & 77.1 & 85.4 & 82.3 & 88.0 & 64.1 & 83.3 & 79.2 \\
 & Ivysaur & 23M & \bfseries \underline{87.3} & 75.6 & 68.6 & 80.5 & 77.6 & 86.2 & 82.8 & \underline{88.6} & \underline{67.4} & 84.2 & 79.9 \\
 & Squirtle & 16M & 71.8 & 77.3 & 70.2 & 78.4 & 74.8 & 82.0 & 78.3 & 85.8 & 61.2 & 79.2 & 75.9 \\
 & Venusaur & 16M & 77.6 & 74.7 & 54.4 & 74.2 & 70.0 & 75.7 & 73.7 & 84.8 & 62.6 & 76.7 & 72.4 \\
 & Wartortle & 17M & 80.8 & 78.2 & \bfseries \underline{75.2} & 79.3 & 76.6 & 84.7 & 81.4 & 86.6 & 63.4 & 81.8 & 78.8 \\
 & snowflake-arctic-embed-xs & 23M & 84.0 & 69.3 & 65.9 & 77.9 & 72.8 & 83.5 & 80.6 & 84.5 & 66.3 & 79.2 & 76.4 \\
 & bge-micro & 17M & 83.4 & 72.4 & 71.9 & 80.9 & 76.6 & 84.9 & 80.7 & 85.6 & 65.9 & 81.3 & 78.4 \\
 & bge-micro-v2 & 17M & 82.9 & 73.6 & 71.9 & 79.8 & 76.9 & 84.8 & 81.9 & 86.8 & 65.4 & 82.5 & 78.7 \\
 & gte-tiny & 23M & \underline{86.6} & 75.8 & 72.6 & \underline{82.4} & \underline{78.0} & \underline{86.5} & \bfseries \underline{83.3} & 88.3 & 66.7 & \underline{84.4} & \underline{80.5} \\
 & GIST-all-MiniLM-L6-v2 & 23M & 81.3 & \underline{79.1} & \underline{75.0} & \bfseries \underline{83.3} & \bfseries \underline{78.6} & \bfseries \underline{87.0} & \underline{83.0} & 87.4 & \bfseries \underline{68.1} & \bfseries \underline{84.4} & \bfseries \underline{80.7} \\
 & multi-qa-MiniLM-L6-cos-v1 & 23M & 79.8 & 70.0 & 64.4 & 76.4 & 69.3 & 80.2 & 79.6 & 81.2 & 65.5 & 76.0 & 74.2 \\
 & all-MiniLM-L6-v2 & 23M & 81.6 & 77.6 & 72.4 & 80.6 & 75.6 & 85.4 & 79.0 & 87.6 & 67.2 & 82.0 & 78.9 \\
\cline{1-14}
MSE & Student-xs & 23M & 76.8 & \bfseries \underline{79.2} & 72.2 & 80.3 & 75.9 & 85.0 & 83.0 & 87.1 & 66.4 & 82.9 & 78.9 \\
\cline{1-14}
NLL & Student-xs & 23M & 78.8 & 77.8 & 71.6 & 80.2 & 77.0 & 85.8 & 82.8 & \bfseries \underline{89.3} & 65.8 & 83.5 & 79.3 \\
\cline{1-14}
\bottomrule
\end{tabular}}
\end{table}

\begin{table}
\caption{Performance of our distilled models compared of models of similar sizes 30M to 50M parameters from the MTEB Benchmark on STS tasks.}
\label{tab:mteb_sts_per_size_30_50}
\resizebox{\textwidth}{!}{\begin{tabular}{llc|cccccccccc|c}
\toprule
 & Task & Size & \rotatebox{90}{\shortstack{BIOSSES}} & \rotatebox{90}{\shortstack{SICK-R}} & \rotatebox{90}{\shortstack{STS12}} & \rotatebox{90}{\shortstack{STS13}} & \rotatebox{90}{\shortstack{STS14}} & \rotatebox{90}{\shortstack{STS15}} & \rotatebox{90}{\shortstack{STS16}} & \rotatebox{90}{\shortstack{STS17}} & \rotatebox{90}{\shortstack{STS22}} & \rotatebox{90}{\shortstack{STSBenchmark}} & Avg. \\
  & Model &  &  &  &  &  &  &  &  &  &  &  &  \\
\midrule
\multirow[c]{11}{*}{MTEB} & bge-small-en-v1.5 & 33M & 83.8 & 79.4 & \bfseries \underline{77.4} & 83.0 & 81.8 & 87.3 & 84.9 & 87.2 & 65.3 & 85.9 & 81.6 \\
 & snowflake-arctic-embed-s & 33M & 86.3 & 69.7 & 68.8 & 79.6 & 75.6 & 84.6 & 82.4 & 86.7 & \bfseries \underline{69.5} & 81.2 & 78.4 \\
 & bge-small-4096 & 35M & 81.6 & 74.2 & 72.2 & 80.5 & 76.2 & 85.2 & 81.9 & 86.6 & 65.5 & 81.9 & 78.6 \\
 & GIST-small-Embedding-v0 & 33M & 87.0 & \bfseries \underline{80.5} & 75.6 & \bfseries \underline{86.3} & \bfseries \underline{82.3} & \underline{88.7} & \bfseries \underline{85.3} & \underline{89.0} & 68.5 & \bfseries \underline{87.1} & \bfseries \underline{83.0} \\
 & NoInstruct-small-Embedding-v0 & 33M & \underline{87.2} & \underline{80.3} & 75.8 & \underline{86.1} & \underline{82.3} & \bfseries \underline{88.9} & \underline{85.2} & 88.7 & \underline{68.5} & \underline{87.0} & \underline{83.0} \\
 & e5-small & 33M & 84.2 & 78.9 & 75.2 & 81.8 & 78.5 & 87.5 & 84.6 & 87.9 & 63.8 & 86.4 & 80.9 \\
 & e5-small-v2 & 33M & 79.4 & 78.5 & \underline{76.2} & 82.4 & 79.0 & 87.8 & 83.8 & 87.7 & 63.1 & 86.0 & 80.4 \\
 & jina-embedding-s-en-v1 & 35M & 83.0 & 76.3 & 74.3 & 78.5 & 73.8 & 83.7 & 80.0 & 87.5 & 64.2 & 79.2 & 78.1 \\
 & jina-embeddings-v2-small-en & 33M & 80.5 & 76.7 & 73.7 & 83.3 & 79.2 & 87.3 & 83.6 & 88.2 & 63.5 & 84.0 & 80.0 \\
 & all-MiniLM-L12-v2 & 33M & 83.6 & 79.3 & 73.1 & 82.1 & 76.7 & 85.6 & 80.2 & 88.6 & 65.7 & 83.1 & 79.8 \\
 & gte-small & 33M & \bfseries \underline{88.2} & 77.9 & 75.1 & 85.1 & 81.0 & 88.3 & 83.9 & 87.6 & 68.0 & 85.6 & 82.1 \\
\cline{1-14}
MSE & Student-s & 33M & 78.9 & 79.5 & 70.6 & 79.7 & 75.4 & 84.1 & 81.8 & 86.7 & 66.6 & 83.1 & 78.6 \\
\cline{1-14}
NLL & Student-s & 33M & 81.5 & 79.3 & 73.0 & 81.4 & 78.2 & 86.3 & 84.2 & \bfseries \underline{90.0} & 66.0 & 84.8 & 80.5 \\
\cline{1-14}
\bottomrule
\end{tabular}}
\end{table}

\begin{table}
\caption{Performance of our distilled models compared of models of similar sizes 100M to 120M parameters from the MTEB Benchmark on STS tasks.}
\label{tab:mteb_sts_per_size_100_120}
\resizebox{\textwidth}{!}{\begin{tabular}{llc|cccccccccc|c}
\toprule
 & Task & Size & \rotatebox{90}{\shortstack{BIOSSES}} & \rotatebox{90}{\shortstack{SICK-R}} & \rotatebox{90}{\shortstack{STS12}} & \rotatebox{90}{\shortstack{STS13}} & \rotatebox{90}{\shortstack{STS14}} & \rotatebox{90}{\shortstack{STS15}} & \rotatebox{90}{\shortstack{STS16}} & \rotatebox{90}{\shortstack{STS17}} & \rotatebox{90}{\shortstack{STS22}} & \rotatebox{90}{\shortstack{STSBenchmark}} & Avg. \\
  & Model &  &  &  &  &  &  &  &  &  &  &  &  \\
\midrule
\multirow[c]{25}{*}{MTEB} & bge-base-en-v1.5 & 109M & 86.9 & 80.3 & 78.0 & 84.2 & 82.3 & 88.0 & \underline{85.5} & 86.4 & 66.0 & 86.4 & \underline{82.4} \\
 & bilingual-embedding-small & 118M & 84.0 & 74.7 & \underline{79.4} & 85.3 & \underline{83.9} & 88.5 & 84.4 & 85.8 & 67.2 & 86.1 & 81.9 \\
 & multilingual-e5-small & 118M & 82.3 & 77.5 & 76.6 & 77.0 & 75.5 & 87.1 & 83.6 & 86.4 & 60.9 & 84.0 & 79.1 \\
 & snowflake-arctic-embed-m & 109M & 86.6 & 69.1 & 67.0 & 79.1 & 68.5 & 79.9 & 78.7 & 81.5 & 65.8 & 74.1 & 75.0 \\
 & snowflake-arctic-embed-m-v1.5 & 109M & 86.4 & 69.9 & 61.8 & 82.7 & 69.0 & 75.5 & 77.3 & 75.0 & \bfseries \underline{69.1} & 69.7 & 73.6 \\
 & GIST-Embedding-v0 & 109M & \bfseries \underline{88.0} & \bfseries \underline{81.3} & 76.2 & \bfseries \underline{87.8} & 83.4 & \bfseries \underline{89.4} & 85.3 & 88.6 & 67.8 & \bfseries \underline{87.3} & \bfseries \underline{83.5} \\
 & ml-nlp-elser.html & 110M & 83.8 & 68.8 & 64.8 & 80.1 & 75.0 & 83.7 & 80.5 & 85.7 & 67.5 & 79.5 & 76.9 \\
 & e5-base-4k & 112M & 81.4 & 78.3 & 75.8 & 83.6 & 80.0 & \underline{88.8} & 84.5 & 87.6 & 64.1 & \underline{86.5} & 81.0 \\
 & instructor-base & 110M & 82.3 & 80.3 & 77.0 & \underline{86.6} & 81.3 & 88.2 & 84.9 & 89.5 & 66.5 & 86.4 & 82.3 \\
 & bert-base-uncased & 110M & 54.7 & 58.6 & 30.9 & 59.9 & 47.7 & 60.3 & 63.7 & 64.1 & 56.4 & 47.3 & 54.4 \\
 & e5-base & 109M & 85.1 & 79.7 & 74.2 & 83.3 & 78.5 & 88.3 & 84.2 & 87.2 & 62.9 & 86.2 & 81.0 \\
 & e5-base-v2 & 110M & 81.4 & 78.3 & 75.8 & 83.6 & 80.0 & \underline{88.8} & 84.5 & 87.6 & 64.1 & \underline{86.5} & 81.0 \\
 & jina-embedding-b-en-v1 & 110M & 83.6 & 79.1 & 75.1 & 80.9 & 76.1 & 85.5 & 81.2 & 89.0 & 66.2 & 82.6 & 79.9 \\
 & contriever-base-msmarco & 110M & 83.3 & 70.2 & 64.3 & 80.0 & 74.5 & 83.3 & 79.7 & 86.3 & 64.6 & 78.8 & 76.5 \\
 & sup-simcse-bert-base-uncased & 110M & 68.4 & 80.8 & 75.3 & 84.7 & 80.2 & 85.4 & 80.8 & 89.4 & 62.0 & 84.2 & 79.1 \\
 & unsup-simcse-bert-base-uncased & 110M & 72.3 & 72.2 & 66.0 & 81.5 & 73.6 & 79.7 & 78.1 & 83.6 & 59.6 & 76.5 & 74.3 \\
 & all-mpnet-base-v2 & 110M & 80.4 & 80.6 & 72.6 & 83.5 & 78.0 & 85.7 & 80.0 & \bfseries \underline{90.6} & \underline{68.0} & 83.4 & 80.3 \\
 & allenai-specter & 110M & 65.0 & 56.4 & 62.5 & 58.7 & 54.9 & 62.5 & 64.3 & 69.6 & 55.1 & 61.3 & 61.0 \\
 & gtr-t5-base & 110M & 79.0 & 71.5 & 68.6 & 79.1 & 74.6 & 84.8 & 81.6 & 85.8 & 66.2 & 79.6 & 77.1 \\
 & msmarco-bert-co-condensor & 110M & 77.3 & 72.0 & 68.2 & 80.4 & 74.0 & 82.6 & 79.8 & 85.9 & 67.5 & 77.0 & 76.5 \\
 & paraphrase-multilingual-MiniLM-L12-v2 & 118M & 74.2 & 79.6 & 76.0 & 80.7 & 78.8 & 85.8 & 81.0 & 86.9 & 62.1 & 84.4 & 79.0 \\
 & sentence-t5-base & 110M & 75.9 & 80.2 & 78.0 & 85.8 & 82.2 & 87.5 & 84.0 & 89.6 & 62.7 & 85.5 & 81.1 \\
 & text2vec-base-multilingual & 118M & 66.2 & 80.0 & \bfseries \underline{80.9} & 82.9 & \bfseries \underline{87.4} & 88.3 & 81.6 & 85.8 & 63.0 & 86.5 & 80.2 \\
 & gte-base & 109M & \underline{87.6} & 78.9 & 75.7 & 85.7 & 81.5 & 88.8 & 83.8 & 87.9 & 67.3 & 85.7 & 82.3 \\
 & ALL\_862873 & 118M & 21.3 & 48.5 & 55.6 & 18.4 & 28.8 & 29.2 & 39.0 & 61.2 & 44.5 & 44.4 & 39.1 \\
\cline{1-14}
MSE & Student-m & 109M & 83.4 & \underline{80.9} & 74.5 & 82.8 & 79.0 & 86.6 & 85.2 & 88.4 & 66.4 & 85.2 & 81.2 \\
\cline{1-14}
NLL & Student-m & 109M & 85.2 & 80.2 & 75.2 & 83.4 & 80.4 & 88.3 & \bfseries \underline{86.0} & \underline{89.9} & 66.2 & 86.4 & 82.1 \\
\cline{1-14}
\bottomrule
\end{tabular}}
\end{table}

\begin{table}
\caption{Performance of our distilled models compared of models of similar sizes 200M to 400M parameters from the MTEB Benchmark on STS tasks.}
\label{tab:mteb_sts_per_size_200_400}
\resizebox{\textwidth}{!}{\begin{tabular}{llc|cccccccccc|c}
\toprule
 & Task & Size & \rotatebox{90}{\shortstack{BIOSSES}} & \rotatebox{90}{\shortstack{SICK-R}} & \rotatebox{90}{\shortstack{STS12}} & \rotatebox{90}{\shortstack{STS13}} & \rotatebox{90}{\shortstack{STS14}} & \rotatebox{90}{\shortstack{STS15}} & \rotatebox{90}{\shortstack{STS16}} & \rotatebox{90}{\shortstack{STS17}} & \rotatebox{90}{\shortstack{STS22}} & \rotatebox{90}{\shortstack{STSBenchmark}} & Avg. \\
  & Model &  &  &  &  &  &  &  &  &  &  &  &  \\
\midrule
\multirow[c]{20}{*}{MTEB} & gte-multilingual-base & 305M & 81.2 & 79.3 & 77.5 & 85.5 & 81.7 & 89.0 & 84.3 & 88.9 & 67.2 & 86.5 & 82.1 \\
 & bge-large-en-v1.5 & 335M & 84.7 & 81.7 & 79.0 & 86.4 & 82.8 & 88.0 & 86.5 & 87.5 & 67.0 & 87.5 & 83.1 \\
 & MUG-B-1.6 & 335M & 88.4 & \bfseries \underline{83.0} & \underline{79.2} & 89.4 & 84.8 & 89.5 & 86.7 & 89.6 & \bfseries \underline{70.3} & 89.0 & 85.0 \\
 & bilingual-embedding-base & 278M & 87.1 & 79.5 & \bfseries \underline{79.6} & 84.7 & 83.9 & \underline{89.9} & 84.9 & 88.7 & 64.3 & 87.4 & 83.0 \\
 & snowflake-arctic-embed-l & 334M & 86.3 & 69.3 & 67.8 & 77.5 & 69.8 & 80.2 & 77.9 & 82.3 & 68.0 & 75.7 & 75.5 \\
 & UAE-Large-V1 & 335M & 86.1 & 82.6 & 79.1 & 89.6 & 85.0 & 89.5 & 86.6 & 89.0 & 68.8 & 89.1 & 84.5 \\
 & GIST-large-Embedding-v0 & 335M & \bfseries \underline{89.2} & 82.8 & 77.1 & 89.3 & 83.8 & 89.7 & 86.4 & 89.7 & 69.6 & 88.3 & 84.6 \\
 & embedder-100p & 278M & 75.3 & 80.9 & 77.0 & 82.6 & 77.8 & 85.9 & 80.7 & 89.0 & 68.3 & 84.2 & 80.2 \\
 & instructor-large & 335M & 84.4 & 81.3 & 76.3 & 88.2 & 81.9 & 89.0 & 85.5 & \bfseries \underline{90.3} & 67.7 & 86.9 & 83.1 \\
 & e5-large & 335M & 84.7 & 80.5 & 75.9 & 85.2 & 80.5 & 88.8 & 85.3 & 89.4 & 63.0 & 87.2 & 82.1 \\
 & e5-large-v2 & 335M & 83.6 & 79.3 & 77.0 & 84.1 & 80.5 & 89.8 & 85.5 & 89.0 & 64.1 & 87.7 & 82.1 \\
 & multilingual-e5-base & 278M & 85.0 & 78.5 & 76.7 & 78.0 & 76.6 & 88.2 & 84.3 & 87.8 & 62.3 & 85.6 & 80.3 \\
 & sf\_model\_e5 & 335M & 86.8 & 82.3 & 77.6 & 88.0 & 83.8 & 88.5 & 86.5 & 88.7 & 68.0 & 88.3 & 83.8 \\
 & jina-embedding-l-en-v1 & 335M & 84.4 & 79.2 & 74.5 & 83.2 & 78.1 & 86.9 & 83.7 & \underline{90.2} & 64.9 & 84.6 & 81.0 \\
 & ember-v1 & 335M & 85.8 & 81.8 & 78.5 & 86.6 & 83.1 & 88.4 & \underline{86.8} & 87.9 & 66.8 & 87.8 & 83.3 \\
 & mxbai-embed-2d-large-v1 & 335M & 88.1 & 82.0 & 78.8 & \bfseries \underline{90.4} & \underline{85.5} & \bfseries \underline{90.0} & \bfseries \underline{87.4} & 88.8 & 68.8 & \bfseries \underline{89.2} & 84.9 \\
 & mxbai-embed-large-v1 & 335M & 88.4 & \underline{82.9} & 78.8 & \underline{90.3} & \bfseries \underline{85.5} & 89.6 & 86.6 & 89.5 & 69.3 & \underline{89.1} & \underline{85.0} \\
 & paraphrase-multilingual-mpnet-base-v2 & 278M & 76.3 & 79.6 & 77.9 & 85.1 & 80.8 & 87.5 & 83.2 & 87.0 & 63.5 & 86.8 & 80.8 \\
 & gte-large & 335M & 88.7 & 79.8 & 76.8 & 88.1 & 82.7 & 88.9 & 84.2 & 88.5 & \underline{69.7} & 86.1 & 83.3 \\
 & b1ade-embed & 335M & \underline{89.2} & 82.8 & 78.7 & 90.0 & 85.0 & 89.8 & 86.7 & 89.8 & 69.7 & 88.8 & \bfseries \underline{85.0} \\
\cline{1-14}
MSE & Student-l & 335M & 79.1 & 80.6 & 73.7 & 82.1 & 78.1 & 87.4 & 84.2 & 89.1 & 67.0 & 85.3 & 80.7 \\
\midrule
NLL & Student-l & 335M & 83.8 & 79.5 & 74.4 & 83.0 & 79.6 & 88.0 & 85.2 & 90.1 & 65.3 & 86.2 & 81.5 \\
\cline{1-14}
\bottomrule
\end{tabular}}
\end{table}

\subsubsection{Analysis and compare with the most recent embedders}
The results at \autoref{tab:recent-head2head} show that our medium model (\textsc{Student-m-nll}, 109M) achieves an average of 80.2 on the selected MTEB classification tasks, tracking much larger recent embedders within single-digit margins. In particular, \textsc{Qwen3-Embedding-0.6B} (595M) reaches 85.8, a +5.6 point gain at \(\sim5.5\times\) the parameters. Substantially larger improvements appear only beyond \(\sim\)1B parameters (\textsc{jasper\_en\_vision\_language\_v1}, 1.0B: 90.3; \textsc{stella\_en\_1.5B\_v5}, 1.5B: 89.4; \textsc{Qwen3-Embedding-4B}, 4.0B: 89.8). Overall, the 109M model delivers competitive accuracy relative to 4–6\(\times\) larger embedders, supporting our claim that multi-teacher distillation yields high information density at compact scales.

\begin{table*}[t]
\caption{Head-to-head comparison on selected MTEB classification tasks, with large embedders (over x5 times the number of parameters).}
\label{tab:recent-head2head}
\resizebox{\textwidth}{!}{%
\begin{tabular}{llc|cccccccc|c}
\toprule
 & Model & Size & \rotatebox{90}{\shortstack{AmazonCtf}} & \rotatebox{90}{\shortstack{Banking77}} & \rotatebox{90}{\shortstack{IMDB}} & \rotatebox{90}{\shortstack{MTOP Dom.}} & \rotatebox{90}{\shortstack{Massive Int.}} & \rotatebox{90}{\shortstack{Massive Scen.}} & \rotatebox{90}{\shortstack{Toxic Conv.}} & \rotatebox{90}{\shortstack{Tweet Sent.}} & Avg. \\
 &  &  &  &  &  &  &  &  &  &  &  \\
\midrule
\multirow[c]{10}{*}{} 
 & Qwen3-Embedding-4B & 4.0B & 93.7 & 86.3 & 97.2 & 97.8 & 85.0 & 88.8 & 91.4 & 78.4 & 89.8 \\
 & stella\_en\_1.5B\_v5 & 1.5B & 94.1 & 89.8 & 96.7 & 98.7 & 84.5 & 89.7 & 86.8 & 74.8 & 89.4 \\
 & jasper\_en\_vision\_language\_v1 & 1.0B & 93.8 & 87.2 & 97.0 & 99.2 & 85.3 & 91.2 & 91.3 & 77.2 & 90.3 \\
 & Qwen3-Embedding-0.6B & 595M & 91.5 & 81.0 & 95.4 & 96.0 & 80.4 & 83.6 & 82.1 & 76.0 & 85.8 \\
 & jina-embeddings-v3 & 572M & 90.9 & 84.1 & 91.9 & --   & 75.2 & 84.1 & 91.3 & 71.4 & 84.1 \\
 & snowflake-arctic-embed-l-v2.0 & 568M & 65.6 & 81.8 & 72.8 & 93.5 & 71.5 & 76.2 & 65.9 & 59.6 & 73.4 \\
 & KaLM-embed-mini-instr-v2 & 494M & 95.3 & 89.5 & 95.2 & 98.9 & 77.8 & 86.0 & 89.3 & 78.6 & 88.8 \\
 & KaLM-embed-mini-instr-v1 & 494M & 81.5 & 84.9 & 95.0 & 92.2 & 69.8 & 74.2 & 89.0 & 76.5 & 82.9 \\
 & KaLM-embed-mini-v1 & 494M & 76.4 & 79.2 & 91.6 & 92.5 & 70.9 & 76.1 & 70.8 & 62.7 & 77.5 \\
 & stella\_en\_400M\_v5 & 435M & 94.3 & 89.3 & 96.5 & 98.3 & 80.5 & 89.6 & 84.0 & 73.6 & 88.2 \\
\midrule
\multirow[c]{2}{*}{NLL} 
 & Student-m-nll & 109M & 79.6 & 88.0 & 88.3 & 96.2 & 78.6 & 82.7 & 67.1 & 61.3 & 80.2 \\
 & Student-s-nll & 32M & 77.3 & 86.7 & 88.3 & 95.5 & 76.7 & 80.7 & 66.1 & 60.6 & 79.0 \\
\cline{1-12}
\bottomrule
\end{tabular}}
\end{table*}

\FloatBarrier

    \section{Vision}
\subsection{Model architecture}
\label{sec:appendix_vision_models}
The models we used for vision as teachers and student are presented in ~\autoref{tab:vision_models}, including the number of parameters of each of them.

\begin{table}
\caption{Number of parameters for each model (in million parameters)}
\label{tab:vision_models}
\centering
\begin{tabular}{lr}
\toprule
Model & \# Parameters \\
\midrule
Swin \citep{liu2021swin} & 87.77M \\ 
DINOv2 \citep{oquab2023dinov2} & 86.58M \\
ViT \citep{dosovitskiy2021imageworth16x16words} & 86.57M \\
BEiT \citep{bao2022beitbertpretrainingimage} & 86.53M \\
PVTv2 \citep{wang2022pvt} & 3.67M \\
WideResNet \citep{zagoruyko2017wideresidualnetworks} & 68.88M \\
DenseNet \citep{huang2017densely} & 28.68M \\
ResNext \citep{xie2017aggregated} & 25.03M \\ 
ResNet18 \citep{he2016deep} & 11.69M \\
GoogLeNet \citep{szegedy2015going} & 6.62M \\ 
MNASNet \citep{tan2019mnasnet} & 4.38M \\
MobileNet \citep{sandler2018mobilenetv2} & 3.50M \\ 
ShuffleNet \citep{ma2018shufflenet} & 2.28M \\
SqueezeNet \citep{iandola2016squeezenetalexnetlevelaccuracy50x} & 1.25M \\
\bottomrule
\end{tabular}
\end{table}

\subsection{Training Set}
\label{sec:appendix_vision_training_set}
~\autoref{tab:vision_datasets} presents the statistics, \textit{i.e.} the number of training and testing samples, of the datasets we used for vision. 
\begin{table}
\caption{Number of classes, training, validation (if any) and testing samples in each vision dataset}
\label{tab:vision_datasets}
\centering \begin{tabular}{lrrrr}
\toprule
Dataset & classes & training samples & validation samples & test samples \\
\midrule
 CIFAR10 \citep{krizhevsky2009learning} & 10 & 50000 & - & 10000\\
 STL10 \citep{coates2011analysis} & 10 & 5000 & - & 8000\\
 SVHN \citep{netzer2011reading} & 10 & 73257 & - & 26032\\
 CUB \citep{WelinderEtal2010} & 200 & 5,994 & - & 5,794 \\
 DTD \citep{cimpoi14describing} & 47 & 1880 & 1880 & 1880 \\
 FGVCAircraft \citep{maji13fine-grained} & 100 & 3334 & 3333 & 3333 \\
 \toprule
 Oxford Pets \citep{6248092} & 37 & 3680  & - & 8041\\
 Food101 \citep{10.1007/978-3-319-10599-4_29} & 101 & 750 & - & 250\\
 Stanford Cars \citep{6755945} & 196 & 8144  & - & 8041\\
\bottomrule
\end{tabular}

\end{table}

\subsection{Vision Details}
\label{sec:appendix_vision_det}
\paragraph{Data processing details:} We use the official train sets of the datasets for the knowledge distillation part. We split the official training part, if there are no official validation sets, to train and validation set with 80 and 20 percents of the data, consequently. For the augmentation we used color jitter with brightness, contrast, saturation and hue equal to 0.2, and random horizontal flip (except for the SVHN dataset).
\paragraph{Distillation details: } For training the distillation, we extract the embeddings of the train set of each dataset, for each teacher and divide the  embeddings to 80 train set and 20 percent validation set. For the optimizer we use Adam, with learning rate of $0.001$, a batch size of 128, trained for $50$ epochs. 
\paragraph{Down-stream task fine-tuning:}
For fine-tuning of down-stream tasks, we add a classifier on the frozen embedders. We again use Adam optimizer for the fine-tuning of downstream tasks.
We perform hyperparameter tuning using grid search to optimize the performance of our models. Our search space includes the learning rate with values (1e-2, 1e-3), the number of fully connected layer units with values (0, 128), and the type of normalization after the fully connected layer, considering (no optimization, batch normalization, layer normalization). The models are trained for a maximum of 1000 epochs with a batch size of 128, but we apply early stopping with a patience of 20 to prevent over-fitting and reduce unnecessary computation. 
\subsection{Complementary Results}
\label{sec:appendix_vision_comp_results}

 ~\autoref{tab:vision_vit_table} shows the detailed results of the Vision Transformer teachers and students. The best among the students are shown with an underline, showing that on average and most of the cases our method improves the baseline. In addition to the main results, we added additional experiments to answer further informative question:

\textbf{How will our method work in vision for unseen datasets?} ~\autoref{tab:vision_vit_table_od} shows the accuracy of our student compared to various distillation baselines: MSE distillation, Cosine distillation,  Correlation Congruence (CC rbf and CC dot)~\cite{cc}, CompRess~\cite{NEURIPS2020_975a1c8b} and relational KD~\cite{rkd}.

for three unseen datasets. As we can see, our method improved the baselines considerably for unseen datasets.
\begin{table}
\caption{Comparison of Vision Transformer teachers, CNN baselines and the ViT student, with their corresponding parameter size, with the underline showing the best students.}
\resizebox{\textwidth}{!}{\begin{tabular}{rrr|cccccc}
\toprule
{Method} & {Model} & {\# Parameters} & {CIFAR10} & {DTD} & {STL10} & {SVHN} & {FGVCAircraft} & {CUB} \\
\midrule
\multirow[c]{9}{*}{NoKD} & Swin & 87.77 & 97.67 & 76.33 & \bfseries 99.60 & 64.42 & 52.45 & 87.11\\
 & ViT & 86.57 & 96.90 & 71.65 & 99.40 & 54.97 & 41.71 & 82.67 \\
 & DINOv2 & 86.58 & \bfseries 98.57 & \bfseries 83.30 & 99.45 & 63.01 & \bfseries 79.40 & \bfseries 89.02 \\
 & BEiT & 86.53 & 97.89 & 77.34 & \bfseries 99.60 & 66.61 & 55.45 & 39.52 \\
 & PVTv2 & 3.67 & 89.27 & 65.05 & 95.80 & 62.03 & 38.58 & 68.97 \\
 & wide resnet & 68.88 & 85.65 & 65.37 & 95.85 & 57.77 & 30.82 & 60.55 \\
 & densenet & 28.68 & 87.49 & 67.93 & 97.11 & 66.91 & 46.84 & 68.62 \\
 & resnet18 & 11.69 & 83.22 & 61.54 & 92.98 & 51.01 & 36.09 & 59.89 \\
 & googlenet & 6.62 & 82.07 & 66.38 & 93.95 & 55.90 
 & 35.85 & 59.09 \\
\midrule
CompRess & PVTv2 & 3.67 & 94.6 & 52.7 & 93.5 & 61.9 & 32.7 & 48.8 \\
MSE & PVTv2 & 3.67 & \underline{96.1} & 65.1 & 96.4 & 70.3 & 34.4 & 67.7 \\
Cosine & PVTv2 & 3.67 & 95.89 & 65.4 & \underline{96.7} & 70.7 & 35.9 & 67.1 \\
RKD & PVTv2  & 3.67 & 87.64 & 52.23 & 89.63 & 61.66 & 30.54 & 47.85 \\
CC grbf     & PVTv2  & 3.67 & 84.07 & 61.86 & 93.03 & 59.96 & 33.48 & 57.55 \\
CC bilinear & PVTv2  & 3.67 & 92.95 & 61.22 & 95.42 & 63.71 & 35.16 & 64.70 \\
NLL & PVTv2  & 3.67 & 94.76 & \underline{65.85} & 96.45 & \bfseries \underline{76.91} & \underline{48.13} & \underline{69.37} \\
\bottomrule
\end{tabular}
}
\label{tab:vision_vit_table}
\end{table}
\begin{table}
\centering
\caption{Comparison of ViT student of our method (NLL), and various distillation baselines for the unseen datasets.}
\begin{tabular}{r|ccc}
        \toprule
        Method & Oxford Pets & Food101 & Stanford Cars \\
        \midrule
        CompRess & 70.23 & 45.48 & 19.43 \\
        MSE & 85.58 & 58.04 & 31.96 \\
        Cosine & 84.38 & 56.37 & 30.92 \\
        RKD & 69.99 & 43.48 & 18.24 \\
        CC rbf & 85.09 & 58.47 & 30.08 \\
        CC dot & 67.42 & 45.93 & 20.88 \\
        NLL & \textbf{87.46} & \textbf{62.62} & \textbf{41.29} \\
        \bottomrule
\end{tabular}

\label{tab:vision_vit_table_od}
\end{table}

\textbf{How our method works for a setting with diverse teachers specialized in different task, and if it will be able to avoid conflicts?} We evaluated the student model's classification performance using three specialized vision teachers: ViT (classification), DETR (~\citep{carion2020end} , object detection), and SegFormer  (~\citep{xie2021segformer}, segmentation). We also included DINOv2, a general-purpose embedding model known for strong performance across multiple benchmarks. As shown in~\autoref{tab:teacher_performance_div}, adding DETR or SegFormer alongside ViT did not significantly improve or degrade classification performance compared to using ViT alone. This suggests that while task-specific teachers may offer limited benefit outside their domain, they do not negatively impact the student’s learning.

To further validate this, we incorporated DINOv2 into the teacher set (~\autoref{tab:teacher_performance_dino}). This addition improved overall performance, while the inclusion of DETR and SegFormer continued to have minimal effect, confirming that our earlier observations hold even in a more competitive setting with a strong general-purpose teacher. These results are consistent with ~\autoref{sec:mol_model_results}
 and ~\autoref{sec:appendix_nlp_single}, where we observe that adding teachers typically boosts student performance. In molecular and text domains, where all teachers are general-purpose embedders, improvements are more uniform. However, in vision tasks, specialized teachers contribute gains primarily in their area of expertise, yet without harming performance elsewhere. Overall, these findings suggest that our method can effectively integrate knowledge from both specialized and generalist teachers without conflict.

\begin{table}[ht]
\centering
\caption{Performance of different teacher combinations across datasets (accuracy \%).}
\resizebox{\textwidth}{!}{\begin{tabular}{r|ccccccc}
\toprule
\textbf{Teachers} & \textbf{CIFAR-10} & \textbf{DTD} & \textbf{STL-10} & \textbf{SVHN} & \textbf{FGVC} & \textbf{CUB} & \textbf{Average} \\
\midrule
ViT + Segformer + DETR & 94.03 & 63.62 & 95.86 & 65.63 & 38.79 & 67.67 & \textbf{70.93} \\
ViT + Segformer        & 94.23 & 63.24 & \textbf{95.91} & \textbf{65.79} & 38.31 & 67.35 & 70.81 \\
ViT + DETR             & \textbf{94.71} & 61.28 & 95.80 & 64.14 & 37.89 & 65.90 & 69.95 \\
ViT                    & 94.69 & 61.70 & 95.75 & 64.13 & \textbf{39.42} & \textbf{69.23} & 70.82 \\
DETR + Segformer       & 87.87 & \textbf{63.72} & 94.81 & 54.71 & 37.89 & 62.43 & 66.91 \\
\bottomrule
\end{tabular}
}
\label{tab:teacher_performance_div}
\end{table}

\begin{table}[ht]
\centering
\caption{Comparison of ViT-based teacher combinations including DINO on multiple datasets (accuracy \%). Bolded values indicate best per column.}
\resizebox{\textwidth}{!}{\begin{tabular}{r|ccccccc}
\toprule
\textbf{Teachers} & \textbf{CIFAR-10} & \textbf{DTD} & \textbf{STL-10} & \textbf{SVHN} & \textbf{FGVC} & \textbf{CUB} & \textbf{Average} \\
\midrule
ViT + Segformer + DETR + DINO & \textbf{95.39} & 64.31 & 96.14 & \textbf{72.88} & \textbf{50.38} & 69.69 & \textbf{74.80} \\
ViT + DINO                    & 95.83 & \textbf{61.92} & \textbf{96.06} & 73.60 & 50.59 & \textbf{69.21} & 74.54 \\
\bottomrule
\end{tabular}
}
\label{tab:teacher_performance_dino}
\end{table}

As another additional experiment, we use CNN based teachers for resnet18, for different relevant datasets. ~\autoref{tab:teacher_performance_cnn} shows the performance improvements, and the effectiveness of using our distillation method, compared to other.
\begin{table}[ht]
\centering
\caption{Comparison of the performance with CNN-based teacher (accuracy \%). Bolded values indicate best per column.}
\resizebox{\textwidth}{!}{\begin{tabular}{rr|cccccccc}
\toprule
{Method} & {Model} & {CIFAR10} & {FMNIST} & {MNIST} & {STL10} & {SVHN} & {QMNIST} & {KMNIST} & {CelebA}  \\
\midrule  
\multirow[c]{9}{*}{NoKD} 
 & resnet18 & 81.89 & 86.94 & 96.6 & 92.98 & 51.01 & 96.89 & 80.43 & 90.82  \\
 & squeezenet & 79.23 & 86.65 & 97.51 & 85.82 & 47.77 & 97.59 & 84.05 & 61.35  \\
 & densenet & \underline{87.49} & 88.69 & 96.80 & \bfseries \underline{97.11} & 66.91 & 97.72 & 86.33& 93.98  \\
 & googlenet & 81.94 & 86.38 & 96.71 & 93.95 & 55.9 & 97.2 & 79.27 & 92.93  \\
 & shufflenet & 81.61 & 87.57 & 95.77 & 71.51 & 49.08 & 95.96 & 76.97 & 92.42  \\
 & mobilenet & 81.67 & 88.07 & 96.05 & 92.26 & 48.57 & 97.5 & 85.64 & 91.02  \\ 
 & mnasnet & 81.41 & 88.76 & 96.09 & 92.79 & 57.63 & 97.00 & 82.35 & 89.01  \\
 & resnext50-32x4d & 83.42 & 87.32 & 95.37 & \underline{95.97} & 52.87 & 96.65 & 83.37 & 91.74  \\
 & wide-resnet50-2 & 84.30 & 87.40 & 95.16 & 95.85 & 57.77 & 96.74 & 76.23 & 90.22  \\ 
\midrule 
Cosine & resnet18 & 84.57 & \underline{89.90} & \underline{98.58} & 88.34 & \underline{76.34} & \underline{98.95} & \underline{91.97} & \underline{95.00}  \\ 
\midrule 
L2 & resnet18 & 82.90 & 89.75 & 98.25 & 88.15 & 74.84 & 98.61 & 88.21 & 94.89  \\ 
\midrule 
NLL & resnet18 & \bfseries \underline{87.51} & \bfseries \underline{90.64} & \bfseries \underline{99.15} & 88.45 & \bfseries \underline{81.99} & \bfseries \underline{99.15} & \bfseries \underline{95.21} & \bfseries \underline{95.47}  \\
\bottomrule
\end{tabular}
}
\label{tab:teacher_performance_cnn}
\end{table}
\FloatBarrier

\FloatBarrier

\section{Detailed Method}
\label{sec:appendices_detailed_method}

\begin{algorithm}[H]
    \caption{Distillation through Gaussian Kernels}
    \label{alg:distillation}
    \begin{algorithmic}
        \STATE \textbf{Input:} Dataset $D = \set{\xbold_i}$, Embedders $(\teachk)_{1 \leq k \leq K}$, Student embedder $\stud$, Number of iterations $T$, Learning rate $\eta$
        
        \STATE Initialize the parameters $\theta_s$ of the student embedder $E_s$ and the parameters $\theta_k$ of the parametric Gaussian kernels
        
        \FOR{$t = 1$ to $T$}
            \STATE Sample a batch of inputs $\set{\xbold_i}$
            \STATE Compute the embeddings $\set{\mathbf{t}^k_i = \teachk(\xbold_i)}_{1 \leq k \leq K}$
            \STATE Compute the student embeddings $\set{\mathbf{s}_i = \stud(\xbold_i)}$
            \STATE Compute the loss $\mathcal{L}_{NLL} = - \sum_{k=1}^K \sum_{i=1}^N \log \mathcal{N}(\mathbf{t}^k_i | \mu_k(\mathbf{s}_i), \Sigma_k(\mathbf{s}_i))$
            \STATE Update the parameters $\theta_s$ and $\theta_k$ using the Adam optimizer.
        \ENDFOR
    \end{algorithmic}
\end{algorithm}

\FloatBarrier

\section{Computaional ressources}
\label{sec:appendices_compute_ressources}

Our experiments were conducted in single GPUs settings. We used NVIDIA V100 GPUs for about $3000$ GPUs hours to train our different models.

\section{Baselines}
\label{sec:appendix_baselines}
For the MSE, we will optimize the following loss function following SimReg strategy~\citep{navaneet2022simregregressionsimpleeffective}.
\begin{equation}
    \mathcal{L}_{MSE} = - \sum_{k=1}^K \sum_{i=1}^N \norm{\stud(\xbold_i) - \teachk(\xbold_i)}^2,
\end{equation}
where it calculates the summation of MSE between the representation produced by each teacher and the student, for each instance of the batch.

Variant of SimReg can be implemented for Cosine multi-teacher feature distillation\citep{gao2022discoremedyselfsupervisedlearning, navaneet2022simregregressionsimpleeffective}, we optimize the summation of cosine of teachers and the students representations of each instance of the batch, \textit{i.e.}:
\begin{equation}
    \mathcal{L}_{Cosine} = - \sum_{k=1}^K \sum_{i=1}^N \frac{\stud(\xbold_i).\teachk(\xbold_i)}{\max(\norm{\stud(\xbold_i)}_2.\norm{\teachk(\xbold_i)}_2, \epsilon)}. 
\end{equation}

\FloatBarrier

    \section{Discussion On MSE distillation}

\begin{figure}
    \centering
    \includegraphics[width=0.5\linewidth]{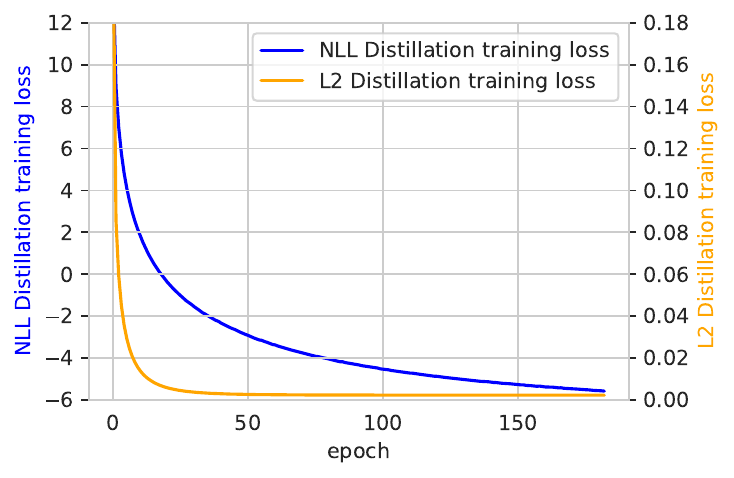}
    \caption{Training curves for the MSE baseline and the NLL student for the molecular experiments.}
    \label{fig:mse_curve}
\end{figure}

We observed that when training with the MSE loss, the loss reaches a minimum in only a few epochs (~40), but the distilled students achieve lower performances on downstream tasks. This could be due to the fact that the NLL loss is more expressive, and harder to optimize (see below). As a result the student learns more informative features compared to when trained with the MSE loss (\autoref{fig:mse_curve}).

We can provide a theoretical insight to explain this phenomenon. Training using the negative log-likelihood over a Gaussian kernel is a simple generalization of the MSE. For a given multivariate Gaussian kernel parameterized by $\mu$ and $\Sigma$, we have:
\[
-\log(p_{\mu, \Sigma}(x)) = \log(C) + \frac{1}{2}\log{\det{\Sigma}} +  \frac{1}{2} (x-\mu)^T\Sigma^{-1}(x-\mu)
\]

Minimizing the MSE loss boils down to minimizing this equation over only, with $\Sigma = I$. Therefore, minimizing the negative log-likelihood of a Gaussian kernel is strictly more expressive than minimizing the MSE directly, which could account for the performance gains we observe.

\section*{NeurIPS Paper Checklist}

\begin{enumerate}

\item {\bf Claims}
    \item[] Question: Do the main claims made in the abstract and introduction accurately reflect the paper's contributions and scope?
    \item[] Answer: \answerYes{}
    \item[] Justification: We propose an objective to train model through multi teacher distillation, give theoretical grounding to it and validate it experimentally that it works by traning embedders in 3 modalities for a wide range of size and domains.
    \item[] Guidelines:
    \begin{itemize}
        \item The answer NA means that the abstract and introduction do not include the claims made in the paper.
        \item The abstract and/or introduction should clearly state the claims made, including the contributions made in the paper and important assumptions and limitations. A No or NA answer to this question will not be perceived well by the reviewers. 
        \item The claims made should match theoretical and experimental results, and reflect how much the results can be expected to generalize to other settings. 
        \item It is fine to include aspirational goals as motivation as long as it is clear that these goals are not attained by the paper. 
    \end{itemize}

\item {\bf Limitations}
    \item[] Question: Does the paper discuss the limitations of the work performed by the authors?
    \item[] Answer: \answerYes{}
    \item[] Justification: Yes, see for example "Embedding space structure"\autoref{sec:nlp_perfs}, and "Computational complexity." in~\autoref{sec:mol_model_results}.
    \item[] Guidelines:
    \begin{itemize}
        \item The answer NA means that the paper has no limitation while the answer No means that the paper has limitations, but those are not discussed in the paper. 
        \item The authors are encouraged to create a separate "Limitations" section in their paper.
        \item The paper should point out any strong assumptions and how robust the results are to violations of these assumptions (e.g., independence assumptions, noiseless settings, model well-specification, asymptotic approximations only holding locally). The authors should reflect on how these assumptions might be violated in practice and what the implications would be.
        \item The authors should reflect on the scope of the claims made, e.g., if the approach was only tested on a few datasets or with a few runs. In general, empirical results often depend on implicit assumptions, which should be articulated.
        \item The authors should reflect on the factors that influence the performance of the approach. For example, a facial recognition algorithm may perform poorly when image resolution is low or images are taken in low lighting. Or a speech-to-text system might not be used reliably to provide closed captions for online lectures because it fails to handle technical jargon.
        \item The authors should discuss the computational efficiency of the proposed algorithms and how they scale with dataset size.
        \item If applicable, the authors should discuss possible limitations of their approach to address problems of privacy and fairness.
        \item While the authors might fear that complete honesty about limitations might be used by reviewers as grounds for rejection, a worse outcome might be that reviewers discover limitations that aren't acknowledged in the paper. The authors should use their best judgment and recognize that individual actions in favor of transparency play an important role in developing norms that preserve the integrity of the community. Reviewers will be specifically instructed to not penalize honesty concerning limitations.
    \end{itemize}

\item {\bf Theory assumptions and proofs}
    \item[] Question: For each theoretical result, does the paper provide the full set of assumptions and a complete (and correct) proof?
    \item[] Answer: \answerYes{}
    \item[] Justification: We provide proofs for our corollary in the supplementary materials. See \autoref{sec:from_setting_to_loss} and \autoref{sec:appendices_detailed_method}.
    \item[] Guidelines:
    \begin{itemize}
        \item The answer NA means that the paper does not include theoretical results. 
        \item All the theorems, formulas, and proofs in the paper should be numbered and cross-referenced.
        \item All assumptions should be clearly stated or referenced in the statement of any theorems.
        \item The proofs can either appear in the main paper or the supplemental material, but if they appear in the supplemental material, the authors are encouraged to provide a short proof sketch to provide intuition. 
        \item Inversely, any informal proof provided in the core of the paper should be complemented by formal proofs provided in appendix or supplemental material.
        \item Theorems and Lemmas that the proof relies upon should be properly referenced. 
    \end{itemize}

    \item {\bf Experimental result reproducibility}
    \item[] Question: Does the paper fully disclose all the information needed to reproduce the main experimental results of the paper to the extent that it affects the main claims and/or conclusions of the paper (regardless of whether the code and data are provided or not)?
    \item[] Answer: \answerYes{}
    \item[] Justification: All datasets, checkpoints and architectures are open source and under permissive licences. The code for our particular experiments is available in different repositories for each modalities with the specific hyperparameters and how to rerun the experiments. 
    Our code is publicly available through anonymous links provided in the core of the paper.
    Regardless of the code we provided the details of datasets, models, hyperparameters and tuning setting (see ~\autoref{sec:appendix_vision_training_set}, ~\autoref{sec:appendix_vision_det}, ~\autoref{sec:appendix_nlp_hyperparameters}, ~\autoref{sec:appendix_nlp_model_dataset_statistics}, ~\autoref{sec:appendix_mol_model_eval_details}, and ~\autoref{sec:appendix_mol_model_model_architecture}).
    \item[] Guidelines:
    \begin{itemize}
        \item The answer NA means that the paper does not include experiments.
        \item If the paper includes experiments, a No answer to this question will not be perceived well by the reviewers: Making the paper reproducible is important, regardless of whether the code and data are provided or not.
        \item If the contribution is a dataset and/or model, the authors should describe the steps taken to make their results reproducible or verifiable. 
        \item Depending on the contribution, reproducibility can be accomplished in various ways. For example, if the contribution is a novel architecture, describing the architecture fully might suffice, or if the contribution is a specific model and empirical evaluation, it may be necessary to either make it possible for others to replicate the model with the same dataset, or provide access to the model. In general. releasing code and data is often one good way to accomplish this, but reproducibility can also be provided via detailed instructions for how to replicate the results, access to a hosted model (e.g., in the case of a large language model), releasing of a model checkpoint, or other means that are appropriate to the research performed.
        \item While NeurIPS does not require releasing code, the conference does require all submissions to provide some reasonable avenue for reproducibility, which may depend on the nature of the contribution. For example
        \begin{enumerate}
            \item If the contribution is primarily a new algorithm, the paper should make it clear how to reproduce that algorithm.
            \item If the contribution is primarily a new model architecture, the paper should describe the architecture clearly and fully.
            \item If the contribution is a new model (e.g., a large language model), then there should either be a way to access this model for reproducing the results or a way to reproduce the model (e.g., with an open-source dataset or instructions for how to construct the dataset).
            \item We recognize that reproducibility may be tricky in some cases, in which case authors are welcome to describe the particular way they provide for reproducibility. In the case of closed-source models, it may be that access to the model is limited in some way (e.g., to registered users), but it should be possible for other researchers to have some path to reproducing or verifying the results.
        \end{enumerate}
    \end{itemize}

\item {\bf Open access to data and code}
    \item[] Question: Does the paper provide open access to the data and code, with sufficient instructions to faithfully reproduce the main experimental results, as described in supplemental material?
    \item[] Answer: \answerYes{}
    \item[] Justification: We leveraged datasets and models available on the huggingface hub and  torchvision datasets for computer vision and NLP (apache 2.0, CC BY-NC-SA 4.0, CC BY-SA, CC BY-NC 4.0, CC BY 4.0 license), in molecules we gathered publicly available datasets under the apache 2.0 license and MIT license.
    All models and datasets are correctly cited in the paper.
    Code and data are publicly available in different repositories available  in the main body of the paper.
    \item[] Guidelines:
    \begin{itemize}
        \item The answer NA means that paper does not include experiments requiring code.
        \item Please see the NeurIPS code and data submission guidelines (\url{https://nips.cc/public/guides/CodeSubmissionPolicy}) for more details.
        \item While we encourage the release of code and data, we understand that this might not be possible, so “No” is an acceptable answer. Papers cannot be rejected simply for not including code, unless this is central to the contribution (e.g., for a new open-source benchmark).
        \item The instructions should contain the exact command and environment needed to run to reproduce the results. See the NeurIPS code and data submission guidelines (\url{https://nips.cc/public/guides/CodeSubmissionPolicy}) for more details.
        \item The authors should provide instructions on data access and preparation, including how to access the raw data, preprocessed data, intermediate data, and generated data, etc.
        \item The authors should provide scripts to reproduce all experimental results for the new proposed method and baselines. If only a subset of experiments are reproducible, they should state which ones are omitted from the script and why.
        \item At submission time, to preserve anonymity, the authors should release anonymized versions (if applicable).
        \item Providing as much information as possible in supplemental material (appended to the paper) is recommended, but including URLs to data and code is permitted.
    \end{itemize}

\item {\bf Experimental setting/details}
    \item[] Question: Does the paper specify all the training and test details (e.g., data splits, hyperparameters, how they were chosen, type of optimizer, etc.) necessary to understand the results?
    \item[] Answer: \answerYes{}
    \item[] Justification: For every modality we share hyperparameter choices and specifics of the model in the specific sections as well as the detailed experiments in appendices (see ~\autoref{sec:appendix_vision_training_set}, ~\autoref{sec:appendix_vision_det},
    ~\autoref{sec:appendix_nlp_hyperparameters}, and ~\autoref{sec:appendix_mol_model_eval_details}).
    \item[] Guidelines:
    \begin{itemize}
        \item The answer NA means that the paper does not include experiments.
        \item The experimental setting should be presented in the core of the paper to a level of detail that is necessary to appreciate the results and make sense of them.
        \item The full details can be provided either with the code, in appendix, or as supplemental material.
    \end{itemize}

\item {\bf Experiment statistical significance}
    \item[] Question: Does the paper report error bars suitably and correctly defined or other appropriate information about the statistical significance of the experiments?
    \item[] Answer: \answerYes{}
    \item[] Justification: We report comprehensive results in appendices for every task with error bars and subset specific evaluations.
    \item[] Guidelines:
    \begin{itemize}
        \item The answer NA means that the paper does not include experiments.
        \item The authors should answer "Yes" if the results are accompanied by error bars, confidence intervals, or statistical significance tests, at least for the experiments that support the main claims of the paper.
        \item The factors of variability that the error bars are capturing should be clearly stated (for example, train/test split, initialization, random drawing of some parameter, or overall run with given experimental conditions).
        \item The method for calculating the error bars should be explained (closed form formula, call to a library function, bootstrap, etc.)
        \item The assumptions made should be given (e.g., Normally distributed errors).
        \item It should be clear whether the error bar is the standard deviation or the standard error of the mean.
        \item It is OK to report 1-sigma error bars, but one should state it. The authors should preferably report a 2-sigma error bar than state that they have a 96\% CI, if the hypothesis of Normality of errors is not verified.
        \item For asymmetric distributions, the authors should be careful not to show in tables or figures symmetric error bars that would yield results that are out of range (e.g. negative error rates).
        \item If error bars are reported in tables or plots, The authors should explain in the text how they were calculated and reference the corresponding figures or tables in the text.
    \end{itemize}

\item {\bf Experiments compute resources}
    \item[] Question: For each experiment, does the paper provide sufficient information on the computer resources (type of compute workers, memory, time of execution) needed to reproduce the experiments?
    \item[] Answer: \answerYes{}
    \item[] Justification: We both discuss the method complexity in the main paper and used ressources in \autoref{sec:appendices_compute_ressources} and \autoref{sec:mol_model_results}.
    \item[] Guidelines:
    \begin{itemize}
        \item The answer NA means that the paper does not include experiments.
        \item The paper should indicate the type of compute workers CPU or GPU, internal cluster, or cloud provider, including relevant memory and storage.
        \item The paper should provide the amount of compute required for each of the individual experimental runs as well as estimate the total compute. 
        \item The paper should disclose whether the full research project required more compute than the experiments reported in the paper (e.g., preliminary or failed experiments that didn't make it into the paper). 
    \end{itemize}
    
\item {\bf Code of ethics}
    \item[] Question: Does the research conducted in the paper conform, in every respect, with the NeurIPS Code of Ethics \url{https://neurips.cc/public/EthicsGuidelines}?
    \item[] Answer: \answerYes{}
    \item[] Justification: The research conducted in the paper conforms with the NeurIPS Code of Ethics.
    \item[] Guidelines:
    \begin{itemize}
        \item The answer NA means that the authors have not reviewed the NeurIPS Code of Ethics.
        \item If the authors answer No, they should explain the special circumstances that require a deviation from the Code of Ethics.
        \item The authors should make sure to preserve anonymity (e.g., if there is a special consideration due to laws or regulations in their jurisdiction).
    \end{itemize}

\item {\bf Broader impacts}
    \item[] Question: Does the paper discuss both potential positive societal impacts and negative societal impacts of the work performed?
    \item[] Answer: \answerNA{}
    \item[] Justification: \answerNA{}
    \item[] Guidelines:
    \begin{itemize}
        \item The answer NA means that there is no societal impact of the work performed.
        \item If the authors answer NA or No, they should explain why their work has no societal impact or why the paper does not address societal impact.
        \item Examples of negative societal impacts include potential malicious or unintended uses (e.g., disinformation, generating fake profiles, surveillance), fairness considerations (e.g., deployment of technologies that could make decisions that unfairly impact specific groups), privacy considerations, and security considerations.
        \item The conference expects that many papers will be foundational research and not tied to particular applications, let alone deployments. However, if there is a direct path to any negative applications, the authors should point it out. For example, it is legitimate to point out that an improvement in the quality of generative models could be used to generate deepfakes for disinformation. On the other hand, it is not needed to point out that a generic algorithm for optimizing neural networks could enable people to train models that generate Deepfakes faster.
        \item The authors should consider possible harms that could arise when the technology is being used as intended and functioning correctly, harms that could arise when the technology is being used as intended but gives incorrect results, and harms following from (intentional or unintentional) misuse of the technology.
        \item If there are negative societal impacts, the authors could also discuss possible mitigation strategies (e.g., gated release of models, providing defenses in addition to attacks, mechanisms for monitoring misuse, mechanisms to monitor how a system learns from feedback over time, improving the efficiency and accessibility of ML).
    \end{itemize}
    
\item {\bf Safeguards}
    \item[] Question: Does the paper describe safeguards that have been put in place for responsible release of data or models that have a high risk for misuse (e.g., pretrained language models, image generators, or scraped datasets)?
    \item[] Answer: \answerNA{}
    \item[] Justification: \answerNA{}
    \item[] Guidelines:
    \begin{itemize}
        \item The answer NA means that the paper poses no such risks.
        \item Released models that have a high risk for misuse or dual-use should be released with necessary safeguards to allow for controlled use of the model, for example by requiring that users adhere to usage guidelines or restrictions to access the model or implementing safety filters. 
        \item Datasets that have been scraped from the Internet could pose safety risks. The authors should describe how they avoided releasing unsafe images.
        \item We recognize that providing effective safeguards is challenging, and many papers do not require this, but we encourage authors to take this into account and make a best faith effort.
    \end{itemize}

\item {\bf Licenses for existing assets}
    \item[] Question: Are the creators or original owners of assets (e.g., code, data, models), used in the paper, properly credited and are the license and terms of use explicitly mentioned and properly respected?
    \item[] Answer: \answerYes{}
    \item[] Justification: We attribute all artefacts used and provide direct links to them for reproducibility purposes. All the ressources we used were publicly available under permissive licences.
    \item[] Guidelines:
    \begin{itemize}
        \item The answer NA means that the paper does not use existing assets.
        \item The authors should cite the original paper that produced the code package or dataset.
        \item The authors should state which version of the asset is used and, if possible, include a URL.
        \item The name of the license (e.g., CC-BY 4.0) should be included for each asset.
        \item For scraped data from a particular source (e.g., website), the copyright and terms of service of that source should be provided.
        \item If assets are released, the license, copyright information, and terms of use in the package should be provided. For popular datasets, \url{paperswithcode.com/datasets} has curated licenses for some datasets. Their licensing guide can help determine the license of a dataset.
        \item For existing datasets that are re-packaged, both the original license and the license of the derived asset (if it has changed) should be provided.
        \item If this information is not available online, the authors are encouraged to reach out to the asset's creators.
    \end{itemize}

\item {\bf New assets}
    \item[] Question: Are new assets introduced in the paper well documented and is the documentation provided alongside the assets?
    \item[] Answer: \answerYes{}
    \item[] Justification: Source code is shared through anonymous git repositories and trained models will be released on the Huggingface hub.
    \item[] Guidelines:
    \begin{itemize}
        \item The answer NA means that the paper does not release new assets.
        \item Researchers should communicate the details of the dataset/code/model as part of their submissions via structured templates. This includes details about training, license, limitations, etc. 
        \item The paper should discuss whether and how consent was obtained from people whose asset is used.
        \item At submission time, remember to anonymize your assets (if applicable). You can either create an anonymized URL or include an anonymized zip file.
    \end{itemize}

\item {\bf Crowdsourcing and research with human subjects}
    \item[] Question: For crowdsourcing experiments and research with human subjects, does the paper include the full text of instructions given to participants and screenshots, if applicable, as well as details about compensation (if any)? 
    \item[] Answer: \answerNA{}
    \item[] Justification: \answerNA{}
    \item[] Guidelines:
    \begin{itemize}
        \item The answer NA means that the paper does not involve crowdsourcing nor research with human subjects.
        \item Including this information in the supplemental material is fine, but if the main contribution of the paper involves human subjects, then as much detail as possible should be included in the main paper. 
        \item According to the NeurIPS Code of Ethics, workers involved in data collection, curation, or other labor should be paid at least the minimum wage in the country of the data collector. 
    \end{itemize}

\item {\bf Institutional review board (IRB) approvals or equivalent for research with human subjects}
    \item[] Question: Does the paper describe potential risks incurred by study participants, whether such risks were disclosed to the subjects, and whether Institutional Review Board (IRB) approvals (or an equivalent approval/review based on the requirements of your country or institution) were obtained?
    \item[] Answer: \answerNA{}
    \item[] Justification: \answerNA{}
    \item[] Guidelines:
    \begin{itemize}
        \item The answer NA means that the paper does not involve crowdsourcing nor research with human subjects.
        \item Depending on the country in which research is conducted, IRB approval (or equivalent) may be required for any human subjects research. If you obtained IRB approval, you should clearly state this in the paper. 
        \item We recognize that the procedures for this may vary significantly between institutions and locations, and we expect authors to adhere to the NeurIPS Code of Ethics and the guidelines for their institution. 
        \item For initial submissions, do not include any information that would break anonymity (if applicable), such as the institution conducting the review.
    \end{itemize}

\item {\bf Declaration of LLM usage}
    \item[] Question: Does the paper describe the usage of LLMs if it is an important, original, or non-standard component of the core methods in this research? Note that if the LLM is used only for writing, editing, or formatting purposes and does not impact the core methodology, scientific rigorousness, or originality of the research, declaration is not required.
    \item[] Answer: \answerNA{}
    \item[] Justification: \answerNA{}
    \item[] Guidelines:
    \begin{itemize}
        \item The answer NA means that the core method development in this research does not involve LLMs as any important, original, or non-standard components.
        \item Please refer to our LLM policy (\url{https://neurips.cc/Conferences/2025/LLM}) for what should or should not be described.
    \end{itemize}

\end{enumerate}

    \section{Funding}
        This work was granted access to the HPC resources of IDRIS under the allocation AD011013290R3, and enabled by support provided by Calcul Quebec and the Digital Research Alliance of Canada.
        This work was funded through scholarships by "École de Technologie Supérieure Montreal", "Université Paris-Saclay" and "McGill University".

\end{document}